\pdfoutput=1
\documentclass[11pt,letterpaper]{article}

\usepackage[T1]{fontenc}
\usepackage{newtxtext}
\usepackage{helvet}
\usepackage{courier}
\usepackage{microtype}
\DisableLigatures[i,I]{encoding = T1}
\usepackage[letterpaper,margin=1in]{geometry}
\usepackage[hyphens]{url}
\usepackage{graphicx}
\usepackage[round]{natbib}
\usepackage{caption}
\usepackage{algorithm}
\usepackage{algorithmic}
\DeclareCaptionStyle{ruled}{labelfont=normalfont,labelsep=colon,strut=off}

\usepackage{booktabs}
\usepackage{amsfonts}
\usepackage{nicefrac}
\usepackage{amsmath}
\usepackage{amssymb}
\usepackage{mathtools}
\usepackage{amsthm}
\usepackage{rotating}
\usepackage{enumitem}
\usepackage{multirow}
\usepackage[export]{adjustbox}
\usepackage{subcaption}
\usepackage{xcolor}

\usepackage{tikz}
\usetikzlibrary{positioning, calc, arrows.meta, fit, shapes.geometric, decorations.pathreplacing, backgrounds}
\newtheorem{theorem}{Theorem}
\newtheorem{lemma}[theorem]{Lemma}

\usepackage[colorlinks=true,linkcolor=blue!50!black,citecolor=blue!50!black,urlcolor=blue!50!black]{hyperref}
\hypersetup{
  pdftitle={Learning to Bound: Monotonic, Calibrated Neural Upper Bounds for MPE Inference},
  pdfauthor={Akshay Vyas, Hailiang Dong, Shivvrat Arya, Brij Malhotra, Tahrima Rahman, Vibhav Gogate, Nicholas Ruozzi}
}
\ifdefined\pdfsuppressptexinfo
\fi

\title{Learning to Bound: Monotonic, Calibrated Neural Upper Bounds for MPE Inference}
\author{%
  Akshay Vyas\thanks{Correspondence to: \texttt{akshay.vyas@utdallas.edu}}\quad
  Hailiang Dong\quad
  Shivvrat Arya\quad
  Brij Malhotra\\
  Tahrima Rahman\quad
  Vibhav Gogate\quad
  Nicholas Ruozzi\\[6pt]
  The University of Texas at Dallas
}
\date{}

\begin{document}

\maketitle
\begin{abstract}
Probabilistic graphical models (PGMs) compactly encode structured joint distributions, but Most Probable Explanation (MPE), which finds the highest-probability assignment to unobserved variables, remains computationally intractable in large, densely structured models. Anytime systematic search, such as branch-and-bound (B\&B), is a popular approach, but its efficiency depends on tight admissible upper bounds that are often prohibitively expensive to compute using classical methods like Mini-Bucket Elimination (MBE). 
We introduce Learning to Bound (L2B), a data-driven framework that amortizes bound computation cost by training a neural predictor. L2B targets two key properties required for effective search bounds: admissibility, which is achieved via statistical calibration techniques, and progressive tightening as evidence accumulates, which is enforced through structural architectural constraints that ensure monotonicity by construction. Across challenging benchmarks, L2B delivers tighter, monotonic bounds at orders-of-magnitude lower cost on every PGM family, and yields measurable improvements when integrated with branch-and-bound on high-treewidth instances. 
\end{abstract}

\section{Introduction}

Probabilistic graphical models (PGMs) \citep{koller,darwiche09} provide a principled representation of structured uncertainty, and have been used for applications in diagnosis, planning, computational biology, and computer vision. A central inference task in PGMs is the most probable explanation (MPE) problem, which seeks an assignment to all non-evidence variables that maximizes the probability of the resulting complete assignment. Exact MPE inference is NP-hard in general \citep{mpe_inf}, and its complexity grows rapidly with model size and treewidth. %
The complexity of this task motivates exact inference algorithms that exploit problem structure, as well as approximate and anytime methods that trade computational cost for solution quality.

The cost of exact inference is governed by the structure of
the graphical model. For example, variable elimination \citep{dechter99} has time
and space complexity exponential in treewidth. Search-based exact
solvers, such as AND/OR Branch-and-Bound (AOBB) \citep{AOBB} and
\textsc{Toulbar2} \citep{toulbar2}, instead explore a structured search
space while maintaining a lower bound given by the best solution found
so far and an upper bound on the best solution that can be obtained
below each search node. %
The quality of the upper bound is critical:
tighter bounds allow the solver to prune larger portions of the search
space, while inexpensive bounds allow the solver to evaluate more
nodes within a fixed time budget.

Classical upper-bounding methods, such as mini-bucket elimination (MBE), dual decomposition, and join graph linear programming (JGLP) \citep{mbe,dual_decomp,jglp,globersonetal}, face a persistent tightness-cost tradeoff. Inexpensive bounds are often too weak to guide search effectively, whereas tighter relaxations yield stronger upper bounds and improve pruning, but they require substantially more time and memory. This tradeoff becomes especially restrictive inside branch-and-bound, where the solver must evaluate bounds repeatedly under many partial assignments. As a result, practical solvers often choose between fast but weak pruning and strong but expensive pruning.

In this work, we address this bottleneck by seeking strong upper bounds that can be computed at low cost. The ideal regime for branch-and-bound combines the pruning effectiveness of tight bounds with the efficiency of inexpensive evaluations, enabling aggressive pruning without slowing the search. To approach this regime, we introduce a learning-based method that amortizes the cost of computing high-quality bounds. Rather than computing expensive bounds at each search node,
we train a neural network offline using high-quality bounds, separately for each PGM, to predict upper bounds on the MPE value from partial assignments. These high-quality bounds are too costly to compute during search, but they provide informative training signals. At test time, the network predicts bounds in milliseconds while retaining much of their quality, enabling fast and effective pruning.

Using neural predictions inside branch-and-bound raises two technical challenges. First, unlike classical bounds, which are admissible by construction, neural predictions may fail to provide valid upper bounds. Second, classical bounds such as MBE and JGLP are monotone along search paths: adding evidence cannot increase the upper bound. Neural predictors do not inherently satisfy this property and may therefore produce inconsistent bounds along AOBB search paths.
We address validity through statistical calibration, which converts neural predictions into upper bounds with high-probability validity. 
We address monotonicity through an encoding scheme that reduces evidence monotonicity to value monotonicity. We instantiate this scheme using a \emph{Scalable Monotonic Neural Network} (SMNN)~\citep{kim2024scalable}, which enforces monotonicity by construction.

Building on these components, we introduce a framework for training neural networks to predict tight upper bounds for MPE inference from partial assignments and integrate the resulting bounds into a branch-and-bound solver. The learned bounds are monotone by construction and statistically calibrated for high-probability validity. To the best of our knowledge, this is the first branch-and-bound solver to combine these properties.
Classical bounds such as MBE and JGLP are deterministically valid and monotone non-increasing in evidence; a learned predictor has neither property by default. To the best of our knowledge, L2B is the first learned bound to recover both---monotonicity by construction, and, in place of deterministic validity, a finite-sample high-probability guarantee---and to be deployed inside a branch-and-bound solver for MPE. Our key contributions are:

\begin{itemize}[itemsep=0pt, parsep=2pt, topsep=0pt, partopsep=0pt,leftmargin=*]
\item A \textbf{monotone-by-construction architecture} that enforces evidence-monotone bounds together with a statistical calibration that converts raw predictions into high-probability-valid bounds.
\item A \textbf{comprehensive empirical evaluation} across PGM families: we show calibrated bounds can be 5--10$\times$ tighter than MBE, computed in just milliseconds, and improve anytime MPE solution quality when integrated into branch-and-bound.
\end{itemize}

\section{Related Work}
In this work, we are concerned with the setting in which a PGM has already been fit to data, and we are simply using the model to make (repeated) inference queries using branch-and-bound (BnB) algorithms for exact MPE. BnB methods such as AND/OR search \citep{marinescu,AOBB} exploit problem structure through context-sensitive caching, while weighted CSP solvers like \textsc{Toulbar2} \citep{toulbar2} and systematic search with caching \citep{bacchus2009solving} provide complementary approaches. The efficiency of these methods depends critically on the quality of upper bounds, see \citep{mbe,dechter2003mini,dechter2022reasoning,jglp,mateescu2010join,globerson2007fixing,ihler2012join,liu2011bounding,rahman2021novel}.  However, achieving bounds tight enough for effective pruning requires configurations with prohibitive time and space complexity, the bottleneck that our work addresses by learning to \emph{amortize} expensive bound computations.

Neural networks have demonstrated remarkable success in guiding combinatorial search across diverse domains, e.g., \citep{gasse2019exact,nair2020solving,khalil2022mip, yonetani2021path, shah2020deep, kool2018attention, cappart2023combinatorial,bengio2021machine}.  However, these methods typically provide no formal guarantees on solution quality. So, predictions may be arbitrarily poor on out-of-distribution instances. Our calibration framework directly addresses this gap by providing finite-sample coverage guarantees.

Other approaches have used neural networks to guide existing  classical approaches to perform better approximate inference\citep{cremer2018inference,satorras2021neural,messaoud2020can}. L2B differs fundamentally from these: we produce \emph{admissible} bounds with formal coverage guarantees, preserving the exactness of branch-and-bound while accelerating convergence.

Most directly related to our work, are approaches that train neural networks for directly answering MPE queries, e.g., \citep{arya2025sine, arya2024neural, malhotra2025learning}. %
However, these methods learn to \emph{predict} MPE solutions without formal guarantees on prediction quality. Our approach is fundamentally different: we learn \emph{upper bounds} rather than solutions, apply statistical calibration to transform neural predictions into provably valid bounds, and use branch-and-bound to find the MPE.

\section{Preliminaries}

A probabilistic graphical model (PGM), given by the tuple $(G,F,X,D)$, represents the joint distribution of a collection of random variables that factorizes over a graph $G$ where $F=\{f_1,...,f_m\}$ is a set of factors, i.e., non-negative functions over the
variables $X=\{X_1,...,X_n\}$ with discrete domains $D = \{D_1,\ldots, D_n\}$, that define a factorization of the joint distribution $p(x_1,\ldots,x_n) \propto \prod_{i=1}^m f_i(x)$. \citep{koller,darwiche09}

In this work, we will primarily be concerned with the MPE inference tasks.  MPE seeks to compute the most probable explanation for non-evidence variables, $x_{\overline{E}}$, given evidence variables, $x_E$: $\arg\max_{x_{\overline{E}}} p(x_{\overline{E}}|x_E).$  

As this task is intractable in general settings, different strategies have been proposed to find an approximate MPE solution, e.g., max-product belief propagation~\citep{globersonetal}, mini-bucket elimination~\citep{dechter2003mini}, and their myriad variants.  Many of these approximation schemes can be viewed as bound minimization schemes:  they iteratively reduce an upper bound on the MPE assignment, but they are not guaranteed to yield tight bounds even at convergence.  The bounds produced by max-product and mini-bucket can be tightened by considering larger groupings of variables at the expense of increased computational complexity, e.g., the cost of computing the bounds scales exponentially in the size of the largest grouping~\citep{mateescu2010join,ihler2012join}.  

Even if the above bounds are quite loose, the upper bounds from these approximation schemes can still be used as heuristics to guide different types of branch and bound search strategies. Branch-and-Bound (BnB) systematically explores the space of partial assignments, starting from evidence at the root of the search tree \citep{AOBB, marinescu}. Branches are generated by selecting a value for an unassigned variable, each choice yields a different branch, and exploration proceeds best-first, prioritizing branches with higher upper bounds. At each node, an upper bound on the optimal completion is computed using a heuristic such as MBE. A branch is explored recursively until it corresponds to a complete assignment, yielding a lower bound (feasible solution). Branches whose upper bounds fall below the best known lower bound are safely pruned.

BnB's efficiency depends critically on bound tightness: tighter bounds enable more aggressive pruning, reducing the search space exponentially. However, achieving tight bounds, e.g., MBE with large $i$-bound, incurs prohibitive per-node cost \citep{mbe,dechter2022reasoning}.

\begin{figure*}[t]
\centering
\begin{minipage}[t]{0.48\textwidth}
\begin{algorithm}[H]
\caption{Data Collection for L2B}
\label{alg:data_collection}
\small
\begin{algorithmic}[1]
\REQUIRE PGM $\mathcal{M}$, size range $[s_{\min}, s_{\max}]$, burn-in $B$, dataset size $N$, oracle $i$-bound $i^*$
\ENSURE Dataset $\mathcal{D} = \{(x_{E_j}, h^*(x_{E_j}))\}_{j=1}^N$

\STATE $\mathcal{D} \leftarrow \emptyset$
\WHILE{$|\mathcal{D}| < N$}
    \STATE $x \leftarrow \text{GibbsSample}(\mathcal{M}, B)$
    \STATE $s \leftarrow \text{ExpDist}(s_{\min}, s_{\max})$
    \STATE $E \leftarrow$ random size-$s$ subset of variables; $x_E \leftarrow x|_E$
    \STATE $h^*(x_E) \leftarrow \text{MBE}(\mathcal{M}, x_E, i\text{-bound}=i^*)$
    \STATE $\mathcal{D} \leftarrow \mathcal{D} \cup \{(x_E, h^*(x_E))\}$
\ENDWHILE
\STATE \textbf{return} $\mathcal{D}$
\end{algorithmic}
\end{algorithm}
\end{minipage}
\hfill
\begin{minipage}[t]{0.48\textwidth}
\begin{algorithm}[H]
\caption{Stratified Monotonic Conformal Calibration}
\label{alg:calibration}
\small
\begin{algorithmic}[1]
\REQUIRE Trained model $f_\theta$, calibration set $\mathcal{D}_{\text{cal}}$, confidence $\gamma$, bins $K$
\ENSURE Per-bin quantiles $\{\hat{q}^b_{\gamma}\}_{b=1}^K$
\STATE Compute residuals: $s_i = h^*(x_{E_i}) - f_\theta(x_{E_i})$ for all $(x_{E_i}, h^*_i) \in \mathcal{D}_{\text{cal}}$
\STATE Partition into bins by $|E_i|$: $\mathcal{S}^b = \{s_i : B(x_{E_i}) = b\}$
\FOR{$b = 1$ to $K$}

    \STATE $\hat{q}^b_{\gamma} \leftarrow$ $\gamma(1 + 1/|\mathcal{S}^b|)$-quantile of $\mathcal{S}^b$
    \IF{$b > 1$}
        \STATE $\hat{q}^b_{\gamma} \leftarrow \min\left(\hat{q}^{b-1}_{\gamma},\ \hat{q}^b_{\gamma}\right)$
    \ENDIF
\ENDFOR
\STATE \textbf{return} $\{\hat{q}^b_{\gamma}\}_{b=1}^K$
\end{algorithmic}
\end{algorithm}
\end{minipage}
\end{figure*}

\section{Learning to Bound}
\label{sec:method}

We propose Learning to Bound (L2B), a neural network-based method for predicting tight upper bounds for MPE inference with formal statistical guarantees. L2B targets a core bottleneck in branch-and-bound search: bounds must be tight to enable aggressive pruning, valid to preserve correctness, and fast to compute since bound evaluations are performed repeatedly during search. Our proposed model learns to approximate high-quality bounds that are expensive to compute with traditional methods using offline supervision, and then applies statistical calibration to convert raw predictions into provably valid upper bounds. %

Given a probabilistic graphical model $\mathcal{M} = (G, F, X, D)$ and evidence $x_E$, which is a partial assignment to variables $E \subseteq X$, let $h^*(x_E) = \max_{x_{\overline{E}}} p(x_{\overline{E}} \mid x_E)$ denote the log-likelihood score of the most likely full assignment consistent with $x_E$. 
In branch-and-bound search, upper bounds on $h^*(x_E)$ enable pruning: when a node’s upper bound falls below the best solution found so far, the algorithm discards the entire subtree. We seek a function $g: \mathcal{X}_E \rightarrow \mathbb{R}$ satisfying the coverage property $\mathbb{P}[g(x_E) \geq h^*(x_E)] \geq \gamma$, for user-specified confidence level $\gamma$. This involves two steps: (1) learning an accurate predictor $f_\theta(x_E) \approx h^*(x_E)$, and (2) calibrating predictions to provide coverage guarantees.

We construct a supervised dataset of partial assignments paired with their optimal log-likelihood scores to train the L2B model. Algorithm~\ref{alg:data_collection} details the data collection procedure. Given a PGM $\mathcal{M}$, we repeatedly sample full assignments $x \sim P_{\mathcal{M}}$ using Gibbs sampling with a burn-in period $B$ to bias samples toward high-probability regions. For each sample, we draw an evidence size $s$ from an exponential distribution biased toward smaller values, reflecting the fact that branch-and-bound search encounters a larger number of partial assignments near the root of the search tree. We then sample a random subset $E \subseteq X$ with $|E| = s$ and extract the corresponding evidence $x_E$. 

Computing the ground-truth labels, $h^*(x_E)$, is intractable in general. Instead, we use MBE with a high $i$-bound to produce upper bounds on them. This procedure produces high-quality upper bounds that remain valid for AOBB, but it incurs a significant per-query cost, making it impractical for use at test time. By incurring this cost during training, the network amortizes the computation of high-quality bounds across subsequent search queries, enabling millisecond-scale bound evaluation at inference time.

\subsection{Monotonicity in Evidence}
\label{sec:smnn}

The optimal log-likelihood score $h^*(x_E)$ in the MPE settings is \emph{monotone non-increasing in evidence}, i.e., for $E \subseteq E'$ with $x_{E'}|_E = x_E$, $h^*(x_{E'}) \leq h^*(x_E)$. This is a desirable property of a predictor as seeing more evidence can only reduce the upper bound.  As such, we require that our predictors are also monotone non-increasing in evidence, mirroring the behavior of MBE/JGLP and preserving correctness along B\&B search paths.%

While standard Monotonic Neural Networks (MNNs) can guarantee monotonicity with respect to input values ($\mathbf{z} \leq \mathbf{z}' \implies f(\mathbf{z}) \leq f(\mathbf{z}')$), achieving monotonicity with respect to the \textbf{presence of evidence} requires a modified strategy: we introduce an input encoding technique that reduces evidence-monotonicity to value-monotonicity. Our approach relies on the following structural principle.
\begin{lemma}
\label{lemma:monotonicity}
Let $\phi(X)$ be an encoding of a discrete variable $X$ taking values in the set $D$, and let $\mathbf{e}$ represent the encoding of the ``unassigned'' state. If $\phi$ satisfies,
\begin{enumerate}[leftmargin=*, itemsep=0pt, label=(\roman*)]
    \item \textbf{Strict Dominance of Evidence:} $\phi(j) > \mathbf{e}$ (element-wise) for all $j \in D$.
    \item \textbf{Value Incomparability:} $\phi(i) \not\leq \phi(j)$ (element-wise) for all distinct $i, j \in D$.
\end{enumerate}
Then, a value-monotonic neural network operating on the concatenated input $\mathbf{z} = [\phi(X_1), \ldots, \phi(X_n)]$ is guaranteed to be monotone non-decreasing with respect to the inclusion of evidence and the encoding does not create a spurious monotonic relationship between different values of $X$. 
\end{lemma}

A natural family of encodings that satisfy these properties is the \textbf{$\mathbf{k}$-hot} scheme, where each value of $X_i$ is mapped to a unique vector of length $M$ containing exactly $k$ ones, and the unassigned sentinel $\mathbf{e}$ is the all-zeros vector. Property (i) is satisfied as any vector with $k \geq 1$ ones strictly dominates the zero vector, and property (ii) follows from the fact that no $k$-hot vector can be a subset of another distinct $k$-hot vector. This is the encoding we adopt here.

We instantiate L2B using a Scalable Monotonic Neural Network (SMNN) architecture~\citep{kim2024scalable}. Specifically, to achieve the final goal of a monotone \textit{non-increasing} bound, we negate the final output,
\begin{equation}
f_\theta(\mathbf{z}) = -\,W^{+}_L\,\sigma\!\Big(\mathrm{BN}\!\big(W^{+}_{L-1}\,\sigma(\dots W^{+}_1 \mathbf{z})\big)\Big).
\end{equation}
Each hidden layer applies a positive-weight linear map $W^{+} = \exp(W')$ (entrywise), batch normalization~\citep{ioffe2015batch} with $\mathrm{affine}=\mathrm{false}$, and a saturating non-decreasing activation $\sigma$ (e.g., $\tanh$). The $\exp$ reparameterization places output-layer weights at $e^{W'}$, which causes output magnitudes to collapse during training. Standard BatchNorm (with affine parameters) fixes the collapse but introduces a learnable scale parameter that breaks monotonicity. We therefore disable the affine transformation in the batch-normalization layer preceding the output, preserving training stability without compromising the architectural monotonicity guarantee. We train $f_\theta$ by minimizing the mean squared error (MSE) between its predictions and the ground-truth labels.

\subsection{Statistical Calibration}
\label{sec:calibration}

Raw neural predictions may underestimate $h^*(x_E)$, violating the upper bound requirement and potentially causing branch-and-bound to prune optimal solutions. Statistical calibration adds a correction term that provides formal coverage guarantees. We consider three approaches.

Gaussian calibration assumes prediction errors are normally distributed: $g_{\text{Gauss}}(x_E) = f_\theta(x_E) + z_{\gamma} \cdot \hat{\sigma}$, where $\hat{\sigma}$ is the root mean squared error on the calibration set and $z_{\gamma}$ is the $\gamma$-quantile of the standard normal distribution. This approach is simple but relies on distributional assumptions that may not hold. Conformal prediction \citep{vovk2005algorithmic} provides distribution-free guarantees: $g_{\text{conf}}(x_E) = f_\theta(x_E) + \hat{q}_{\gamma}$, where $\hat{q}_{\gamma}$ is the $\gamma(1 + 1/n_{\text{cal}})$-quantile of calibration residuals $s_i = h^*(x_{E_i}) - f_\theta(x_{E_i})$. Under exchangeability of calibration and test data, this provides finite-sample coverage $\mathbb{P}[g(x_E) \geq h^*(x_E)] \geq \gamma$ without distributional assumptions.  We adopt this approach here.

We observe that prediction difficulty varies systematically with the amount of evidence, $|E|$. Partial assignments near the root of the search tree exhibit higher uncertainty, while more complete assignments closer to the leaves are more predictable. We stratify by $|E|$ into $K$ bins and compute bin-specific quantiles,
\begin{equation}
g_{\text{strat}}(x_E) = f_\theta(x_E) + \hat{q}_{\gamma}^{B(x_E)},
\end{equation}
where $B(x_E) \in \{1, \ldots, K\}$ is the bin index for assignment size $|E|$. This provides tighter bounds for easier instances while maintaining per-stratum coverage: $\mathbb{P}[g(x_E) \geq h^*(x_E) \mid B(x_E) = b] \geq \gamma$ for all $b$. However, independently computed per-bin quantiles $\hat{q}^b_\gamma$ can violate path-monotonicity when search crosses a bin boundary. To preserve the path-monotonicity guarantee, we enforce a left-to-right monotone cap on the quantile schedule: $\hat{q}^b_{\gamma} \leftarrow \min(\hat{q}^{b-1}_{\gamma}, \hat{q}^b_{\gamma})$ for $b > 1$. This ensures that the calibrated bound $g(x_E)$ remains monotone non-increasing along all search paths. Algorithm~\ref{alg:calibration} summarizes the calibration procedure.

At inference time, L2B integrates with AND/OR branch-and-bound (AOBB) solvers as a drop-in replacement for classical bounding methods, replacing MBE/JGLP and reducing per-node overhead from seconds to sub-milliseconds. For each node with evidence $x_E$, the solver queries the neural network to obtain $\hat{h} = f_\theta(\text{encode}(x_E))$ and adds the calibrated correction, $g = \hat{h} + \hat{q}_{\gamma}^{B(x_E)}$. Nodes with bounds below the current best solution are pruned. The complete integration procedure is given in Appendix~\ref{app:algorithm}.

 The calibrated bound $g$ satisfies the per-node coverage $\mathbb{P}[g \geq h^*] \geq \gamma$, but a single node whose bound underestimates $h^*$ can prune the optimal subtree. Without pruning, exactness of AOBB is preserved by construction and the only effect is on anytime solution quality. When using pruning, $\gamma$ trades off between violation probability and speed (higher $\gamma$ produces looser bounds but stronger guarantees while lower $\gamma$ produces tighter bounds that may prune optimal regions).

\section{Experiments}
\label{sec:experiments}

We evaluated L2B along two complementary dimensions. First, we assessed whether its calibrated neural predictors provide tighter upper bounds than classical bounding methods at substantially lower per-query cost. Second, we measured whether these bounds improve anytime MPE solving when integrated into AND/OR branch-and-bound (AOBB).

We evaluated L2B on 30 probabilistic graphical models from the UAI inference competitions \citep{gogate2014uai,UAICompetition2022}. The benchmark suite contains five model families with induced widths between 10 and 100: Bayesian networks, grids, segmentation models, random uniform models, and structured prediction models. Table~\ref{tab:dataset_stats} in the appendix reports detailed information for all instances.

For each benchmark PGM, we generated 20,000 training examples using Algorithm~\ref{alg:data_collection}.
We obtained partial assignments via Gibbs sampling with 1,000 burn-in steps and an exponential size distribution (bias 2.0), spanning evidence ratios 0.1--0.9. We compute ground-truth labels using DAOOPT~\citep{AOBB} with $i$-bound 10, requiring 0.5--300 seconds per query. For computationally intensive multi-domain instances (Object Detection, Protein Folding), we used evidence ratio 0.5--0.9 and $i$-bound 6 to ensure tractable label generation; all baseline comparisons use $i$-bound 4. We split the data into training (50\%), validation (10\%), calibration (20\%), and test (20\%). We held out the calibration set from training and used it only to estimate calibration parameters.%

We trained separate neural networks for each PGM using the architecture in Section~\ref{sec:smnn}. We used AdamW~\citep{loshchilov2017decoupled} with learning rate $10^{-3}$, weight decay $10^{-4}$, and a cosine annealing schedule over 50 epochs, batch size 256, and early stopping based on validation loss (patience 5). Because target values $h^*(x)$ span wide ranges (e.g., $-500$ to $+5000$ in log-probability space), we standardized targets to zero mean and unit variance during training and store the parameters for denormalization at inference. We implemented all models in PyTorch~\citep{paszke2019pytorch} and train them on a single NVIDIA V100 GPU, completing training in under 10 minutes per PGM. After training, we computed calibration parameters on the held-out calibration set using $K=6$ bins for stratified conformal prediction at confidence levels $\gamma \in {0.95, 0.99, 0.999}$.

\subsection{Upper Bound Quality}
\label{sec:bound_quality}
\begin{table*}[!tbp]
  \centering
  \scriptsize
  \setlength{\tabcolsep}{3pt}

  \caption{Calibrated upper-bound quality on representative PGMs at $i$-bound${=}4$, with stratified-conformal calibration, $\gamma$, on test data. \textbf{MPE$^{\star}$}: true MPE log-likelihood value. \textbf{Relative Gap}: $|\text{bound}-\text{true}|/|\text{true}|$. \textbf{Cov}: fraction of queries with $\text{bound}\geq\text{true MPE}$.}

  \label{tab:smnn_vs_ufoh_3gamma_locked}
  \resizebox{\textwidth}{!}{%
  \begin{tabular}{l r | c | cc cc | cc cc}
  \toprule
    \multirow{3}{*}{Dataset} & \multirow{3}{*}{MPE$^{\star}$} & \multirow{3}{*}{JGLP} & \multicolumn{4}{c|}{$\gamma=0.95$} & \multicolumn{4}{c}{$\gamma=0.999$} \\
    & & & \multicolumn{2}{c}{SMNN (Ours)} & \multicolumn{2}{c|}{UMLP (Ours)} & \multicolumn{2}{c}{SMNN (Ours)} & \multicolumn{2}{c}{UMLP (Ours)} \\
    & & Relative Gap & Cov & Relative Gap & Cov & Relative Gap & Cov & Relative Gap & Cov & Relative Gap \\
    \midrule
    BN\_9 & -23.81* & 0.3649$\pm$0.1745 & 0.97 & 0.0683$\pm$0.0351 & 0.99 & 0.0696$\pm$0.0357 & 1.00 & 0.1285$\pm$0.0396 & 1.00 & 0.1331$\pm$0.0414 \\
    BN\_12 & -19.71* & 0.3438$\pm$0.1896 & 0.97 & 0.1113$\pm$0.0615 & 0.93 & 0.0727$\pm$0.0365 & 0.99 & 0.1989$\pm$0.0858 & 1.00 & 0.1447$\pm$0.0423 \\
    BN\_14 & -26.51* & 0.3411$\pm$0.1954 & 0.92 & 0.0656$\pm$0.0385 & 0.95 & 0.0667$\pm$0.0365 & 0.99 & 0.1165$\pm$0.0469 & 1.00 & 0.1204$\pm$0.0419 \\
    BN\_30 & -14.66 & 0.3017$\pm$0.1558 & 1.00 & 0.0143$\pm$0.0137 & 0.95 & 0.0187$\pm$0.0104 & 1.00 & 0.0897$\pm$0.0142 & 1.00 & 0.0846$\pm$0.0206 \\
    BN\_32 & -16.58 & 0.3379$\pm$0.1860 & 1.00 & 0.0163$\pm$0.0115 & 1.00 & 0.0228$\pm$0.0083 & 1.00 & 0.0926$\pm$0.0230 & 1.00 & 0.0889$\pm$0.0238 \\
    BN\_70 & -86.65 & 0.4475$\pm$0.2403 & 1.00 & 0.0239$\pm$0.0115 & 1.00 & 0.0262$\pm$0.0108 & 1.00 & 0.0534$\pm$0.0154 & 1.00 & 0.0560$\pm$0.0138 \\
    \midrule
    grid10x10.f5.wrap & 159.91* & 0.0248$\pm$0.0156 & 1.00 & 0.0140$\pm$0.0105 & 1.00 & 0.0268$\pm$0.0107 & 1.00 & 0.0454$\pm$0.0109 & 1.00 & 0.0692$\pm$0.0112 \\
    grid20x20.f5.wrap & 617.89* & 0.0143$\pm$0.0113 & 1.00 & 0.0066$\pm$0.0063 & 1.00 & 0.0212$\pm$0.0095 & 1.00 & 0.0279$\pm$0.0107 & 1.00 & 0.0432$\pm$0.0109 \\
    grid20x20.f10 & 1230.40* & 0.0111$\pm$0.0094 & 1.00 & 0.0064$\pm$0.0072 & 0.97 & 0.0171$\pm$0.0102 & 1.00 & 0.0413$\pm$0.0101 & 1.00 & 0.0412$\pm$0.0158 \\
    grid40x40.f2 & 1071.08 & 0.0064$\pm$0.0036 & 1.00 & 0.0029$\pm$0.0019 & 1.00 & 0.0131$\pm$0.0079 & 1.00 & 0.0207$\pm$0.0078 & 1.00 & 0.0259$\pm$0.0084 \\
    grid40x40.f10.wrap & 5308.25 & 0.0092$\pm$0.0024 & 0.33 & 0.0025$\pm$0.0018 & 0.95 & 0.0068$\pm$0.0049 & 1.00 & 0.0108$\pm$0.0028 & 1.00 & 0.0197$\pm$0.0051 \\
    grid80x80.f10 & 20140.10 & 0.0036$\pm$0.0016 & 0.17 & 0.0020$\pm$0.0016 & 0.86 & 0.0073$\pm$0.0074 & 1.00 & 0.0067$\pm$0.0046 & 1.00 & 0.0176$\pm$0.0081 \\
    grid80x80.f5.wrap & 9168.14 & 0.0084$\pm$0.0026 & 1.00 & 0.0050$\pm$0.0010 & 1.00 & 0.0156$\pm$0.0079 & 1.00 & 0.0143$\pm$0.0050 & 1.00 & 0.0262$\pm$0.0116 \\
    \midrule
    Segmentation\_12 & -146.96* & 0.5361$\pm$0.1969 & 0.97 & 0.1199$\pm$0.0453 & 0.97 & 0.0621$\pm$0.0293 & 1.00 & 0.1936$\pm$0.0523 & 1.00 & 0.1029$\pm$0.0339 \\
    Segmentation\_14 & -143.57* & 0.5134$\pm$0.2128 & 0.99 & 0.1356$\pm$0.0477 & 0.98 & 0.0628$\pm$0.0269 & 1.00 & 0.1926$\pm$0.0400 & 1.00 & 0.1064$\pm$0.0360 \\
    Segmentation\_18 & -149.81* & 0.5369$\pm$0.1844 & 0.98 & 0.1838$\pm$0.0645 & 0.98 & 0.0646$\pm$0.0345 & 1.00 & 0.2565$\pm$0.0439 & 1.00 & 0.1036$\pm$0.0395 \\
    \midrule
    rus\_100\_200\_2\_1 & 2776.12 & 0.3779$\pm$0.1738 & 0.99 & 0.0556$\pm$0.0341 & 0.96 & 0.0142$\pm$0.0109 & 1.00 & 0.0704$\pm$0.0364 & 1.00 & 0.0228$\pm$0.0139 \\
    rus\_100\_200\_3\_1 & 2726.97 & 0.6128$\pm$0.1284 & 0.98 & 0.1995$\pm$0.1393 & 0.98 & 0.0127$\pm$0.0072 & 1.00 & 0.2324$\pm$0.1377 & 1.00 & 0.0222$\pm$0.0109 \\
    rus\_100\_200\_3\_3 & 2745.43 & 0.4392$\pm$0.2550 & 0.98 & 0.0805$\pm$0.0583 & 0.98 & 0.0164$\pm$0.0119 & 1.00 & 0.1089$\pm$0.0574 & 1.00 & 0.0292$\pm$0.0148 \\
    rus\_100\_200\_4\_3 & 2740.65 & 0.4341$\pm$0.2447 & 0.96 & 0.0825$\pm$0.0538 & 1.00 & 0.0162$\pm$0.0114 & 1.00 & 0.1066$\pm$0.0563 & 1.00 & 0.0343$\pm$0.0191 \\
    rus\_100\_200\_5\_3 & 2739.05 & 0.5134$\pm$0.2328 & 0.98 & 0.0835$\pm$0.0633 & 0.98 & 0.0156$\pm$0.0105 & 1.00 & 0.1070$\pm$0.0706 & 1.00 & 0.0348$\pm$0.0221 \\
    rus\_100\_200\_6\_1 & 2741.41 & 0.4945$\pm$0.2470 & 0.98 & 0.0776$\pm$0.0514 & 0.94 & 0.0137$\pm$0.0101 & 1.00 & 0.1024$\pm$0.0561 & 1.00 & 0.0280$\pm$0.0189 \\
    rus\_100\_200\_7\_1 & 2739.09 & 0.4508$\pm$0.2410 & 0.95 & 0.0925$\pm$0.0667 & 0.98 & 0.0170$\pm$0.0122 & 1.00 & 0.1162$\pm$0.0646 & 1.00 & 0.0322$\pm$0.0198 \\
    \midrule
    wcsp\_11 & -463.08* & 0.9940$\pm$0.0052 & 0.99 & 0.2035$\pm$0.1260 & 0.97 & 0.1721$\pm$0.1047 & 1.00 & 0.3551$\pm$0.1633 & 1.00 & 0.3481$\pm$0.1421 \\
    wcsp\_12 & -562.67 & 0.9955$\pm$0.0048 & 0.99 & 0.1890$\pm$0.1454 & 0.97 & 0.1283$\pm$0.0826 & 1.00 & 0.3271$\pm$0.1766 & 1.00 & 0.2517$\pm$0.1356 \\
    wcsp\_15 & -328.00 & 0.6536$\pm$0.1156 & 0.97 & 0.1933$\pm$0.1026 & 1.00 & 0.1236$\pm$0.0440 & 1.00 & 0.2945$\pm$0.1063 & 1.00 & 0.2029$\pm$0.0476 \\
    \midrule
    ObjectDetection\_11 & 1999.86 & 0.7080$\pm$0.5687 & 0.81 & 0.2650$\pm$0.2800 & 0.75 & 0.2143$\pm$0.2489 & 1.00 & 0.5001$\pm$0.4505 & 1.00 & 0.4099$\pm$0.3996 \\
    ObjectDetection\_18 & 3313.81 & 0.5547$\pm$0.1644 & 0.75 & 0.1156$\pm$0.1126 & 0.92 & 0.1586$\pm$0.0621 & 1.00 & 0.3699$\pm$0.1213 & 1.00 & 0.2629$\pm$0.0988 \\
    \midrule
    ProteinFolding\_19 & 1557.28 & 0.5385$\pm$0.1350 & 1.00 & 0.0651$\pm$0.0432 & 1.00 & 0.1245$\pm$0.0292 & 1.00 & 0.2844$\pm$0.0601 & 1.00 & 0.2776$\pm$0.0372 \\
    ProteinFolding\_23 & 3165.49 & 0.4682$\pm$0.1535 & 0.94 & 0.1687$\pm$0.1564 & 0.94 & 0.1409$\pm$0.1010 & 1.00 & 0.2818$\pm$0.1579 & 1.00 & 0.2883$\pm$0.0791 \\
  \bottomrule
  \end{tabular}%
  }
\end{table*}

Table~\ref{tab:smnn_vs_ufoh_3gamma_locked} summarizes our main results against JGLP on held-out test queries; the full comparison, which additionally includes the MB column and the intermediate $\gamma=0.99$ level, is deferred to Table~\ref{tab:smnn_vs_ufoh_3gamma_locked_2deterministic} in the appendix.

Neural bounds with stratified conformal calibration achieve substantially tighter gaps than classical methods across all network families while maintaining target coverage (MB columns referenced below appear in Table~\ref{tab:smnn_vs_ufoh_3gamma_locked_2deterministic}). On Bayesian Networks, neural gaps of 0.02--0.08 compare to 0.19--0.45 for MB, a 5--10$\times$ improvement. Coverage exceeds target at both $\gamma = 0.95$ (empirical 0.96--1.00) and $\gamma = 0.999$ (empirical 1.00). On Random Uniform instances, these challenging high-treewidth benchmarks show neural gaps of 0.02--0.03 compared to 0.15--0.29 for MB, a 5--14$\times$ improvement. This demonstrates that neural bounds excel precisely where classical methods struggle. On Segmentation, neural gaps of 0.06--0.07 compare to 0.46--0.49 for MB (7$\times$ tighter). Grid Models show neural bounds achieving comparable gaps to MB on regular grid structures; JGLP achieves tighter bounds on grids, but at 10--300$\times$ higher computational cost (seconds vs.\ milliseconds).

The key insight is that neural bounds trained on $i$-bound 10 supervision achieve this quality in milliseconds (0.001--0.4 seconds) versus seconds to minutes for classical methods, see Figure~\ref{fig:pareto} in the appendix.  %
Stratified conformal consistently achieves 10--20\% tighter gaps than Gaussian or standard conformal at the same coverage level by adapting margins to assignment-size-dependent difficulty (see Appendix~\ref{app:calibration_results} for the full comparison).

We also conducted ablation studies on calibration methods. On BN\_70 at $\gamma = 0.999$, raw neural predictions (no calibration) achieve the tightest gap of 0.008 but violate coverage with only 47\% of bounds being valid. Gaussian calibration restores coverage to 1.00 with the tightest calibrated gap of 0.036, but assumes normally distributed errors. Among distribution-free methods, stratified conformal achieves gap 0.050, 10\% tighter than standard conformal (gap 0.056), by adapting margins to assignment difficulty. Performance improves steadily with training data up to 10,000 examples, with diminishing returns beyond 20,000; our default of 20,000 examples balances quality and data generation cost.

Bound-quality results across PGM families are mixed. On Bayesian networks, SMNN and UMLP perform comparably; on grids, SMNN is statistically better outside one standard deviation; on segmentation and the rus family, UMLP achieves tighter bounds; on WCSP, Object Detection, and Protein Folding, the two methods are statistically equivalent. We do not bold winners in Table~\ref{tab:smnn_vs_ufoh_3gamma_locked} because most differences fall within one standard deviation of each other.

We note one expected edge case: \texttt{grid80x80.f10} at $\gamma=0.95$ shows SMNN coverage 0.17. This is consistent with theory, a 95\% target permits up to 5\% violations in expectation, and small per-stratum sample sizes can amplify this, and at $\gamma=0.999$ coverage returns to 1.00, exactly as the calibration framework predicts.

\subsection{Path Monotonicity}
\label{sec:monotonicity_results}
\begin{table}[!htbp]
  \centering
  \small
  \setlength{\tabcolsep}{4pt}
  \caption{Path-monotonicity violations, aggregated by family. For each of the 30 PGMs we sample 100 parent$\to$child evidence pairs (child $=$ parent plus one assigned variable) and count pairs with $g(\text{child}) > g(\text{parent})$, which breaks admissibility of the heuristic along an AOBB path. \textbf{Raw} is the direct network output; \textbf{Calib.} is the deployed bound after monotone-stratified conformal calibration at $\gamma = 0.999$. SMNN is monotone by construction and stays at zero end-to-end; the unconstrained UMLP violates on every family.}
  \label{tab:monotonicity_main}
  \begin{tabular}{l r | rr | rr}
    \toprule
    \multirow{2}{*}{Family} & \multirow{2}{*}{Pairs} & \multicolumn{2}{c|}{UMLP} & \multicolumn{2}{c}{SMNN (Ours)} \\
    & & Raw & Calib. & Raw & Calib. \\
    \midrule
    Bayesian networks & 600  & 203 & 213 & 0 & \textbf{0} \\
    Grids             & 700  & 145 & 150 & 0 & \textbf{0} \\
    Segmentation      & 300  &  65 &  66 & 0 & \textbf{0} \\
    Random uniform    & 700  & 110 & 111 & 0 & \textbf{0} \\
    WCSP              & 300  & 117 & 114 & 0 & \textbf{0} \\
    Object detection  & 200  & 118 &  92 & 0 & \textbf{0} \\
    Protein folding   & 200  &  91 &  77 & 0 & \textbf{0} \\
    \midrule
    \textbf{All}      & \textbf{3000} & \textbf{849} & \textbf{823} & \textbf{0} & \textbf{0} \\
                      &      & \small(28.3\%) & \small(27.4\%) & \small(0\%) & \small(\textbf{0\%}) \\
    \bottomrule
  \end{tabular}
\end{table}

Tightness alone does not make a bound usable inside search. AOBB evaluates the heuristic at every node of a root-to-leaf path, and each child extends its parent by one assigned variable; if the bound can \emph{rise} as evidence accumulates, the heuristic is inconsistent, node ordering is corrupted, and a prune justified at one node may contradict the bound reported by its parent. Section~\ref{sec:smnn} enforces $g(x_E^{\text{child}}) \leq g(x_E^{\text{parent}})$ by construction. Table~\ref{tab:monotonicity_main} tests whether that property survives into the deployed system: for each of the 30 PGMs we sample 100 pairs of parent$\to$ child evidence and count order violations.

The separation is categorical rather than statistical. SMNN records $0/3000$ violations, both raw and after calibration, in every family. UMLP, the same data, the same calibration, an unconstrained architecture violates in $849/3000$ pairs raw ($28.3\%$) and $823/3000$ after calibration ($27.4\%$), with per-family rates from $15.7\%$ on random-uniform instances to $59.0\%$ on Object Detection. Calibration does not repair the property: it is a per-node offset and cannot restore an ordering the network did not produce.

This reframes the mixed bound-quality picture of Section~\ref{sec:bound_quality}. UMLP is often the tighter predictor by an order of magnitude on the random-uniform family, yet it inverts the bound on roughly one search transition in four. The two architectures therefore offer different advantages: UMLP prioritizes bound tightness, whereas SMNN provides a structural guarantee that tighter predictions alone cannot replace. The AOBB results below reveal which property yields greater practical benefit during search.

The guarantee also has to be defended through calibration, which is where it is easiest to lose. The stratified conformal adds a per-bin offset $\hat{q}^b_\gamma$ indexed by the evidence size, and this schedule is naturally \emph{increasing} in $b$, since larger evidence sets carry greater residual variance. A parent and child that cross a bin boundary can then have their correct raw ordering reversed by the calibrator alone. This is not hypothetical: the unmodified wrapper produces $63/3000$ calibrated violations on a model with $0/3000$ raw violations (Table~\ref{tab:monotonicity_violations_v2} in the appendix). The monotone cap $\hat{q}^b_\gamma \leftarrow \min(\hat{q}^{b-1}_\gamma, \hat{q}^b_\gamma)$ restores a non-increasing schedule and returns the count to zero. It is not free: capping an offset downward can only tighten the bound, so per-bin coverage is spent to buy the ordering, mean worst-bin calibration coverage falls from $1.000$ to $0.674$ at $\gamma=0.999$, though $22/30$ PGMs still hold test coverage above $0.999$ (Appendix~\ref{app:monotone_calib}). We consider the trade favorable for search: a non-monotone bound can prune the optimum \emph{deterministically} through a single inversion at a bin boundary, whereas an under-covered monotone bound does so only probabilistically, at a rate $\gamma$ controls.

\subsection{Branch-and-Bound Integration}
\label{sec:aobb_results}
We integrated L2B into AND/OR branch-and-bound (AOBB)~\citep{AOBB} on binary-domain instances and reported relative MPE improvement $\Delta = (\text{MPE}_\text{neural} - \text{MPE}_\text{baseline}) / |\text{MPE}_\text{baseline}|$ over the MBE-$i4$ baseline at matched wall-clock budgets in Table~\ref{tab:aobb_main} ($\gamma \in \{0.95, 0.999\}$, 10-minute timeout, 3 candidate queries per PGM averaged).

\begin{table}[t]
  \centering
  \scriptsize
  \setlength{\tabcolsep}{2pt}
  \caption{Anytime AOBB MPE improvement over MBE on high-treewidth instances.%
  $\Delta = (\text{NN MPE}-\text{MBE MPE})/|\text{MBE MPE}|$; positive values indicate the NN-augmented search found a strictly better MPE within the time budget. %
  \textbf{Bold} marks $\Delta>10^{-4}$. }
  \label{tab:aobb_main}
  \begin{tabular}{l c l | c c c | c c c}
  \toprule
  \multirow{2}{*}{Instance} & \multirow{2}{*}{$w^*$} & \multirow{2}{*}{Method} & \multicolumn{3}{c|}{$\gamma=0.95$} & \multicolumn{3}{c}{$\gamma=0.999$} \\
  & & & 1\,m & 5\,m & 10\,m & 1\,m & 5\,m & 10\,m \\
  \midrule
  \multirow{2}{*}{rus2\_1} & \multirow{2}{*}{105} & SMNN & +0.0000 & \textbf{+0.0225} & \textbf{+0.0225} & -0.2629 & \textbf{+0.0225} & \textbf{+0.0225} \\
   & & UMLP & -0.1416 & -0.1541 & -0.1399 & -0.3020 & -0.1490 & -0.1399 \\
  \addlinespace[1pt]
  \multirow{2}{*}{rus3\_1} & \multirow{2}{*}{96} & SMNN & \textbf{+0.0013} & \textbf{+0.0013} & \textbf{+0.0013} & \textbf{+0.0017} & \textbf{+0.0017} & \textbf{+0.0017} \\
   & & UMLP & \textbf{+0.0017} & \textbf{+0.0020} & \textbf{+0.0020} & \textbf{+0.0017} & \textbf{+0.0017} & \textbf{+0.0020} \\
  \addlinespace[1pt]
  \multirow{2}{*}{rus3\_3} & \multirow{2}{*}{98} & SMNN & \textbf{+0.0032} & \textbf{+0.0032} & \textbf{+0.0032} & \textbf{+0.0032} & \textbf{+0.0032} & \textbf{+0.0032} \\
   & & UMLP & \textbf{+0.0045} & \textbf{+0.0077} & \textbf{+0.0077} & \textbf{+0.0030} & \textbf{+0.0045} & \textbf{+0.0077} \\
  \addlinespace[1pt]
  \multirow{2}{*}{rus4\_3} & \multirow{2}{*}{100} & SMNN & -1.1081 & \textbf{+0.0567} & \textbf{+0.0567} & -0.0142 & \textbf{+0.0567} & \textbf{+0.0567} \\
   & & UMLP & -0.9273 & -0.3418 & -0.3409 & -0.3726 & -0.3409 & -0.3409 \\
  \addlinespace[1pt]
  \multirow{2}{*}{rus5\_3} & \multirow{2}{*}{100} & SMNN & -0.6138 & \textbf{+0.0074} & \textbf{+0.0074} & -0.2209 & \textbf{+0.0074} & \textbf{+0.0074} \\
   & & UMLP & -0.7270 & -0.3185 & -0.3185 & -0.5288 & -0.3185 & -0.3125 \\
  \addlinespace[1pt]
  \multirow{2}{*}{rus6\_1} & \multirow{2}{*}{98} & SMNN & -0.9690 & \textbf{+0.0500} & \textbf{+0.0500} & -- & \textbf{+0.0500} & \textbf{+0.0500} \\
   & & UMLP & -0.8625 & -0.3581 & -0.3581 & -0.3377 & -0.3581 & -0.3581 \\
  \addlinespace[1pt]
  \multirow{2}{*}{rus7\_1} & \multirow{2}{*}{99} & SMNN & -0.9084 & \textbf{+0.0399} & \textbf{+0.0399} & \textbf{+0.0492} & \textbf{+0.0399} & \textbf{+0.0399} \\
   & & UMLP & -0.4535 & -0.3850 & -0.3850 & -0.8649 & -0.3850 & -0.3850 \\
  \midrule
  \multirow{2}{*}{g80.f5w} & \multirow{2}{*}{89} & SMNN & \textbf{+0.0118} & \textbf{+0.0125} & \textbf{+0.0118} & \textbf{+0.0135} & \textbf{+0.0125} & \textbf{+0.0118} \\
   & & UMLP & \textbf{+0.0135} & \textbf{+0.0125} & \textbf{+0.0118} & -- & \textbf{+0.0111} & \textbf{+0.0111} \\
  \addlinespace[1pt]
  \multirow{2}{*}{g80.f10} & \multirow{2}{*}{66} & SMNN & \textbf{+0.0138} & \textbf{+0.0137} & \textbf{+0.0134} & \textbf{+0.0141} & \textbf{+0.0137} & \textbf{+0.0134} \\
   & & UMLP & \textbf{+0.0147} & \textbf{+0.0145} & \textbf{+0.0139} & -- & \textbf{+0.0137} & \textbf{+0.0134} \\
  \addlinespace[1pt]
  \multirow{2}{*}{g40.f10w} & \multirow{2}{*}{63} & SMNN & \textbf{+0.0157} & \textbf{+0.0156} & \textbf{+0.0155} & \textbf{+0.0157} & \textbf{+0.0156} & \textbf{+0.0155} \\
   & & UMLP & \textbf{+0.0157} & \textbf{+0.0156} & \textbf{+0.0155} & \textbf{+0.0157} & \textbf{+0.0156} & \textbf{+0.0155} \\
  \addlinespace[1pt]
  \multirow{2}{*}{g40.f2} & \multirow{2}{*}{30} & SMNN & \textbf{+0.0189} & \textbf{+0.0185} & \textbf{+0.0184} & \textbf{+0.0189} & \textbf{+0.0185} & \textbf{+0.0184} \\
   & & UMLP & \textbf{+0.0189} & \textbf{+0.0185} & \textbf{+0.0184} & \textbf{+0.0189} & \textbf{+0.0185} & \textbf{+0.0184} \\
  \bottomrule
  \end{tabular}
\end{table}

Although bound-quality results are mixed across PGM families (Section~\ref{sec:bound_quality}), the AOBB integration tells a sharper story. Even on the rus family, where UMLP achieves tighter raw bounds, SMNN-augmented search produces substantially better MPE solutions, with the gap reaching an order of magnitude on multiple instances at 10 minutes. This is what the violation counts of Table~\ref{tab:monotonicity_main} predict. AOBB calls the bound repeatedly along each path, and a monotone-by-construction bound is guaranteed to move in the correct direction as conditioning grows; an unconstrained predictor that inverts on $27\%$ of transitions is occasionally too optimistic at one node and too pessimistic at the next, which is precisely the regime that produces pruning errors. Tightness measured one query at a time does not see this failure mode, which is why UMLP can win Table~\ref{tab:smnn_vs_ufoh_3gamma_locked} and lose Table~\ref{tab:aobb_main}. On high-treewidth instances ($w^*\geq 60$), SMNN achieves an average improvement of 1-6\% over MBE; we view this as substantial given that these are precisely the regime where MBE-$i4$ bounds are loosest and the search space is largest.

Saturated families (BN, WCSP, ObjDet, PF), AOBB results on the remaining families are reported in Tables~\ref{tab:final_aobb_combined_merged_avg_top3}--\ref{tab:final_aobb_combined_lt_avg_top3} in the appendix.

\section{Conclusion}
\label{sec:conclusion}

We introduced Learning to Bound (L2B), a framework for predicting tight upper bounds on MPE values with formal statistical guarantees. By training on high-quality supervision from expensive MBE configurations ($i$-bound 10) and applying conformal calibration, L2B produces bounds that are simultaneously \textbf{tight} (5--10$\times$ lower gap than MBE), \textbf{fast} (milliseconds vs.\ minutes, enabling deployment in search inner loops), and \textbf{valid} with high probability (finite-sample coverage guarantees, with a measured and $\gamma$-controllable violation rate). The key empirical finding is that neural bounds provide the greatest benefit where they are most needed: on \textbf{high-treewidth instances} where classical bounds are loosest. When integrated into AOBB, neural bounds achieve 1--6\% better MPE solutions on high-treewidth Random Uniform and large grid instances ($w^*\geq 60$), which demonstrates that tight bounds enable discovery of high-quality regions more quickly than looser classical bounds, which waste time exploring suboptimal search paths.

\paragraph{Limitations}
\label{sec:limitations}
L2B specializes a predictor to each PGM with fixed structure and parameters, favoring tight instance-specific bounds over cross-model generalization. Conformal calibration provides finite-sample coverage at each search node; although this guarantee does not extend jointly over entire search paths, AOBB remains anytime and preserves the feasibility and objective value of incumbent solutions.

\bibliographystyle{plainnat}
\bibliography{references}

\clearpage
\appendix

\section*{Appendix}

\section{UMLP}
The UMLP predictor $f_\theta$ is a feedforward network with four fully-connected layers:
\begin{equation*}
f_\theta: \mathbb{R}^{n \cdot b} \xrightarrow{\text{FC}_{512}} \mathbb{R}^{512} \xrightarrow{\text{FC}_{512}} \mathbb{R}^{512} \xrightarrow{\text{FC}_{256}} \mathbb{R}^{256} \xrightarrow{\text{FC}_{1}} \mathbb{R}^{1}.
\end{equation*}
Each hidden layer is followed by batch normalization, ReLU activation, and dropout ($p=0.2$).

\section{Neural Architecture and Path Monotonicity}
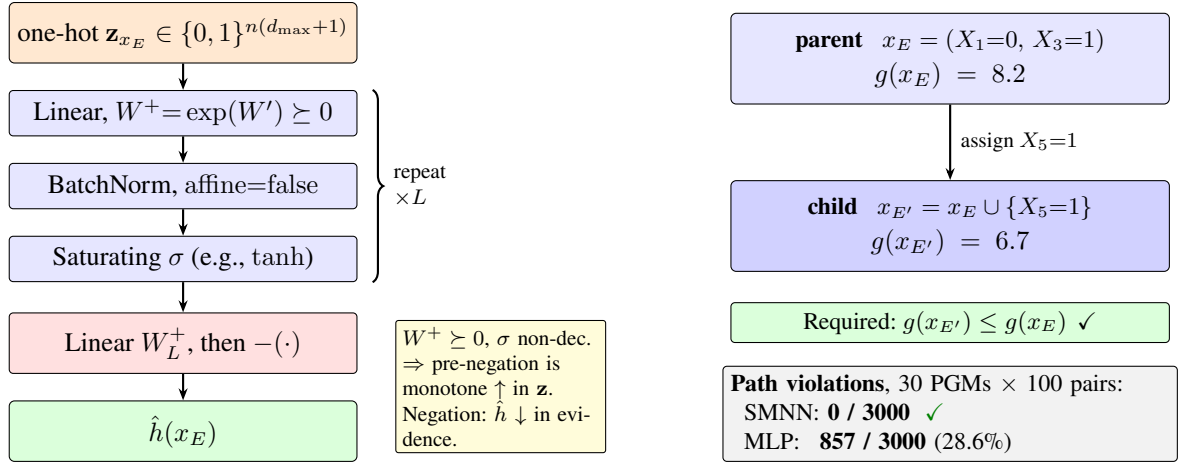
\begin{figure*}[t]
\centering
\begin{subfigure}[t]{0.56\textwidth}
\centering
\begin{tikzpicture}[
  node distance=3.5mm,
  every node/.style={font=\small},
  io/.style={draw, rounded corners=2pt, minimum width=46mm, minimum height=8mm,
             fill=orange!18, align=center, inner sep=2pt},
  blk/.style={draw, rounded corners=1.5pt, minimum width=46mm, minimum height=6mm,
              fill=blue!10, align=center, inner sep=2pt},
  finblk/.style={draw, rounded corners=1.5pt, minimum width=46mm, minimum height=8mm,
                 fill=red!12, align=center, inner sep=2pt},
  outblk/.style={draw, rounded corners=2pt, minimum width=46mm, minimum height=8mm,
                 fill=green!14, align=center, inner sep=2pt},
  arr/.style={-{Stealth[length=1.6mm]}, thick}
]
\node[io] (in) {one-hot $\mathbf{z}_{x_E}\in\{0,1\}^{n(d_{\max}+1)}$};

\node[blk, below=of in] (lin1) {Linear, $W^{+}{=}\exp(W')\succeq 0$};
\node[blk, below=of lin1] (bn1) {BatchNorm, $\mathrm{affine}{=}\mathrm{false}$};
\node[blk, below=of bn1] (sig1) {Saturating $\sigma$ (e.g., $\tanh$)};

\node[finblk, below=4mm of sig1] (out) {Linear $W^{+}_L$, then $-(\cdot)$};
\node[outblk, below=of out] (h) {$\hat h(x_E)$};

\draw[arr] (in)   -- (lin1);
\draw[arr] (lin1) -- (bn1);
\draw[arr] (bn1)  -- (sig1);
\draw[arr] (sig1) -- (out);
\draw[arr] (out)  -- (h);

\draw[decorate, decoration={brace, amplitude=4pt}, thick]
  ($(lin1.north east)+(2mm,0)$) -- ($(sig1.south east)+(2mm,0)$)
  node[midway, right=4pt, font=\scriptsize, align=left, text width=18mm]
  {repeat\\$\times L$};

\node[anchor=north west, font=\scriptsize, align=left, text width=26mm,
      fill=yellow!18, draw, rounded corners=1pt, inner sep=2.5pt]
  at ($(sig1.south east)+(5mm, -5mm)$)
  {$W^{+}\succeq 0$, $\sigma$ non-dec.\\
   $\Rightarrow$ pre-negation is\\
   monotone $\uparrow$ in $\mathbf{z}$.\\
   Negation: $\hat h\downarrow$ in evidence.};
\end{tikzpicture}
\caption{SMNN architecture. Positive weights $W^{+}=\exp(W')$ and a non-decreasing saturating $\sigma$ make the pre-output composition monotone non-decreasing in $\mathbf{z}_{x_E}$; final negation yields $\hat h$ monotone non-increasing in evidence. $\mathrm{BN}(\mathrm{affine}{=}\mathrm{false})$ before the output prevents $e^{W'}$ collapse without re-introducing learnable signs.}
\label{fig:smnn_arch}
\end{subfigure}\hfill
\begin{subfigure}[t]{0.41\textwidth}
\centering
\begin{tikzpicture}[
  node distance=4mm,
  every node/.style={font=\small},
  pn/.style={draw, rounded corners=2pt, minimum width=58mm, minimum height=12mm,
             fill=blue!10, align=center, inner sep=3pt},
  cn/.style={draw, rounded corners=2pt, minimum width=58mm, minimum height=12mm,
             fill=blue!18, align=center, inner sep=3pt},
  arr/.style={-{Stealth[length=1.6mm]}, thick}
]
\node[pn] (parent) {%
  \footnotesize \textbf{parent}\;\; $x_E = (X_1{=}0,\,X_3{=}1)$\\[1pt]
  $g(x_E)\;=\;8.2$};
\node[cn, below=10mm of parent] (child) {%
  \footnotesize \textbf{child}\;\; $x_{E'} = x_E \cup \{X_5{=}1\}$\\[1pt]
  $g(x_{E'})\;=\;6.7$};

\draw[arr] (parent) -- node[right, font=\scriptsize] {assign $X_5{=}1$} (child);

\node[font=\footnotesize, below=4mm of child, fill=green!14, draw,
      rounded corners=1.5pt, inner sep=3pt, align=center, minimum width=58mm] (req)
  {Required: $g(x_{E'})\leq g(x_E)$ \,\checkmark};

\node[font=\footnotesize, below=3mm of req, draw, rounded corners=1.5pt,
      inner sep=3pt, align=left, text width=58mm, fill=gray!10]
  {\textbf{Path violations}, 30 PGMs $\times$ 100 pairs:\\
   \;\;SMNN: \textbf{0 / 3000} \;\textcolor{green!50!black}{$\checkmark$}\\
   \;\;MLP:\;\;\;\textbf{857 / 3000}\,\,(28.6\%)};
\end{tikzpicture}
\caption{Monotonicity along search paths. Conditioning on more evidence ($X_5{=}1$) takes parent $x_E$ to child $x_{E'}\supset x_E$. Classical bounds and SMNN are monotone: $g(x_{E'})\le g(x_E)$. An unconstrained MLP frequently violates this; SMNN exhibits zero violations. (Bound values illustrative.)}
\label{fig:monotonicity_paths}
\end{subfigure}

\caption{(a)~SMNN architecture: positive weights, batch normalization without affine parameters, saturating activation, and output negation give monotonicity in evidence \emph{by construction}. (b)~Empirical effect along AOBB search paths: SMNN exhibits zero violations across 30 benchmark PGMs; an unconstrained MLP violates monotonicity on 28.6\% of parent--child pairs.}
\label{fig:smnn}
\end{figure*}

Figure~\ref{fig:smnn} provides the structural blueprint for the L2B model. Panel (a) details the internal feedforward pipeline: the use of positive-weight reparameterization ($W^+ = \exp(W')$) and non-decreasing saturating activations ($\sigma$) ensures that the pre-negation composition remains monotone increasing with respect to the input encoding $\mathbf{z}_{x_E}$. As established by Lemma~\ref{lemma:monotonicity}, our encoding scheme ensures that the accumulation of evidence corresponds to a bitwise increase in the input vector. When combined with the final output negation, this provides a structural guarantee that the predicted bound $g(x_E)$ is monotone non-increasing as search deepens.

Panel (b) illustrates how this guarantee preserves the integrity of the search process. In classical branch-and-bound, a child node's bound must be at least as tight as its parent's. While unconstrained neural predictors frequently violate this "path-correctness" property—leading to unstable pruning and search artifacts—the SMNN architecture eliminates these violations by design. The figure includes aggregate results from our 30 benchmark PGMs, where the SMNN exhibited zero monotonicity violations across 3,000 parent-child evidence pairs, compared to a 28.3\% violation rate for a standard MLP.

\label{app:neural_arch}

\section{L2B Pipeline}
\label{app:l2b_pipeline}
\begin{figure*}[htbp]
\centering
\includegraphics[width=0.92\textwidth]{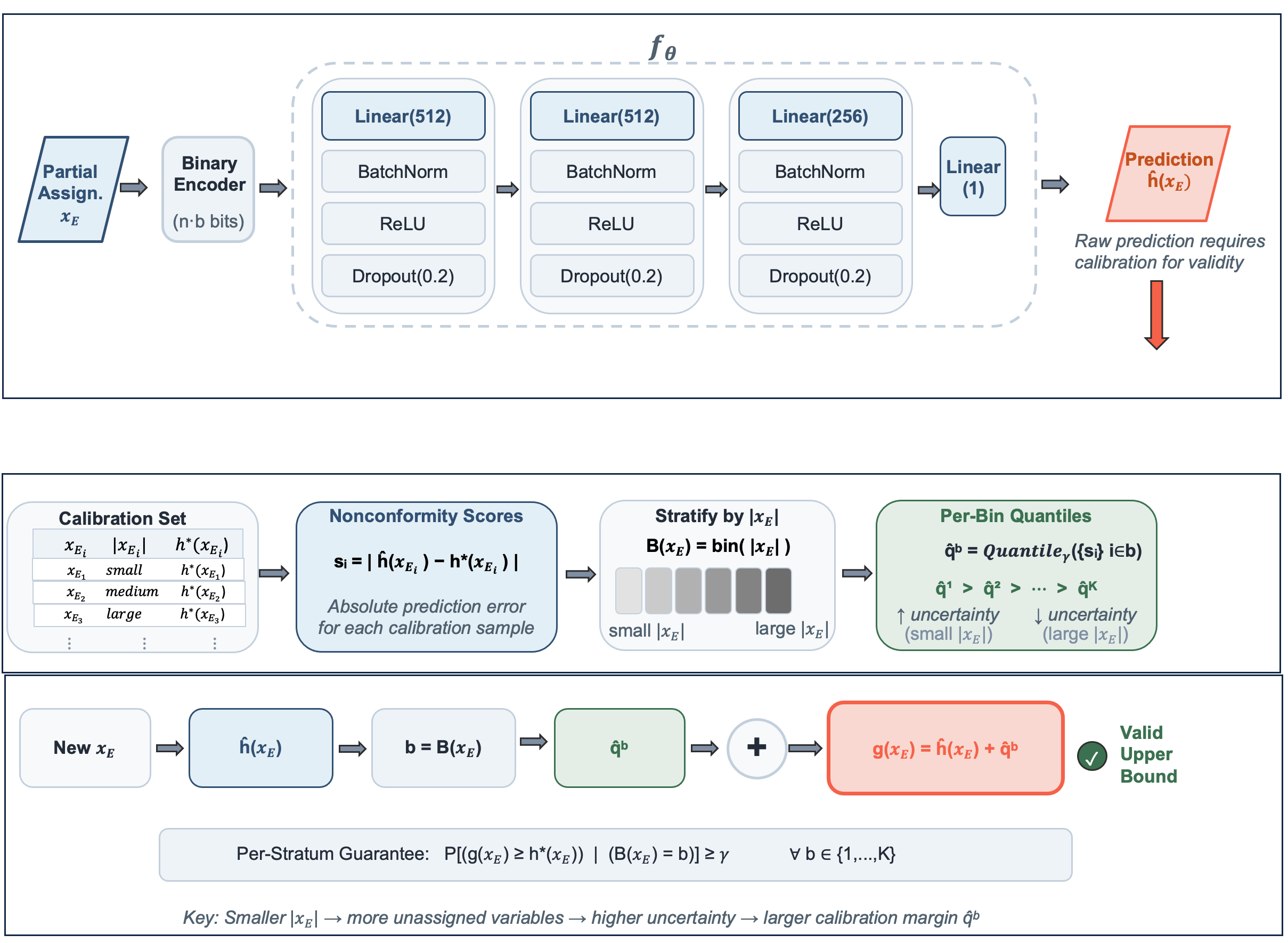}

\caption{L2B pipeline for calibrated neural upper bounds. \textbf{Top:} A binary-encoded partial assignment $x_E$ is processed through MLP $f_\theta$ with three [Linear $\to$ BatchNorm $\to$ ReLU $\to$ Dropout] blocks, producing raw prediction $\hat{h}(x_E)$. \textbf{Middle:} Stratified conformal calibration computes nonconformity scores $s_i = h^*(x_{E_i}) - \hat{h}(x_{E_i})$, stratifies by $|x_E|$ into $K$ bins, and extracts per-bin quantiles $\hat{q}^1 > \cdots > \hat{q}^K$ adapted to prediction difficulty. \textbf{Bottom:} At inference, the calibrated bound $g(x_E) = \hat{h}(x_E) + \hat{q}^{B(x_E)}$ satisfies per-stratum coverage $\mathbb{P}[g(x_E) \geq h^*(x_E) \mid B(x_E) = b] \geq \gamma$ for all $b$.}

\label{fig:pipeline}
\end{figure*}

To further clarify the end-to-end execution of our framework, we provide a detailed visualization of the Learning to Bound (L2B) pipeline in Figure~\ref{fig:pipeline}. This diagram illustrates the three distinct phases of our method: encoding, neural prediction, and statistical calibration.

As depicted in the top panel, the process begins with the transformation of a partial assignment $x_E$ into a high-dimensional vector $\mathbf{z}_{x_E}$ via the encoding scheme defined in Lemma~\ref{lemma:monotonicity}. This vector is processed by the SMNN, which amortizes the high cost of classical MPE bounds by predicting a raw log-likelihood score $\hat{h}(x_E)$ in milliseconds.

The middle and bottom panels visualize our primary mechanism for ensuring admissibility: \textit{stratified conformal calibration}. Because raw neural predictions lack formal guarantees, we compute nonconformity scores $s_i$ on a held-out calibration set. By stratifying these scores into $K$ bins based on the evidence size $|x_E|$, we extract per-bin quantiles $\hat{q}^b_\gamma$ that capture how prediction uncertainty evolves during search. The final calibrated bound $g(x_E)$ is formed by adding the appropriate bin-specific quantile to the neural output. This provides a formal coverage guarantee—ensuring the bound is valid with probability at least $\gamma$—while adapting the margin to be tighter in regions where the model is more confident.

\section{Benchmark Details}
\label{app:benchmarks}
We evaluated on 30 probabilistic graphical models from the UAI inference competitions \citep{gogate2014uai,UAICompetition2022}. The benchmark suite contains five model families with induced widths between 10 and 100:
(1) \textbf{Bayesian Networks}, comprising six real-world instances (BN\_9 through BN\_70) with 90--2315 variables and induced widths $w^* \in [12, 28]$; (2) \textbf{Grid Models}, consisting of seven Ising grids ranging from $10 \times 10$ to $80 \times 80$, with induced widths equal to the grid dimension; (3) \textbf{Segmentation Models}, including three image segmentation instances with 231--235 variables and $w^* \in [23, 25]$; (4) \textbf{Random Uniform Models}, comprising seven synthetic instances with 1094 variables and high induced widths $w^* \in [95, 100]$; and (5) \textbf{WCSP and Structured Prediction Models}, consisting of seven instances, including protein folding tasks, to assess generalization across diverse optimization settings. 

Table~\ref{tab:pgm_details} provides statistics for all benchmark PGMs. We selected instances spanning induced widths from 10 to 80 to evaluate L2B across diverse difficulty levels.

\begin{table}[t]
\centering
\caption{Benchmark PGM statistics.}
\label{tab:pgm_details}
\small
\begin{tabular}{l cccc l}
\toprule
\textbf{Dataset} & $n$ & $m$ & $d_{\max}$ & $w^*$ & \textbf{Source} \\
\midrule
\multicolumn{6}{l}{\textit{Bayesian Networks}} \\
BN\_9, BN\_12, BN\_14 & 90--155 & 90--115 & 2 & 12--18 & UAI \\
BN\_30, BN\_32 & 1156--1444 & 1156--1444 & 2 & 15--22 & UAI \\
BN\_70 & 2315 & 2315 & 36 & - & UAI \\
\midrule
\multicolumn{6}{l}{\textit{Grid Models}} \\
grid10$\times$10 & 100 & 300 & 2 & 10 & Ising \\
grid20$\times$20 & 400 & 1200 & 2 & 20 & Ising \\
grid40$\times$40 & 400 & 1160 & 2 & 40 & Ising \\
grid80$\times$80 & 1600 & 4720 & 2 & 80 & Ising \\
\midrule
\multicolumn{6}{l}{\textit{Segmentation}} \\
Seg\_12, Seg\_14, Seg\_18 & 231--235 & 856--882 & 2 & 23--25 & UAI \\
\midrule
\multicolumn{6}{l}{\textit{Random Uniform (High Treewidth)}} \\
rus\_100\_200\_* & 1094 & 100710 & 2 & 95--100 & UAI \\
\midrule
\multicolumn{6}{l}{\textit{Structured Prediction}} \\
ObjectDetection & 60 & 1830 & 11--16 & 18--20 & UAI \\
ProteinFolding & 60 & 1830 & 11 & 15--17 & UAI \\
\midrule
\multicolumn{6}{l}{\textit{WCSP}} \\
wcsp\_11, wcsp\_12, wcsp\_15 & 125--311 & 736--5732 & 4 & 20--28 & UAI \\
\bottomrule
\end{tabular}
\end{table}

\section{Additional Calibration Results}
\label{app:calibration_results}

The main paper reports stratified conformal results, which achieve the best trade-off between tightness and coverage. Here we present Gaussian and standard conformal results for completeness.

Tables~\ref{tab:ub_smnn_gaussian_locked} and~\ref{tab:ub_smnn_conformal_locked} show results using Gaussian calibration (which assumes normally distributed prediction errors) and standard conformal calibration (using a single global quantile), respectively.

\begin{figure}[t]
\centering
\includegraphics[width=\textwidth]{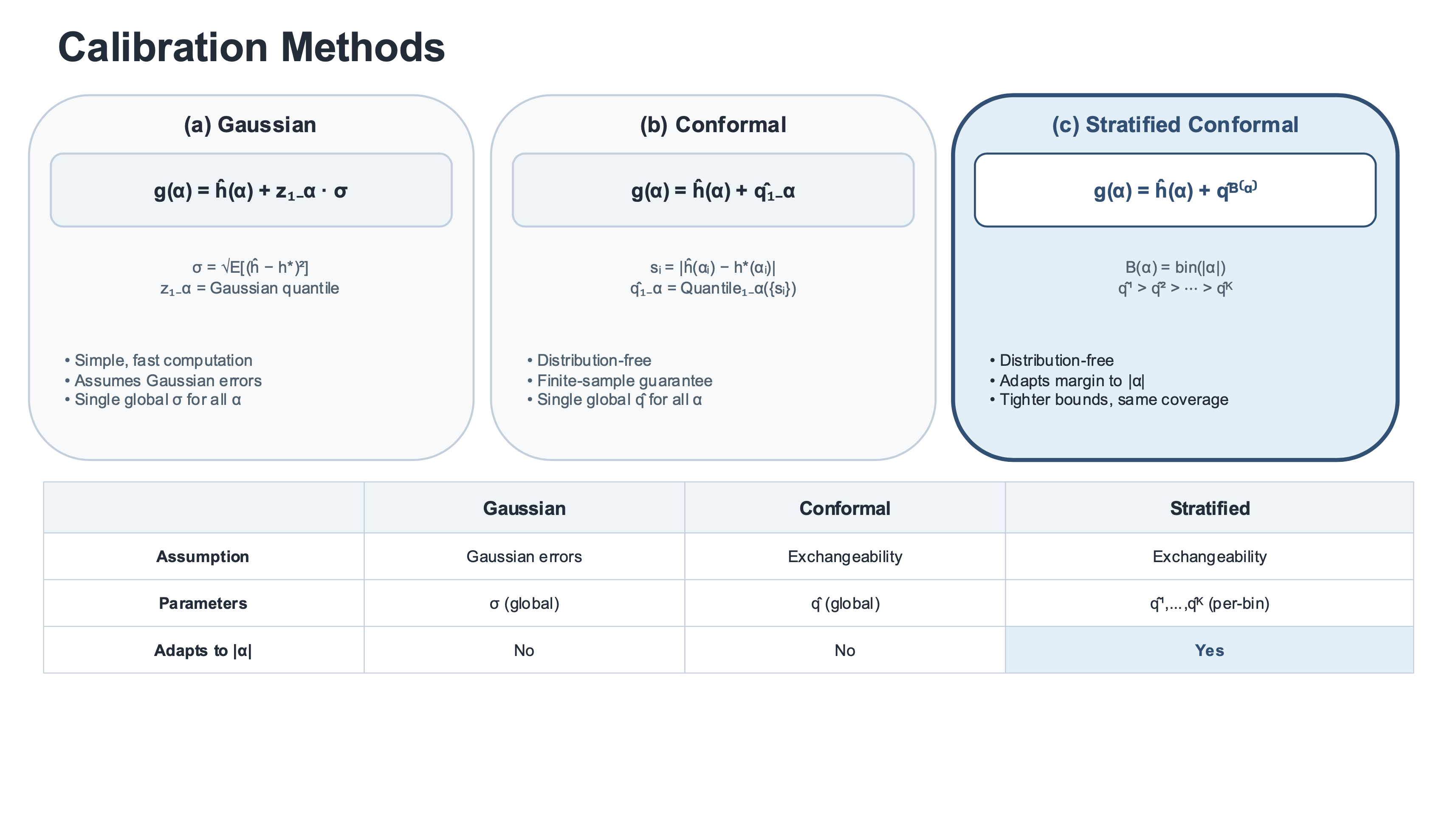}
\caption{Comparison of calibration methods: (a) Gaussian, (b) standard conformal, (c) stratified conformal prediction.}
\label{fig:calibration-comparison}
\end{figure}

Across all calibration methods, neural bounds achieve lower gaps than mini-bucket elimination while maintaining target coverage. Stratified conformal achieves 10--20\% tighter gaps than other methods by adapting calibration margins to assignment-size-dependent difficulty. Figure~\ref{fig:calibration-comparison} illustrates the differences: Gaussian calibration uses a single standard deviation $\sigma$ estimated from calibration errors; standard conformal computes a global quantile $\hat{q}_{\gamma}$ of nonconformity scores; stratified conformal partitions samples by assignment size into $K$ bins and computes per-bin quantiles, enabling tighter bounds when more variables are assigned.

Across all calibration methods, neural bounds achieve lower gaps than mini-bucket elimination while maintaining target coverage. Stratified conformal achieves 10--20\% tighter gaps than other methods by adapting calibration margins to assignment-size-dependent difficulty. Table~\ref{tab:calibration-comparison} summarizes the three approaches.

\begin{table*}[ht]
\centering
\caption{Comparison of calibration methods. Stratified conformal adapts margins to evidence size $|x_E|$, yielding tighter bounds at the same coverage level.}
\label{tab:calibration-comparison}
\small
\begin{tabular}{lccc}
\toprule
& \textbf{Gaussian} & \textbf{Conformal} & \textbf{Stratified Conformal} \\
\midrule
\textbf{Calibrated bound} & $g(x_E) = \hat{h}(x_E) + z_\gamma \cdot \sigma$ & $g(x_E) = \hat{h}(x_E) + \hat{q}_\gamma$ & $g(x_E) = \hat{h}(x_E) + \hat{q}_\gamma^{B(x_E)}$ \\[4pt]
\textbf{Parameters} & $\sigma = \sqrt{\mathbb{E}[(\hat{h} - h^*)^2]}$ & $\hat{q}_\gamma = \text{Quantile}_\gamma(\{s_i\})$ & $\hat{q}_\gamma^b = \text{Quantile}_\gamma(\{s_i\}_{i \in b})$ \\[2pt]
& $z_\gamma$ = Gaussian quantile & $s_i = h^*(x_{E_i}) - \hat{h}(x_{E_i})$ & $B(x_E) = \text{bin}(|x_E|)$ \\[4pt]
\textbf{Assumption} & Gaussian errors & Exchangeability & Exchangeability \\[4pt]
\textbf{Adapts to $|x_E|$} & No & No & \textbf{Yes} \\[4pt]
\textbf{Guarantee} & Asymptotic & Finite-sample & Finite-sample (per-stratum) \\
\bottomrule
\end{tabular}
\end{table*}

\paragraph{High-Treewidth Performance.} The strongest results are on Random Uniform instances ($w^*$=45--52), the highest-treewidth benchmarks. Neural bounds consistently find 5--14\% better solutions, with improvements appearing early and remaining stable. Large grids ($w^*$=40--80) show similar positive gaps. This demonstrates L2B provides greatest benefit where classical methods struggle most.

\paragraph{Calibration Differentiation.} On Bayesian Networks and Segmentation, $\gamma=0.999$ achieves smaller negative gaps than $\gamma=0.95$ (e.g., --0.068 vs --0.089 on BN\_9). This systematic difference validates that our calibration framework produces meaningfully different bounds at different confidence levels.

\paragraph{Convergence Behavior.} On high-treewidth instances, positive gaps appear within the first minute and remain stable, indicating rapid convergence. Both calibration levels converge identically on these problems.

\section{Implementation Details}
\label{app:implementation}

\subsection{Data Generation Details}
\label{app:data_generation}

\paragraph{Gibbs Sampling.} Training data is generated via parallel Gibbs sampling. We first find a feasible starting assignment using topological ordering for Bayesian networks. Each sample uses a burn-in of $2B$ steps (default $B=100$) to ensure mixing. Partial assignment sizes follow an exponential distribution biased toward smaller sizes (bias parameter 2.0), which improves coverage of harder instances with more unassigned variables.

\paragraph{Data Splits.} The full dataset is split as: 50\% training, 10\% validation, 20\% calibration, 20\% test. Validation data is used for early stopping and model selection; calibration data is held out entirely until after training to ensure valid coverage guarantees. Test data uses fixed partial assignment sizes for consistent evaluation.

\paragraph{Evidence Ratio Selection for AOBB.} For AOBB experiments, we select queries with evidence ratio (assigned variables / total variables) below 0.1 to focus on challenging instances with significant search spaces.

\paragraph{Neural Network Training.} Training uses AdamW optimizer with learning rate $10^{-3}$ and weight decay $10^{-4}$, following a cosine annealing schedule over 50 epochs with batch size 256. Gradient clipping with max norm 1.0 prevents unstable updates. Training runs on a single NVIDIA GPU and completes in under 30 minutes per PGM. The checkpoint with lowest validation loss is selected.

\paragraph{Input Encoding.} Partial assignments are encoded as binary vectors where each of $n$ variables uses $b = \lceil \log_2(d_{\max} + 1) \rceil$ bits. A sentinel value ($d_{\max}$) represents unassigned (query) variables. This encoding allows efficient TorchScript export for C++ integration with total input dimension $n \cdot b$.

\paragraph{Target Preprocessing.} During training, target values are standardized (zero mean, unit variance) for numerical stability. At inference, predictions are denormalized using the stored center and scale parameters. This is purely for gradient stability, not uncertainty quantification.

\paragraph{Ground Truth Labels.} Training labels use DAOOPT with $i$-bound 10, providing high-quality bounds that would be prohibitively expensive online. This offline investment enables the network to approach the quality of expensive classical configurations. For small residual problems ($|\mathbf{X} \setminus \alpha| \leq 10$), we use exact brute-force enumeration. Alternative solvers (SCIP, Toulbar2, Merlin) are supported for for data labelling as well in codebase.

\paragraph{Calibration Parameters.} The calibration set comprises 20\% of data, held out after training for valid coverage guarantees. For stratified conformal, we partition assignment sizes into $K=6$ equal-width bins between network-specific minimum and maximum sizes. We evaluate $\alpha \in \{0.50, 0.80, 0.90, 0.95, 0.99, 0.999\}$ during calibration; main results report $\alpha \in \{0.95, 0.999\}$ for brevity. Calibration parameters are serialized to both PyTorch (`.pt`) and text formats for C++ integration.

\paragraph{AOBB Integration.} We extend DAOOPT with neural bound queries via TorchScript export for C++ integration. Calibration files specify the method (Gaussian, conformal, or stratified conformal) and confidence level at runtime. Speculative lookahead batching with depth 8 generates up to 256 partial assignments per batch, amortizing GPU overhead. A two-level cache (L1 direct-mapped with $2^{16}$ entries, L2 hash map) achieves $>$95\% hit rates. All experiments use OR-chain search (\texttt{--or}) with $i$-bound 4 for baseline MBE, auxiliary heuristics disabled (\texttt{--slsX 0 --slsT 0 --lds -1 --match -1 --mplp -1 --jglp -1}), 10-minute budget, and fixed seed 42.
\subsection{Data Generation and Training}

Gibbs sampling initializes from a feasible assignment found via topological ordering for Bayesian networks. Partial assignments are encoded as binary vectors where each of $n$ variables uses $b = \lceil \log_2(d_{\max} + 1) \rceil$ bits, with a sentinel value ($d_{\max}$) indicating unassigned variables. For small residual problems ($|X \setminus E| \leq 10$), we compute exact solutions via brute-force enumeration rather than DAOOPT.

Training uses gradient clipping (max norm 1.0) for stability. For AOBB integration, we select queries with evidence ratio below 0.1 to focus on challenging early-search scenarios. Calibration parameters are serialized to both PyTorch and text formats for C++ solver integration.
\subsection{AOBB Integration}

We extend DAOOPT with neural bound queries via TorchScript export. Speculative lookahead batching with depth 8 generates up to 256 partial assignments per batch, amortizing GPU overhead. A two-level cache (L1 direct-mapped with $2^{16}$ entries, L2 hash map) achieves $>$95\% hit rates. All experiments use OR-chain search with $i$-bound 4 for baseline MBE, auxiliary heuristics disabled, 10-minute budget, and fixed seed 42.

\begin{table*}[ht]
\centering
\small
\caption{PGM details.}
\label{tab:dataset_stats}
\resizebox{0.7\columnwidth}{!}{%
\begin{tabular}{lrrr}
\toprule
\textbf{Dataset} & \textbf{\#Variables } & \textbf{\#Factors} & \textbf{Max domain } \\
\midrule
BN\_9                & 105  & 105  & 2  \\
BN\_12               & 90  & 90  & 2  \\
BN\_14               & 115  & 115  & 2  \\
BN\_30               & 1156  & 1156 & 2  \\
BN\_32               & 1444 & 1444 & 2  \\
BN\_70               & 2315 & 2315 & 36  \\
\midrule
grid10x10.f5.wrap    & 100  & 300  & 2  \\
grid20x20.f5.wrap    & 400  & 1200 & 2  \\
grid20x20.f10        & 400  & 1160 & 2 \\
grid40x40.f2        & 1600 & 4720 & 2 \\
grid40x40.f10.wrap  & 1600 & 4800 & 2 \\
grid80x80.f10       & 6400 & 19040 & 2 \\
grid80x80.f5.wrap   & 6400 & 19200 & 2 \\
\midrule
Segmentation\_12     & 231  & 856 & 2 \\
Segmentation\_14     & 231  & 863 & 2 \\
Segmentation\_18     & 235  & 882 & 2 \\
\midrule
rus\_100\_200\_2\_1  & 1094  & 100710  & 2  \\
rus\_100\_200\_3\_1  & 1094  & 100710  & 2   \\
rus\_100\_200\_3\_3  & 1094  & 100710  & 2   \\
rus\_100\_200\_4\_3  & 1094  & 100710  & 2   \\
rus\_100\_200\_5\_3  & 1094  & 100710  & 2   \\
rus\_100\_200\_6\_1  & 1094  & 100710  & 2   \\
rus\_100\_200\_7\_1  & 1094  & 100710  & 2   \\
\midrule
wcsp\_11             & 230   & 5226  & 4 \\
wcsp\_12             & 311   & 5732  & 4 \\
wcsp\_15             & 125   & 736  & 4 \\
\midrule
ObjectDetection\_11  & 60  & 1830 & 11 \\
ObjectDetection\_18  & 60  & 1830 & 16 \\
\midrule
ProteinFolding\_19   & 60   & 1830   & 11 \\
ProteinFolding\_23   & 60   & 1830   & 11 \\
\bottomrule
\end{tabular}
}
\end{table*}

\textbf{Anytime AOBB MAP improvement over MBE for SMNN (monotone-by-construction NN) and UMLP (unconstrained NN) on all 30 PGMs (single merged view; high-treewidth bucket on top, moderate/low-treewidth bucket below).}
  For each PGM both methods run on \emph{the same evidence query and the same AOBB ordering configuration}, so differences isolate the architecture (monotone vs unconstrained). $\Delta = ($NN MAP $-$ MBE MAP$) / |$MBE MAP$|$; positive values mean the NN-augmented search found a strictly better MAP than the MBE baseline within the wall-clock budget. Two stratified-conformal calibration levels are reported side-by-side. Reported on the 3 candidate queries per PGM (mean over the 3); \textbf{bold} = strict win over MBE ($\Delta>10^{-4}$). Where UMLP goes deeply negative (rus, BN\_70) it produces invalid bounds because the SMNN-tuned desc/nn-prune configuration is only safe for monotone heuristics. \emph{Both buckets} use a 600\,s wall-clock deep budget per query, so all 1\,m / 5\,m / 10\,m anytime checkpoints are genuinely measured bounds (no extrapolation). For LT-bucket PGMs where daoopt's search verifiably terminates within seconds (BN\_9/12/14, Seg\_12/18, grid10$\times$10) the optimal bound is reached early and the three checkpoints repeat the same value by construction; for the moderate-treewidth grid20$\times$20 instances and Seg\_14 the search continues past 1\,m and the 5\,m / 10\,m checkpoints reflect MBE's anytime catch-up effect (MBE narrows the gap as it finds better solutions, hence $\Delta$ shrinks slightly with time). BN\_9/12/14 and grid10$\times$10 saturate at $\Delta=0$ because MBE-$i$4 already finds the optimum at root, leaving no headroom for either NN. The two unconstrained-only failure modes (rus and BN\_70 going $-$0.14 to $-$0.42) are reported verbatim and motivate the monotone-by-construction architecture for safe deployment in anytime AOBB search.

\section{Computational Costs}
L2B is an \emph{amortized} method, trading time to train against time to infer. Label generation dominates: computing $i$-bound-10 labels costs $0.5$--$300$\,s per query, and a $20{,}000$-example dataset therefore takes a one-time $\approx 3$\,h (Bayesian networks) to $\approx 60$\,h (random-uniform), against which predictor training ($\le 10$\,min) and calibration ($<0.01$\,ms per query) are negligible. This upfront cost is justified precisely when a model is queried repeatedly: a standalone MBE-$i4$ bound already costs $35$--$235$\,s per query on the random-uniform family (Table~\ref{tab:pgm_details}), so the one-time training cost breaks even after roughly $1{,}000$--$6{,}000$ queries---well within the regime of a solver that evaluates a bound at every search node. This is also why the method targets the amortized setting rather than one-shot inference: for a single query one should spend the budget on inference directly, but branch-and-bound issues millions of bound evaluations against one fixed model
so a learned bound is the only route to a cheap tight bound at all.

\section{Timing Results}
\begin{figure}[t]
\centering
\includegraphics[width=0.95\columnwidth]{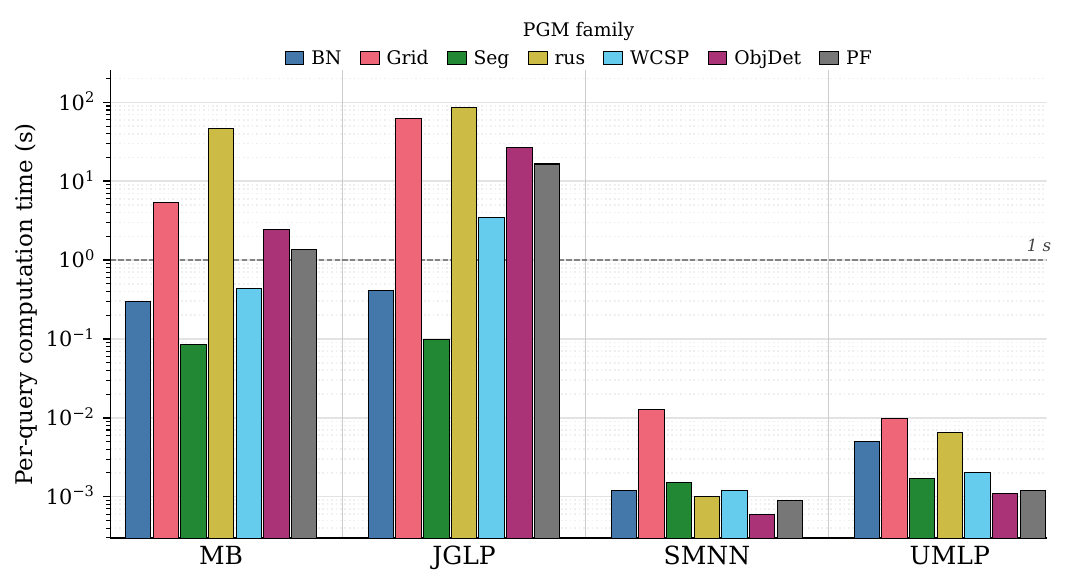}
\caption{Per-query bound computation time across seven PGM families, grouped by method (log scale). Bars report mean wall-clock time per test query, with $n$ instances per family in parentheses: BN (6), Grid (7), Seg (3), rus (7), WCSP (3), Object Detection (2), Protein Folding (2). MB and JGLP run in seconds to minutes; SMNN (Section~\ref{sec:smnn}) and UMLP compute bounds in milliseconds---3--5 orders of magnitude faster---on every family. SMNN is marginally slower than UMLP due to its positive-weight reparameterization and batch-norm-without-affine layers, but both remain below $10^{-2}$\,s.}

\label{fig:pareto}
\end{figure}

\section{Alternative Encoding: Consecutive 2-Hot}
\label{app:twohot_results}

\begin{sidewaystable}[!htbp]
  \centering
  \scriptsize
  \setlength{\tabcolsep}{3pt}
  \caption{Upper-bound calibrated quality of \textbf{SMNN-1hot} (existing monotone-by-construction with one-hot encoding, length $d$ per var) vs \textbf{SMNN-2hot} (same architecture / data / calibration but with consecutive 2-hot encoding, length $d{+}1$ per var, observed value $k\to$ bits $k$ and $k{+}1$ set; see Scheme A). Both methods evaluated on the same per-PGM \emph{locked test subsets} and the same calibration set with stratified-conformal calibration; the only axis of variation is the input encoding. Gap = mean$\pm$std of $|$bound$-$true$|/|$true$|$; Cov = empirical coverage. Three target levels $\gamma\in\{0.95, 0.99, 0.999\}$ are reported side-by-side. \textbf{Bold} = strictly lower gap between 1hot and 2hot at that (PGM, $\gamma$). Deterministic MB-$i$4 / JGLP-$i$4 gaps included for context.}
  \label{tab:smnn_1hot_vs_2hot_3gamma_locked}
  \resizebox{\textheight}{!}{%

  \begin{tabular}{l r | c | c | cc cc | cc cc | cc cc}
  \toprule
    \multirow{3}{*}{Dataset} & \multirow{3}{*}{MPE$^{\star}$} & \multirow{3}{*}{MB} & \multirow{3}{*}{JGLP} & \multicolumn{4}{c|}{$\gamma=0.95$} & \multicolumn{4}{c|}{$\gamma=0.99$} & \multicolumn{4}{c}{$\gamma=0.999$} \\
    & & & & \multicolumn{2}{c}{1hot} & \multicolumn{2}{c|}{2hot} & \multicolumn{2}{c}{1hot} & \multicolumn{2}{c|}{2hot} & \multicolumn{2}{c}{1hot} & \multicolumn{2}{c}{2hot} \\
    & & Gap & Gap & Cov & Gap & Cov & Gap & Cov & Gap & Cov & Gap & Cov & Gap & Cov & Gap \\
    \midrule
    BN\_9 & -23.81 & 0.3656$\pm$0.1738 & 0.3649$\pm$0.1745 & 0.97 & \textbf{0.0683$\pm$0.0351} & 1.00 & 0.2612$\pm$0.1264 & 1.00 & \textbf{0.0947$\pm$0.0366} & 1.00 & 0.3219$\pm$0.1314 & 1.00 & \textbf{0.1285$\pm$0.0396} & 1.00 & 0.3856$\pm$0.1414 \\
    BN\_12 & -19.71 & 0.3490$\pm$0.1847 & 0.3438$\pm$0.1896 & 0.97 & 0.1113$\pm$0.0615 & 0.97 & \textbf{0.1038$\pm$0.0551} & 0.99 & 0.1473$\pm$0.0680 & 0.99 & \textbf{0.1365$\pm$0.0582} & 0.99 & 0.1989$\pm$0.0858 & 0.99 & \textbf{0.1862$\pm$0.0741} \\
    BN\_14 & -26.51 & 0.3429$\pm$0.1935 & 0.3411$\pm$0.1954 & 0.92 & \textbf{0.0656$\pm$0.0385} & 0.95 & 0.0986$\pm$0.0562 & 0.99 & \textbf{0.0911$\pm$0.0425} & 0.97 & 0.1287$\pm$0.0624 & 0.99 & \textbf{0.1165$\pm$0.0469} & 1.00 & 0.1623$\pm$0.0676 \\
    BN\_30 & -14.66 & 0.3017$\pm$0.1558 & 0.3017$\pm$0.1558 & 1.00 & \textbf{0.0143$\pm$0.0137} & 1.00 & 0.0290$\pm$0.0131 & 1.00 & \textbf{0.0624$\pm$0.0119} & 1.00 & 0.0659$\pm$0.0124 & 1.00 & \textbf{0.0897$\pm$0.0142} & 1.00 & 0.0914$\pm$0.0200 \\
    BN\_32 & -16.58 & 0.3379$\pm$0.1860 & 0.3379$\pm$0.1860 & 1.00 & \textbf{0.0163$\pm$0.0115} & 1.00 & 0.0258$\pm$0.0122 & 1.00 & \textbf{0.0491$\pm$0.0197} & 1.00 & 0.0569$\pm$0.0180 & 1.00 & \textbf{0.0926$\pm$0.0230} & 1.00 & 0.0927$\pm$0.0249 \\
    BN\_70 & -86.65 & 0.4617$\pm$0.2349 & 0.4475$\pm$0.2403 & 1.00 & \textbf{0.0239$\pm$0.0115} & 1.00 & 0.0240$\pm$0.0117 & 1.00 & \textbf{0.0371$\pm$0.0147} & 1.00 & 0.0372$\pm$0.0150 & 1.00 & \textbf{0.0534$\pm$0.0154} & 1.00 & 0.0535$\pm$0.0154 \\
    \midrule
    grid10x10.f5.wrap & 159.91 & 0.0438$\pm$0.0209 & 0.0248$\pm$0.0156 & 1.00 & 0.0140$\pm$0.0105 & 0.73 & \textbf{0.0126$\pm$0.0108} & 1.00 & 0.0360$\pm$0.0108 & 1.00 & \textbf{0.0294$\pm$0.0133} & 1.00 & 0.0454$\pm$0.0109 & 1.00 & \textbf{0.0436$\pm$0.0136} \\
    grid20x20.f5.wrap & 617.89 & 0.0615$\pm$0.0252 & 0.0143$\pm$0.0113 & 1.00 & \textbf{0.0066$\pm$0.0063} & 0.97 & 0.0097$\pm$0.0063 & 1.00 & 0.0189$\pm$0.0098 & 1.00 & \textbf{0.0187$\pm$0.0090} & 1.00 & \textbf{0.0279$\pm$0.0107} & 1.00 & 0.0324$\pm$0.0080 \\
    grid20x20.f10 & 1230.40 & 0.0443$\pm$0.0180 & 0.0111$\pm$0.0094 & 1.00 & \textbf{0.0064$\pm$0.0072} & 1.00 & 0.0080$\pm$0.0073 & 1.00 & \textbf{0.0188$\pm$0.0117} & 1.00 & 0.0211$\pm$0.0114 & 1.00 & 0.0413$\pm$0.0101 & 1.00 & \textbf{0.0398$\pm$0.0091} \\
    grid40x40.f2 & 1071.08 & 0.0329$\pm$0.0130 & 0.0064$\pm$0.0036 & 1.00 & \textbf{0.0029$\pm$0.0019} & 1.00 & 0.0030$\pm$0.0024 & 1.00 & \textbf{0.0098$\pm$0.0031} & 1.00 & 0.0100$\pm$0.0034 & 1.00 & \textbf{0.0207$\pm$0.0078} & 1.00 & 0.0210$\pm$0.0078 \\
    grid40x40.f10.wrap & 5308.25 & 0.0596$\pm$0.0146 & 0.0092$\pm$0.0024 & 0.33 & \textbf{0.0025$\pm$0.0018} & 0.40 & 0.0034$\pm$0.0020 & 1.00 & \textbf{0.0051$\pm$0.0028} & 0.97 & 0.0070$\pm$0.0038 & 1.00 & \textbf{0.0108$\pm$0.0028} & 1.00 & 0.0132$\pm$0.0039 \\
    grid80x80.f10 & 20140.10 & 0.0447$\pm$0.0268 & 0.0036$\pm$0.0016 & 0.17 & \textbf{0.0020$\pm$0.0016} & 0.17 & 0.0020$\pm$0.0016 & 1.00 & \textbf{0.0021$\pm$0.0016} & 1.00 & 0.0023$\pm$0.0016 & 1.00 & \textbf{0.0067$\pm$0.0046} & 1.00 & 0.0071$\pm$0.0048 \\
    grid80x80.f5.wrap & 9168.14 & 0.0107$\pm$0.0022 & 0.0084$\pm$0.0026 & 1.00 & \textbf{0.0050$\pm$0.0010} & 1.00 & 0.0244$\pm$0.0146 & 1.00 & \textbf{0.0102$\pm$0.0031} & 1.00 & 0.0308$\pm$0.0118 & 1.00 & \textbf{0.0143$\pm$0.0050} & 1.00 & 0.0357$\pm$0.0096 \\
    \midrule
    Segmentation\_12 & -146.96 & 0.5361$\pm$0.1969 & 0.5361$\pm$0.1969 & 0.97 & \textbf{0.1199$\pm$0.0453} & 0.97 & 0.1787$\pm$0.0618 & 1.00 & \textbf{0.1498$\pm$0.0480} & 1.00 & 0.2186$\pm$0.0562 & 1.00 & \textbf{0.1936$\pm$0.0523} & 1.00 & 0.2513$\pm$0.0466 \\
    Segmentation\_14 & -143.57 & 0.5133$\pm$0.2128 & 0.5134$\pm$0.2128 & 0.99 & \textbf{0.1356$\pm$0.0477} & 0.99 & 0.1868$\pm$0.0770 & 1.00 & \textbf{0.1632$\pm$0.0435} & 0.99 & 0.2183$\pm$0.0721 & 1.00 & \textbf{0.1926$\pm$0.0400} & 1.00 & 0.2518$\pm$0.0632 \\
    Segmentation\_18 & -149.81 & 0.5369$\pm$0.1844 & 0.5369$\pm$0.1844 & 0.98 & 0.1838$\pm$0.0645 & 0.97 & \textbf{0.1500$\pm$0.0487} & 1.00 & 0.2209$\pm$0.0540 & 1.00 & \textbf{0.1855$\pm$0.0465} & 1.00 & 0.2565$\pm$0.0439 & 1.00 & \textbf{0.2183$\pm$0.0435} \\
    \midrule
    rus\_100\_200\_2\_1 & 2776.12 & 0.4418$\pm$0.2617 & 0.3779$\pm$0.1738 & 0.99 & \textbf{0.0556$\pm$0.0341} & 0.99 & 0.1409$\pm$0.0855 & 1.00 & \textbf{0.0652$\pm$0.0347} & 1.00 & 0.1530$\pm$0.0788 & 1.00 & \textbf{0.0704$\pm$0.0364} & 1.00 & 0.1596$\pm$0.0747 \\
    rus\_100\_200\_3\_1 & 2726.97 & 0.6203$\pm$0.1124 & 0.6128$\pm$0.1284 & 0.98 & \textbf{0.1995$\pm$0.1393} & 1.00 & 0.2574$\pm$0.0638 & 1.00 & \textbf{0.2230$\pm$0.1375} & 1.00 & 0.2652$\pm$0.0631 & 1.00 & \textbf{0.2324$\pm$0.1377} & 1.00 & 0.2677$\pm$0.0624 \\
    rus\_100\_200\_3\_3 & 2745.43 & 0.5360$\pm$0.2253 & 0.4392$\pm$0.2550 & 0.98 & \textbf{0.0805$\pm$0.0583} & 0.97 & 0.2265$\pm$0.1343 & 1.00 & \textbf{0.0991$\pm$0.0607} & 1.00 & 0.2402$\pm$0.1253 & 1.00 & \textbf{0.1089$\pm$0.0574} & 1.00 & 0.2456$\pm$0.1251 \\
    rus\_100\_200\_4\_3 & 2740.65 & 0.5523$\pm$0.2164 & 0.4341$\pm$0.2447 & 0.96 & \textbf{0.0825$\pm$0.0538} & 0.98 & 0.2557$\pm$0.1464 & 1.00 & \textbf{0.0966$\pm$0.0564} & 1.00 & 0.2635$\pm$0.1461 & 1.00 & \textbf{0.1066$\pm$0.0563} & 1.00 & 0.2707$\pm$0.1393 \\
    rus\_100\_200\_5\_3 & 2739.05 & 0.6095$\pm$0.1520 & 0.5134$\pm$0.2328 & 0.98 & \textbf{0.0835$\pm$0.0633} & 0.98 & 0.2638$\pm$0.1246 & 1.00 & \textbf{0.1000$\pm$0.0706} & 1.00 & 0.2755$\pm$0.1237 & 1.00 & \textbf{0.1070$\pm$0.0706} & 1.00 & 0.2807$\pm$0.1216 \\
    rus\_100\_200\_6\_1 & 2741.41 & 0.5734$\pm$0.2214 & 0.4945$\pm$0.2470 & 0.98 & \textbf{0.0776$\pm$0.0514} & 1.00 & 0.1739$\pm$0.0897 & 1.00 & \textbf{0.0910$\pm$0.0550} & 1.00 & 0.1842$\pm$0.0864 & 1.00 & \textbf{0.1024$\pm$0.0561} & 1.00 & 0.1951$\pm$0.0845 \\
    rus\_100\_200\_7\_1 & 2739.09 & 0.5514$\pm$0.2083 & 0.4508$\pm$0.2410 & 0.95 & \textbf{0.0925$\pm$0.0667} & 0.95 & 0.2259$\pm$0.1238 & 0.96 & \textbf{0.1074$\pm$0.0699} & 0.96 & 0.2363$\pm$0.1209 & 1.00 & \textbf{0.1162$\pm$0.0646} & 1.00 & 0.2439$\pm$0.1158 \\
    \midrule
    wcsp\_11 & -463.08 & 0.9940$\pm$0.0052 & 0.9940$\pm$0.0052 & 0.99 & \textbf{0.2035$\pm$0.1260} & 0.97 & 0.4826$\pm$0.2328 & 1.00 & \textbf{0.2981$\pm$0.1572} & 1.00 & 0.6300$\pm$0.2627 & 1.00 & \textbf{0.3551$\pm$0.1633} & 1.00 & 0.7820$\pm$0.2619 \\
    wcsp\_12 & -562.67 & 0.9956$\pm$0.0048 & 0.9955$\pm$0.0048 & 0.99 & \textbf{0.1890$\pm$0.1454} & 0.99 & 0.3697$\pm$0.1686 & 1.00 & \textbf{0.2479$\pm$0.1625} & 1.00 & 0.4834$\pm$0.1713 & 1.00 & \textbf{0.3271$\pm$0.1766} & 1.00 & 0.5555$\pm$0.1681 \\
    wcsp\_15 & -328.00 & 0.6592$\pm$0.1099 & 0.6536$\pm$0.1156 & 0.97 & \textbf{0.1933$\pm$0.1026} & 1.00 & 0.3205$\pm$0.0668 & 1.00 & \textbf{0.2688$\pm$0.1058} & 1.00 & 0.3807$\pm$0.0703 & 1.00 & \textbf{0.2945$\pm$0.1063} & 1.00 & 0.4248$\pm$0.0734 \\
    \midrule
    ObjectDetection\_11 & 1999.86 & 0.6288$\pm$0.0979 & 0.7080$\pm$0.5687 & 0.81 & \textbf{0.2650$\pm$0.2800} & 0.94 & 0.5413$\pm$0.7482 & 0.81 & \textbf{0.3744$\pm$0.3626} & 0.94 & 0.7159$\pm$0.9001 & 1.00 & \textbf{0.5001$\pm$0.4505} & 0.94 & 0.8025$\pm$0.9360 \\
    ObjectDetection\_18 & 3313.81 & 0.8564$\pm$0.0000 & 0.5547$\pm$0.1644 & 0.75 & \textbf{0.1156$\pm$0.1126} & 0.83 & 0.2295$\pm$0.1981 & 1.00 & \textbf{0.2036$\pm$0.1369} & 1.00 & 0.3349$\pm$0.2303 & 1.00 & \textbf{0.3699$\pm$0.1213} & 1.00 & 0.4523$\pm$0.2578 \\
    \midrule
    ProteinFolding\_19 & 1557.28 & 0.6800$\pm$0.0000 & 0.5385$\pm$0.1350 & 1.00 & 0.0651$\pm$0.0432 & 0.75 & \textbf{0.0649$\pm$0.0471} & 1.00 & 0.1674$\pm$0.0430 & 1.00 & \textbf{0.1644$\pm$0.0521} & 1.00 & 0.2844$\pm$0.0601 & 1.00 & \textbf{0.2296$\pm$0.0802} \\
    ProteinFolding\_23 & 3165.49 & 0.7055$\pm$0.0879 & 0.4682$\pm$0.1535 & 0.94 & \textbf{0.1687$\pm$0.1564} & 0.94 & 0.1948$\pm$0.1655 & 0.94 & \textbf{0.2416$\pm$0.1721} & 0.94 & 0.2783$\pm$0.2045 & 1.00 & \textbf{0.2818$\pm$0.1579} & 0.94 & 0.3168$\pm$0.1979 \\
    \bottomrule
  \end{tabular}
  }
  \par\smallskip\footnotesize
  \textbf{Strict-better-gap counts (out of 30):} $\gamma{=}0.95$: 1hot 26, 2hot 4, tie 0; $\gamma{=}0.99$: 1hot 25, 2hot 5, tie 0; $\gamma{=}0.999$: 1hot 25, 2hot 5, tie 0.
\end{sidewaystable}

\begin{sidewaystable}[!htbp]
  \centering
  \scriptsize
  \caption{Upper-bound quality of \textbf{SMNN-2hot} (monotone-by-construction + \emph{consecutive 2-hot} encoding: per variable of domain $d$, length $d{+}1$, observed value $k\to$ bits $k$ and $k{+}1$ set; preserves monotonicity for any $d$) on per-PGM \emph{locked test subsets} (same queries used in the AOBB integration), with \textbf{Gaussian} calibration at $i$-bound=4 reported at three $\gamma$ levels (0.95, 0.99, 0.999). Deterministic baselines MB-$i$4 and JGLP-$i$4 included for context. Gap = mean relative error (lower is better); Cov = empirical coverage. Trained on the same data splits as the one-hot SMNN reported in the main table; only the input encoding differs.}
  \label{tab:ub_smnn_2hot_gaussian_locked_3g}
  \begin{adjustbox}{max width=\textwidth}
  \begin{tabular}{l r | ccccc | cc cc cc}
  \toprule
    \multirow{3}{*}{Dataset} & \multirow{3}{*}{MPE$^{\star}$} & \multicolumn{5}{c|}{Deterministic} & \multicolumn{6}{c}{Gaussian (SMNN-2hot)} \\
    \cmidrule(lr){3-7} \cmidrule(lr){8-13}
    & & \multicolumn{2}{c}{MB} & \multicolumn{2}{c}{JGLP} & SMNN-2hot & \multicolumn{2}{c}{$\gamma=0.95$} & \multicolumn{2}{c}{$\gamma=0.99$} & \multicolumn{2}{c}{$\gamma=0.999$} \\
    & & Gap & Time(s) & Gap & Time(s) & Time(s) & Cov & Gap & Cov & Gap & Cov & Gap \\
    \midrule
    BN\_9 & -23.81 & 0.3656$\pm$0.1738 & 0.0821 & 0.3649$\pm$0.1745 & 0.0968 & 0.0011 & 1.00 & 0.3094$\pm$0.0621 & 1.00 & 0.3881$\pm$0.0527 & 1.00 & 0.4782$\pm$0.0421 \\
    BN\_12 & -19.71 & 0.3490$\pm$0.1847 & 0.0799 & 0.3438$\pm$0.1896 & 0.1053 & 0.0018 & 0.99 & 0.1010$\pm$0.0440 & 1.00 & 0.1377$\pm$0.0429 & 1.00 & 0.1802$\pm$0.0412 \\
    BN\_14 & -26.51 & 0.3429$\pm$0.1935 & 0.0745 & 0.3411$\pm$0.1954 & 0.0974 & 0.0016 & 0.99 & 0.0995$\pm$0.0490 & 1.00 & 0.1321$\pm$0.0459 & 1.00 & 0.1695$\pm$0.0425 \\
    BN\_30 & -14.66 & 0.3017$\pm$0.1558 & 0.1316 & 0.3017$\pm$0.1558 & 0.1387 & 0.0184 & 1.00 & 0.0239$\pm$0.0138 & 1.00 & 0.0315$\pm$0.0137 & 1.00 & 0.0402$\pm$0.0136 \\
    BN\_32 & -16.58 & 0.3379$\pm$0.1860 & 0.1834 & 0.3379$\pm$0.1860 & 0.2325 & 0.0125 & 1.00 & 0.0208$\pm$0.0118 & 1.00 & 0.0275$\pm$0.0117 & 1.00 & 0.0352$\pm$0.0116 \\
    BN\_70 & -86.65 & 0.4617$\pm$0.2349 & 1.2170 & 0.4475$\pm$0.2403 & 1.7728 & 0.0115 & 1.00 & 0.0208$\pm$0.0110 & 1.00 & 0.0272$\pm$0.0109 & 1.00 & 0.0345$\pm$0.0108 \\
    \midrule
    grid10x10.f5.wrap & 159.91 & 0.0438$\pm$0.0209 & 0.3944 & 0.0248$\pm$0.0156 & 2.1251 & 0.0044 & 1.00 & 0.0240$\pm$0.0131 & 1.00 & 0.0643$\pm$0.0140 & 1.00 & 0.1104$\pm$0.0151 \\
    grid20x20.f5.wrap & 617.89 & 0.0615$\pm$0.0252 & 0.6182 & 0.0143$\pm$0.0113 & 14.9056 & 0.0019 & 1.00 & 0.0294$\pm$0.0118 & 1.00 & 0.0691$\pm$0.0134 & 1.00 & 0.1147$\pm$0.0153 \\
    grid20x20.f10 & 1230.40 & 0.0443$\pm$0.0180 & 0.6024 & 0.0111$\pm$0.0094 & 13.6322 & 0.0045 & 1.00 & 0.0243$\pm$0.0148 & 1.00 & 0.0602$\pm$0.0165 & 1.00 & 0.1013$\pm$0.0185 \\
    grid40x40.f2 & 1071.08 & 0.0329$\pm$0.0130 & 2.1179 & 0.0064$\pm$0.0036 & 77.6967 & 0.0037 & 1.00 & 0.0486$\pm$0.0221 & 1.00 & 0.1021$\pm$0.0257 & 1.00 & 0.1634$\pm$0.0297 \\
    grid40x40.f10.wrap & 5308.25 & 0.0596$\pm$0.0146 & 2.7172 & 0.0092$\pm$0.0024 & 87.3251 & 0.0026 & 1.00 & 0.0338$\pm$0.0041 & 1.00 & 0.1024$\pm$0.0048 & 1.00 & 0.1810$\pm$0.0056 \\
    grid80x80.f10 & 20140.10 & 0.0447$\pm$0.0268 & 16.8307 & 0.0036$\pm$0.0016 & 120.2345 & 0.0114 & 1.00 & 0.0455$\pm$0.0148 & 1.00 & 0.0871$\pm$0.0182 & 1.00 & 0.1346$\pm$0.0220 \\
    grid80x80.f5.wrap & 9168.14 & 0.0107$\pm$0.0022 & 14.3095 & 0.0084$\pm$0.0026 & 120.2831 & 0.0181 & 1.00 & 0.1401$\pm$0.0539 & 1.00 & 0.2146$\pm$0.0621 & 1.00 & 0.2999$\pm$0.0715 \\
    \midrule
    Segmentation\_12 & -146.96 & 0.5361$\pm$0.1969 & 0.0825 & 0.5361$\pm$0.1969 & 0.0903 & 0.0164 & 1.00 & 0.2627$\pm$0.0475 & 1.00 & 0.3406$\pm$0.0400 & 1.00 & 0.4299$\pm$0.0400 \\
    Segmentation\_14 & -143.57 & 0.5133$\pm$0.2128 & 0.0876 & 0.5134$\pm$0.2128 & 0.1062 & 0.0166 & 1.00 & 0.2547$\pm$0.0709 & 1.00 & 0.3350$\pm$0.0505 & 1.00 & 0.4270$\pm$0.0324 \\
    Segmentation\_18 & -149.81 & 0.5369$\pm$0.1844 & 0.0841 & 0.5369$\pm$0.1844 & 0.0952 & 0.0172 & 0.99 & 0.2076$\pm$0.0468 & 0.99 & 0.2701$\pm$0.0480 & 1.00 & 0.3456$\pm$0.0409 \\
    \midrule
    rus\_100\_200\_2\_1 & 2776.12 & 0.4418$\pm$0.2617 & 56.0038 & 0.3779$\pm$0.1738 & 104.7771 & 0.0202 & 0.91 & 0.2214$\pm$0.1110 & 0.97 & 0.2758$\pm$0.1294 & 1.00 & 0.3442$\pm$0.1394 \\
    rus\_100\_200\_3\_1 & 2726.97 & 0.6203$\pm$0.1124 & 21.4693 & 0.6128$\pm$0.1284 & 24.5238 & 0.0130 & 0.98 & 0.3745$\pm$0.0622 & 1.00 & 0.4576$\pm$0.0700 & 1.00 & 0.5534$\pm$0.0753 \\
    rus\_100\_200\_3\_3 & 2745.43 & 0.5360$\pm$0.2253 & 53.4856 & 0.4392$\pm$0.2550 & 97.6758 & 0.0124 & 0.92 & 0.3312$\pm$0.2023 & 1.00 & 0.4169$\pm$0.2265 & 1.00 & 0.5185$\pm$0.2485 \\
    rus\_100\_200\_4\_3 & 2740.65 & 0.5523$\pm$0.2164 & 54.8099 & 0.4341$\pm$0.2447 & 104.7959 & 0.0121 & 0.95 & 0.3599$\pm$0.2269 & 1.00 & 0.4538$\pm$0.2510 & 1.00 & 0.5634$\pm$0.2753 \\
    rus\_100\_200\_5\_3 & 2739.05 & 0.6095$\pm$0.1520 & 45.8743 & 0.5134$\pm$0.2328 & 79.5065 & 0.0125 & 0.91 & 0.3606$\pm$0.1842 & 1.00 & 0.4462$\pm$0.2076 & 1.00 & 0.5479$\pm$0.2274 \\
    rus\_100\_200\_6\_1 & 2741.41 & 0.5734$\pm$0.2214 & 44.0984 & 0.4945$\pm$0.2470 & 89.2765 & 0.0130 & 0.86 & 0.2632$\pm$0.1375 & 1.00 & 0.3302$\pm$0.1585 & 1.00 & 0.4115$\pm$0.1742 \\
    rus\_100\_200\_7\_1 & 2739.09 & 0.5514$\pm$0.2083 & 48.7584 & 0.4508$\pm$0.2410 & 94.1580 & 0.0126 & 0.86 & 0.3367$\pm$0.1905 & 0.96 & 0.4184$\pm$0.2203 & 1.00 & 0.5196$\pm$0.2415 \\
    \midrule
    wcsp\_11 & -463.08 & 0.9940$\pm$0.0052 & 0.6401 & 0.9940$\pm$0.0052 & 5.3184 & 0.0163 & 1.00 & 0.7707$\pm$0.2027 & 1.00 & 1.0709$\pm$0.3912 & 1.00 & 1.4147$\pm$0.6155 \\
    wcsp\_12 & -562.67 & 0.9956$\pm$0.0048 & 0.6004 & 0.9955$\pm$0.0048 & 4.8372 & 0.0176 & 1.00 & 0.6317$\pm$0.2867 & 1.00 & 0.8915$\pm$0.4747 & 1.00 & 1.1890$\pm$0.6983 \\
    wcsp\_15 & -328.00 & 0.6592$\pm$0.1099 & 0.0859 & 0.6536$\pm$0.1156 & 0.1499 & 0.0245 & 1.00 & 0.3923$\pm$0.0579 & 1.00 & 0.4840$\pm$0.0618 & 1.00 & 0.5890$\pm$0.0677 \\
    \midrule
    ObjectDetection\_11 & 1999.86 & 0.6288$\pm$0.0979 & 1.7946 & 0.7080$\pm$0.5687 & 27.9568 & 0.0224 & 0.81 & 0.5663$\pm$0.7840 & 0.94 & 0.7070$\pm$0.8978 & 1.00 & 0.8838$\pm$1.0191 \\
    ObjectDetection\_18 & 3313.81 & 0.8564$\pm$0.0000 & 3.0544 & 0.5547$\pm$0.1644 & 25.7126 & 0.0137 & 0.83 & 0.2265$\pm$0.1973 & 1.00 & 0.3590$\pm$0.2378 & 1.00 & 0.5157$\pm$0.2793 \\
    \midrule
    ProteinFolding\_19 & 1557.28 & 0.6800$\pm$0.0000 & 1.6974 & 0.5385$\pm$0.1350 & 21.9289 & 0.0235 & 1.00 & 0.0891$\pm$0.0338 & 1.00 & 0.1956$\pm$0.0456 & 1.00 & 0.3176$\pm$0.0670 \\
    ProteinFolding\_23 & 3165.49 & 0.7055$\pm$0.0879 & 1.0269 & 0.4682$\pm$0.1535 & 11.1632 & 0.0240 & 0.94 & 0.2122$\pm$0.1826 & 1.00 & 0.3522$\pm$0.2426 & 1.00 & 0.5255$\pm$0.3018 \\
    \bottomrule
  \end{tabular}
  \end{adjustbox}
\end{sidewaystable}

\begin{sidewaystable}[!htbp]
  \centering
  \scriptsize
  \caption{Upper-bound quality of \textbf{SMNN-2hot} (monotone-by-construction + \emph{consecutive 2-hot} encoding: per variable of domain $d$, length $d{+}1$, observed value $k\to$ bits $k$ and $k{+}1$ set; preserves monotonicity for any $d$) on per-PGM \emph{locked test subsets} (same queries used in the AOBB integration), with \textbf{Conformal} calibration at $i$-bound=4 reported at three $\gamma$ levels (0.95, 0.99, 0.999). Deterministic baselines MB-$i$4 and JGLP-$i$4 included for context. Gap = mean relative error (lower is better); Cov = empirical coverage. Trained on the same data splits as the one-hot SMNN reported in the main table; only the input encoding differs.}
  \label{tab:ub_smnn_2hot_conformal_locked_3g}
  \begin{adjustbox}{max width=\textwidth}
  \begin{tabular}{l r | ccccc | cc cc cc}
  \toprule
    \multirow{3}{*}{Dataset} & \multirow{3}{*}{MPE$^{\star}$} & \multicolumn{5}{c|}{Deterministic} & \multicolumn{6}{c}{Conformal (SMNN-2hot)} \\
    \cmidrule(lr){3-7} \cmidrule(lr){8-13}
    & & \multicolumn{2}{c}{MB} & \multicolumn{2}{c}{JGLP} & SMNN-2hot & \multicolumn{2}{c}{$\gamma=0.95$} & \multicolumn{2}{c}{$\gamma=0.99$} & \multicolumn{2}{c}{$\gamma=0.999$} \\
    & & Gap & Time(s) & Gap & Time(s) & Time(s) & Cov & Gap & Cov & Gap & Cov & Gap \\
    \midrule
    BN\_9 & -23.81 & 0.3656$\pm$0.1738 & 0.0821 & 0.3649$\pm$0.1745 & 0.0968 & 0.0011 & 1.00 & 0.3084$\pm$0.0622 & 1.00 & 0.4001$\pm$0.0512 & 1.00 & 0.4808$\pm$0.0418 \\
    BN\_12 & -19.71 & 0.3490$\pm$0.1847 & 0.0799 & 0.3438$\pm$0.1896 & 0.1053 & 0.0018 & 0.99 & 0.1026$\pm$0.0440 & 1.00 & 0.1600$\pm$0.0419 & 1.00 & 0.2138$\pm$0.0405 \\
    BN\_14 & -26.51 & 0.3429$\pm$0.1935 & 0.0745 & 0.3411$\pm$0.1954 & 0.0974 & 0.0016 & 0.97 & 0.0968$\pm$0.0490 & 1.00 & 0.1333$\pm$0.0458 & 1.00 & 0.1931$\pm$0.0406 \\
    BN\_30 & -14.66 & 0.3017$\pm$0.1558 & 0.1316 & 0.3017$\pm$0.1558 & 0.1387 & 0.0184 & 1.00 & 0.0292$\pm$0.0137 & 1.00 & 0.0615$\pm$0.0133 & 1.00 & 0.0849$\pm$0.0129 \\
    BN\_32 & -16.58 & 0.3379$\pm$0.1860 & 0.1834 & 0.3379$\pm$0.1860 & 0.2325 & 0.0125 & 1.00 & 0.0223$\pm$0.0118 & 1.00 & 0.0450$\pm$0.0115 & 1.00 & 0.1022$\pm$0.0108 \\
    BN\_70 & -86.65 & 0.4617$\pm$0.2349 & 1.2170 & 0.4475$\pm$0.2403 & 1.7728 & 0.0115 & 1.00 & 0.0205$\pm$0.0110 & 1.00 & 0.0359$\pm$0.0108 & 1.00 & 0.0564$\pm$0.0106 \\
    \midrule
    grid10x10.f5.wrap & 159.91 & 0.0438$\pm$0.0209 & 0.3944 & 0.0248$\pm$0.0156 & 2.1251 & 0.0044 & 0.93 & 0.0188$\pm$0.0127 & 1.00 & 0.0657$\pm$0.0141 & 1.00 & 0.1258$\pm$0.0154 \\
    grid20x20.f5.wrap & 617.89 & 0.0615$\pm$0.0252 & 0.6182 & 0.0143$\pm$0.0113 & 14.9056 & 0.0019 & 0.97 & 0.0128$\pm$0.0111 & 1.00 & 0.0419$\pm$0.0123 & 1.00 & 0.0696$\pm$0.0134 \\
    grid20x20.f10 & 1230.40 & 0.0443$\pm$0.0180 & 0.6024 & 0.0111$\pm$0.0094 & 13.6322 & 0.0045 & 1.00 & 0.0122$\pm$0.0143 & 1.00 & 0.0352$\pm$0.0154 & 1.00 & 0.0615$\pm$0.0166 \\
    grid40x40.f2 & 1071.08 & 0.0329$\pm$0.0130 & 2.1179 & 0.0064$\pm$0.0036 & 77.6967 & 0.0037 & 1.00 & 0.0269$\pm$0.0207 & 1.00 & 0.0493$\pm$0.0222 & 1.00 & 0.0681$\pm$0.0234 \\
    grid40x40.f10.wrap & 5308.25 & 0.0596$\pm$0.0146 & 2.7172 & 0.0092$\pm$0.0024 & 87.3251 & 0.0026 & 0.40 & 0.0034$\pm$0.0020 & 1.00 & 0.0166$\pm$0.0040 & 1.00 & 0.0339$\pm$0.0041 \\
    grid80x80.f10 & 20140.10 & 0.0447$\pm$0.0268 & 16.8307 & 0.0036$\pm$0.0016 & 120.2345 & 0.0114 & 1.00 & 0.0327$\pm$0.0138 & 1.00 & 0.0490$\pm$0.0151 & 1.00 & 0.0596$\pm$0.0160 \\
    grid80x80.f5.wrap & 9168.14 & 0.0107$\pm$0.0022 & 14.3095 & 0.0084$\pm$0.0026 & 120.2831 & 0.0181 & 1.00 & 0.1128$\pm$0.0509 & 1.00 & 0.1337$\pm$0.0532 & 1.00 & 0.1504$\pm$0.0551 \\
    \midrule
    Segmentation\_12 & -146.96 & 0.5361$\pm$0.1969 & 0.0825 & 0.5361$\pm$0.1969 & 0.0903 & 0.0164 & 1.00 & 0.2637$\pm$0.0474 & 1.00 & 0.3250$\pm$0.0411 & 1.00 & 0.3746$\pm$0.0388 \\
    Segmentation\_14 & -143.57 & 0.5133$\pm$0.2128 & 0.0876 & 0.5134$\pm$0.2128 & 0.1062 & 0.0166 & 0.97 & 0.2331$\pm$0.0716 & 1.00 & 0.2767$\pm$0.0652 & 1.00 & 0.3173$\pm$0.0548 \\
    Segmentation\_18 & -149.81 & 0.5369$\pm$0.1844 & 0.0841 & 0.5369$\pm$0.1844 & 0.0952 & 0.0172 & 0.99 & 0.2057$\pm$0.0470 & 0.99 & 0.2473$\pm$0.0455 & 1.00 & 0.2961$\pm$0.0458 \\
    \midrule
    rus\_100\_200\_2\_1 & 2776.12 & 0.4418$\pm$0.2617 & 56.0038 & 0.3779$\pm$0.1738 & 104.7771 & 0.0202 & 0.93 & 0.2366$\pm$0.1173 & 0.99 & 0.2800$\pm$0.1303 & 1.00 & 0.3011$\pm$0.1334 \\
    rus\_100\_200\_3\_1 & 2726.97 & 0.6203$\pm$0.1124 & 21.4693 & 0.6128$\pm$0.1284 & 24.5238 & 0.0130 & 0.98 & 0.3784$\pm$0.0630 & 1.00 & 0.4388$\pm$0.0690 & 1.00 & 0.4885$\pm$0.0717 \\
    rus\_100\_200\_3\_3 & 2745.43 & 0.5360$\pm$0.2253 & 53.4856 & 0.4392$\pm$0.2550 & 97.6758 & 0.0124 & 0.91 & 0.3282$\pm$0.2011 & 0.98 & 0.3827$\pm$0.2188 & 1.00 & 0.4141$\pm$0.2259 \\
    rus\_100\_200\_4\_3 & 2740.65 & 0.5523$\pm$0.2164 & 54.8099 & 0.4341$\pm$0.2447 & 104.7959 & 0.0121 & 0.89 & 0.3451$\pm$0.2209 & 0.98 & 0.4106$\pm$0.2413 & 1.00 & 0.4290$\pm$0.2455 \\
    rus\_100\_200\_5\_3 & 2739.05 & 0.6095$\pm$0.1520 & 45.8743 & 0.5134$\pm$0.2328 & 79.5065 & 0.0125 & 0.88 & 0.3516$\pm$0.1800 & 0.98 & 0.4215$\pm$0.2020 & 1.00 & 0.4555$\pm$0.2094 \\
    rus\_100\_200\_6\_1 & 2741.41 & 0.5734$\pm$0.2214 & 44.0984 & 0.4945$\pm$0.2470 & 89.2765 & 0.0130 & 0.86 & 0.2669$\pm$0.1399 & 1.00 & 0.3320$\pm$0.1588 & 1.00 & 0.3609$\pm$0.1644 \\
    rus\_100\_200\_7\_1 & 2739.09 & 0.5514$\pm$0.2083 & 48.7584 & 0.4508$\pm$0.2410 & 94.1580 & 0.0126 & 0.88 & 0.3421$\pm$0.1935 & 0.95 & 0.3879$\pm$0.2108 & 1.00 & 0.4219$\pm$0.2211 \\
    \midrule
    wcsp\_11 & -463.08 & 0.9940$\pm$0.0052 & 0.6401 & 0.9940$\pm$0.0052 & 5.3184 & 0.0163 & 1.00 & 0.8367$\pm$0.2426 & 1.00 & 1.3575$\pm$0.5779 & 1.00 & 1.9935$\pm$0.9974 \\
    wcsp\_12 & -562.67 & 0.9956$\pm$0.0048 & 0.6004 & 0.9955$\pm$0.0048 & 4.8372 & 0.0176 & 1.00 & 0.7200$\pm$0.3492 & 1.00 & 1.1128$\pm$0.6406 & 1.00 & 1.4450$\pm$0.8929 \\
    wcsp\_15 & -328.00 & 0.6592$\pm$0.1099 & 0.0859 & 0.6536$\pm$0.1156 & 0.1499 & 0.0245 & 1.00 & 0.3446$\pm$0.0563 & 1.00 & 0.3726$\pm$0.0572 & 1.00 & 0.4324$\pm$0.0594 \\
    \midrule
    ObjectDetection\_11 & 1999.86 & 0.6288$\pm$0.0979 & 1.7946 & 0.7080$\pm$0.5687 & 27.9568 & 0.0224 & 0.81 & 0.5725$\pm$0.7896 & 0.94 & 0.7554$\pm$0.9319 & 1.00 & 1.0654$\pm$1.1332 \\
    ObjectDetection\_18 & 3313.81 & 0.8564$\pm$0.0000 & 3.0544 & 0.5547$\pm$0.1644 & 25.7126 & 0.0137 & 0.83 & 0.2260$\pm$0.1969 & 1.00 & 0.3424$\pm$0.2334 & 1.00 & 0.4957$\pm$0.2739 \\
    \midrule
    ProteinFolding\_19 & 1557.28 & 0.6800$\pm$0.0000 & 1.6974 & 0.5385$\pm$0.1350 & 21.9289 & 0.0235 & 1.00 & 0.0880$\pm$0.0337 & 1.00 & 0.1911$\pm$0.0450 & 1.00 & 0.4004$\pm$0.0832 \\
    ProteinFolding\_23 & 3165.49 & 0.7055$\pm$0.0879 & 1.0269 & 0.4682$\pm$0.1535 & 11.1632 & 0.0240 & 0.88 & 0.1956$\pm$0.1737 & 0.94 & 0.2792$\pm$0.2123 & 1.00 & 0.3951$\pm$0.2569 \\
    \bottomrule
  \end{tabular}
  \end{adjustbox}
\end{sidewaystable}

\begin{sidewaystable}[!htbp]
  \centering
  \scriptsize
  \caption{Upper-bound quality of \textbf{SMNN-2hot} (monotone-by-construction + \emph{consecutive 2-hot} encoding: per variable of domain $d$, length $d{+}1$, observed value $k\to$ bits $k$ and $k{+}1$ set; preserves monotonicity for any $d$) on per-PGM \emph{locked test subsets} (same queries used in the AOBB integration), with \textbf{Stratified Conformal} calibration at $i$-bound=4 reported at three $\gamma$ levels (0.95, 0.99, 0.999). Deterministic baselines MB-$i$4 and JGLP-$i$4 included for context. Gap = mean relative error (lower is better); Cov = empirical coverage. Trained on the same data splits as the one-hot SMNN reported in the main table; only the input encoding differs.}
  \label{tab:ub_smnn_2hot_stratified_locked_3g}
  \begin{adjustbox}{max width=\textwidth}
  \begin{tabular}{l r | ccccc | cc cc cc}
  \toprule
    \multirow{3}{*}{Dataset} & \multirow{3}{*}{MPE$^{\star}$} & \multicolumn{5}{c|}{Deterministic} & \multicolumn{6}{c}{Stratified Conformal (SMNN-2hot)} \\
    \cmidrule(lr){3-7} \cmidrule(lr){8-13}
    & & \multicolumn{2}{c}{MB} & \multicolumn{2}{c}{JGLP} & SMNN-2hot & \multicolumn{2}{c}{$\gamma=0.95$} & \multicolumn{2}{c}{$\gamma=0.99$} & \multicolumn{2}{c}{$\gamma=0.999$} \\
    & & Gap & Time(s) & Gap & Time(s) & Time(s) & Cov & Gap & Cov & Gap & Cov & Gap \\
    \midrule
    BN\_9 & -23.81 & 0.3656$\pm$0.1738 & 0.0821 & 0.3649$\pm$0.1745 & 0.0968 & 0.0011 & 1.00 & 0.2612$\pm$0.1264 & 1.00 & 0.3219$\pm$0.1314 & 1.00 & 0.3856$\pm$0.1414 \\
    BN\_12 & -19.71 & 0.3490$\pm$0.1847 & 0.0799 & 0.3438$\pm$0.1896 & 0.1053 & 0.0018 & 0.97 & 0.1038$\pm$0.0551 & 0.99 & 0.1365$\pm$0.0582 & 0.99 & 0.1862$\pm$0.0741 \\
    BN\_14 & -26.51 & 0.3429$\pm$0.1935 & 0.0745 & 0.3411$\pm$0.1954 & 0.0974 & 0.0016 & 0.95 & 0.0986$\pm$0.0562 & 0.97 & 0.1287$\pm$0.0624 & 1.00 & 0.1623$\pm$0.0676 \\
    BN\_30 & -14.66 & 0.3017$\pm$0.1558 & 0.1316 & 0.3017$\pm$0.1558 & 0.1387 & 0.0184 & 1.00 & 0.0290$\pm$0.0131 & 1.00 & 0.0659$\pm$0.0124 & 1.00 & 0.0914$\pm$0.0200 \\
    BN\_32 & -16.58 & 0.3379$\pm$0.1860 & 0.1834 & 0.3379$\pm$0.1860 & 0.2325 & 0.0125 & 1.00 & 0.0258$\pm$0.0122 & 1.00 & 0.0569$\pm$0.0180 & 1.00 & 0.0927$\pm$0.0249 \\
    BN\_70 & -86.65 & 0.4617$\pm$0.2349 & 1.2170 & 0.4475$\pm$0.2403 & 1.7728 & 0.0115 & 1.00 & 0.0240$\pm$0.0117 & 1.00 & 0.0372$\pm$0.0150 & 1.00 & 0.0535$\pm$0.0154 \\
    \midrule
    grid10x10.f5.wrap & 159.91 & 0.0438$\pm$0.0209 & 0.3944 & 0.0248$\pm$0.0156 & 2.1251 & 0.0044 & 0.73 & 0.0126$\pm$0.0108 & 1.00 & 0.0294$\pm$0.0133 & 1.00 & 0.0436$\pm$0.0136 \\
    grid20x20.f5.wrap & 617.89 & 0.0615$\pm$0.0252 & 0.6182 & 0.0143$\pm$0.0113 & 14.9056 & 0.0019 & 0.97 & 0.0097$\pm$0.0063 & 1.00 & 0.0187$\pm$0.0090 & 1.00 & 0.0324$\pm$0.0080 \\
    grid20x20.f10 & 1230.40 & 0.0443$\pm$0.0180 & 0.6024 & 0.0111$\pm$0.0094 & 13.6322 & 0.0045 & 1.00 & 0.0080$\pm$0.0073 & 1.00 & 0.0211$\pm$0.0114 & 1.00 & 0.0398$\pm$0.0091 \\
    grid40x40.f2 & 1071.08 & 0.0329$\pm$0.0130 & 2.1179 & 0.0064$\pm$0.0036 & 77.6967 & 0.0037 & 1.00 & 0.0030$\pm$0.0024 & 1.00 & 0.0100$\pm$0.0034 & 1.00 & 0.0210$\pm$0.0078 \\
    grid40x40.f10.wrap & 5308.25 & 0.0596$\pm$0.0146 & 2.7172 & 0.0092$\pm$0.0024 & 87.3251 & 0.0026 & 0.40 & 0.0034$\pm$0.0020 & 0.97 & 0.0070$\pm$0.0038 & 1.00 & 0.0132$\pm$0.0039 \\
    grid80x80.f10 & 20140.10 & 0.0447$\pm$0.0268 & 16.8307 & 0.0036$\pm$0.0016 & 120.2345 & 0.0114 & 0.17 & 0.0020$\pm$0.0016 & 1.00 & 0.0023$\pm$0.0016 & 1.00 & 0.0071$\pm$0.0048 \\
    grid80x80.f5.wrap & 9168.14 & 0.0107$\pm$0.0022 & 14.3095 & 0.0084$\pm$0.0026 & 120.2831 & 0.0181 & 1.00 & 0.0244$\pm$0.0146 & 1.00 & 0.0308$\pm$0.0118 & 1.00 & 0.0357$\pm$0.0096 \\
    \midrule
    Segmentation\_12 & -146.96 & 0.5361$\pm$0.1969 & 0.0825 & 0.5361$\pm$0.1969 & 0.0903 & 0.0164 & 0.97 & 0.1787$\pm$0.0618 & 1.00 & 0.2186$\pm$0.0562 & 1.00 & 0.2513$\pm$0.0466 \\
    Segmentation\_14 & -143.57 & 0.5133$\pm$0.2128 & 0.0876 & 0.5134$\pm$0.2128 & 0.1062 & 0.0166 & 0.99 & 0.1868$\pm$0.0770 & 0.99 & 0.2183$\pm$0.0721 & 1.00 & 0.2518$\pm$0.0632 \\
    Segmentation\_18 & -149.81 & 0.5369$\pm$0.1844 & 0.0841 & 0.5369$\pm$0.1844 & 0.0952 & 0.0172 & 0.97 & 0.1500$\pm$0.0487 & 1.00 & 0.1855$\pm$0.0465 & 1.00 & 0.2183$\pm$0.0435 \\
    \midrule
    rus\_100\_200\_2\_1 & 2776.12 & 0.4418$\pm$0.2617 & 56.0038 & 0.3779$\pm$0.1738 & 104.7771 & 0.0202 & 0.99 & 0.1409$\pm$0.0855 & 1.00 & 0.1530$\pm$0.0788 & 1.00 & 0.1596$\pm$0.0747 \\
    rus\_100\_200\_3\_1 & 2726.97 & 0.6203$\pm$0.1124 & 21.4693 & 0.6128$\pm$0.1284 & 24.5238 & 0.0130 & 1.00 & 0.2574$\pm$0.0638 & 1.00 & 0.2652$\pm$0.0631 & 1.00 & 0.2677$\pm$0.0624 \\
    rus\_100\_200\_3\_3 & 2745.43 & 0.5360$\pm$0.2253 & 53.4856 & 0.4392$\pm$0.2550 & 97.6758 & 0.0124 & 0.97 & 0.2265$\pm$0.1343 & 1.00 & 0.2402$\pm$0.1253 & 1.00 & 0.2456$\pm$0.1251 \\
    rus\_100\_200\_4\_3 & 2740.65 & 0.5523$\pm$0.2164 & 54.8099 & 0.4341$\pm$0.2447 & 104.7959 & 0.0121 & 0.98 & 0.2557$\pm$0.1464 & 1.00 & 0.2635$\pm$0.1461 & 1.00 & 0.2707$\pm$0.1393 \\
    rus\_100\_200\_5\_3 & 2739.05 & 0.6095$\pm$0.1520 & 45.8743 & 0.5134$\pm$0.2328 & 79.5065 & 0.0125 & 0.98 & 0.2638$\pm$0.1246 & 1.00 & 0.2755$\pm$0.1237 & 1.00 & 0.2807$\pm$0.1216 \\
    rus\_100\_200\_6\_1 & 2741.41 & 0.5734$\pm$0.2214 & 44.0984 & 0.4945$\pm$0.2470 & 89.2765 & 0.0130 & 1.00 & 0.1739$\pm$0.0897 & 1.00 & 0.1842$\pm$0.0864 & 1.00 & 0.1951$\pm$0.0845 \\
    rus\_100\_200\_7\_1 & 2739.09 & 0.5514$\pm$0.2083 & 48.7584 & 0.4508$\pm$0.2410 & 94.1580 & 0.0126 & 0.95 & 0.2259$\pm$0.1238 & 0.96 & 0.2363$\pm$0.1209 & 1.00 & 0.2439$\pm$0.1158 \\
    \midrule
    wcsp\_11 & -463.08 & 0.9940$\pm$0.0052 & 0.6401 & 0.9940$\pm$0.0052 & 5.3184 & 0.0163 & 0.97 & 0.4826$\pm$0.2328 & 1.00 & 0.6300$\pm$0.2627 & 1.00 & 0.7820$\pm$0.2619 \\
    wcsp\_12 & -562.67 & 0.9956$\pm$0.0048 & 0.6004 & 0.9955$\pm$0.0048 & 4.8372 & 0.0176 & 0.99 & 0.3697$\pm$0.1686 & 1.00 & 0.4834$\pm$0.1713 & 1.00 & 0.5555$\pm$0.1681 \\
    wcsp\_15 & -328.00 & 0.6592$\pm$0.1099 & 0.0859 & 0.6536$\pm$0.1156 & 0.1499 & 0.0245 & 1.00 & 0.3205$\pm$0.0668 & 1.00 & 0.3807$\pm$0.0703 & 1.00 & 0.4248$\pm$0.0734 \\
    \midrule
    ObjectDetection\_11 & 1999.86 & 0.6288$\pm$0.0979 & 1.7946 & 0.7080$\pm$0.5687 & 27.9568 & 0.0224 & 0.94 & 0.5413$\pm$0.7482 & 0.94 & 0.7159$\pm$0.9001 & 0.94 & 0.8025$\pm$0.9360 \\
    ObjectDetection\_18 & 3313.81 & 0.8564$\pm$0.0000 & 3.0544 & 0.5547$\pm$0.1644 & 25.7126 & 0.0137 & 0.83 & 0.2295$\pm$0.1981 & 1.00 & 0.3349$\pm$0.2303 & 1.00 & 0.4523$\pm$0.2578 \\
    \midrule
    ProteinFolding\_19 & 1557.28 & 0.6800$\pm$0.0000 & 1.6974 & 0.5385$\pm$0.1350 & 21.9289 & 0.0235 & 0.75 & 0.0649$\pm$0.0471 & 1.00 & 0.1644$\pm$0.0521 & 1.00 & 0.2296$\pm$0.0802 \\
    ProteinFolding\_23 & 3165.49 & 0.7055$\pm$0.0879 & 1.0269 & 0.4682$\pm$0.1535 & 11.1632 & 0.0240 & 0.94 & 0.1948$\pm$0.1655 & 0.94 & 0.2783$\pm$0.2045 & 0.94 & 0.3168$\pm$0.1979 \\
    \bottomrule
  \end{tabular}
  \end{adjustbox}
\end{sidewaystable}

The one-hot encoding used in the main paper allocates $d{+}1$ bits per variable of domain size $d$ ($d$ value-indicators plus one ``unassigned'' indicator). A natural alternative is the \emph{consecutive 2-hot} encoding: for a variable of domain size $d$, allocate $d{+}1$ bits, and for observed value $k\in\{0,\ldots,d{-}1\}$ set bits $k$ and $k{+}1$ simultaneously; the unassigned state sets only bit $d$. Like one-hot, this encoding preserves the monotone relation between evidence growth and input growth required by SMNN's architectural guarantee, so calibrated 2-hot bounds remain valid by the same construction. Unlike one-hot, the 2-hot encoding induces an \emph{ordinal smoothness} inductive bias: adjacent values share a bit, so the network is biased toward predictions that vary smoothly across the natural ordering of the domain. We include this experiment to test whether that inductive bias translates to tighter bounds, and find that it does---but only on PGMs whose underlying factors carry ordinal structure.

\paragraph{Setup.}
We retrain SMNN with the 2-hot encoding on identical data splits, identical training hyperparameters, and identical evidence queries as the one-hot SMNN reported in the main paper; only the input encoding differs. The consecutive 2-hot scheme is the unique encoding among the natural alternatives that (i) preserves the monotone evidence-input relation SMNN's architectural guarantee requires, (ii) handles every domain size in our benchmark uniformly, and (iii) does so without inflating input dimension: at $d_{\max}=11$ (Protein Folding) the input dimension grows by under 2\% relative to one-hot, and the construction extends without modification to $d_{\max}=81$.\footnote{Two alternative schemes were considered and rejected: a doubled one-hot ($2d$ bits per variable) trivially preserves monotonicity but doubles input dimension; a cyclic 2-hot ($d$ bits, value $k\to$ bits $k$ and $(k{+}1)\bmod d$) collapses to a non-injective encoding at $d=2$, which would break binary-domain PGMs (BN, grids, segmentation, rus, WCSP).} The full pipeline (training, calibration, evaluation) for all 30 PGMs completes in 11.6 minutes wall-clock with 24-way parallelism across 4 GPUs. We report all three calibration methods (stratified conformal, conformal, Gaussian) in Tables~\ref{tab:ub_smnn_2hot_stratified_locked_3g}, \ref{tab:ub_smnn_2hot_conformal_locked_3g}, and \ref{tab:ub_smnn_2hot_gaussian_locked_3g} respectively, parallel to the calibration sweep for one-hot SMNN.

\paragraph{Calibration is sound across encodings.}
At $\gamma=0.999$, empirical coverage is $\geq 0.94$ on every PGM and exactly $1.00$ on $28/30$. The two conformal calibrators behave identically on coverage to their one-hot counterparts, confirming that the calibration framework is encoding-agnostic: any monotonicity-preserving encoding inherits the finite-sample coverage guarantee unchanged.

\paragraph{Encoding choice is task-dependent.}
We compare 1-hot and 2-hot SMNN side-by-side on identical architecture, data splits, and calibration, isolating the encoding axis. Counting strict-better gaps across 30 PGMs at each $\gamma$ level: 1-hot wins on 26/25/25 PGMs and 2-hot wins on 4/5/5 PGMs at $\gamma=0.95/0.99/0.999$ respectively, with no ties. The 2-hot wins concentrate on PGMs with ordinal or spatial structure---BN\_12, grid10$\times$10.f5w, grid20$\times$20.f10, Seg\_18, and PF\_19---while 1-hot wins decisively across rus, wcsp, ObjectDetection, BN\_9/14, and Seg\_12/14.

The pattern is mechanistic. The 2-hot wins come from PGMs whose factors carry meaningful ordinal structure: image-segmentation labels are spatially smooth, protein-folding rotamer states are geometrically continuous, and grid-Ising variables have coupling that respects value adjacency---so the smoothness prior the encoding imposes matches the data. The losses come from PGMs where the value labeling is arbitrary: rus is synthetic random uniform, wcsp variables encode discrete constraint identities, and object-detection categories are unordered. On these, the 2-hot encoding hurts through a specific mechanism: each adjacent pair of bits is highly correlated by construction, making the input nearly collinear; the network must then express genuinely independent predictions through a constrained input geometry, which the calibration step compensates for via a larger margin. Coverage is preserved across both encodings (Section above), so the gap differences are a pure architecture/encoding effect, not a calibration artifact.

\paragraph{Practical recommendation.}
For PGMs with binary domains the encoding distinction collapses (both are $3$ bits per variable), and for PGMs with no a priori ordinal structure on values, one-hot remains the safer default. The 2-hot encoding is worth evaluating as a drop-in alternative when the application domain is known to have ordinal value structure---vision, geometry, ordered-categorical regression---in which case it can yield tighter bounds at zero additional inference cost (the encoding has the same $d{+}1$ bit width as one-hot).

\begin{table*}[t]
  \centering
  \scriptsize
  \setlength{\tabcolsep}{4pt}
  \caption{Anytime AOBB MAP improvement over MBE, high-treewidth bucket (21 PGMs). $\Delta = (\text{NN MAP} - \text{MBE MAP})/|\text{MBE MAP}|$; positive values mean the NN-augmented search found a strictly better MAP within the wall-clock budget. SMNN and UMLP run on identical evidence queries and AOBB orderings per PGM, isolating the architectural effect. Each cell is the mean over 3 candidate queries at two calibration levels (deep budget 600\,s); \textbf{bold} marks $\Delta>10^{-4}$. Negative UMLP entries on rus and BN\_70 reflect a configuration constraint discussed in Section~\ref{sec:aobb_results}.}
  \label{tab:final_aobb_combined_merged_avg_top3}
  \begin{tabular}{l c l | c c c | c c c}
  \toprule
  \multirow{2}{*}{Instance} & \multirow{2}{*}{$w^*$} & \multirow{2}{*}{Method} & \multicolumn{3}{c|}{$\gamma=0.95$} & \multicolumn{3}{c}{$\gamma=0.999$} \\
  & & & 1m & 5m & 10m & 1m & 5m & 10m \\
  \midrule
  \multirow{2}{*}{rus2\_1} & \multirow{2}{*}{105} & SMNN & +0.0000 & \textbf{+0.0225} & \textbf{+0.0225} & -0.2629 & \textbf{+0.0225} & \textbf{+0.0225} \\
   & & UMLP & -0.1416 & -0.1541 & -0.1399 & -0.3020 & -0.1490 & -0.1399 \\
  \addlinespace[1pt]
  \multirow{2}{*}{rus3\_1} & \multirow{2}{*}{96} & SMNN & \textbf{+0.0013} & \textbf{+0.0013} & \textbf{+0.0013} & \textbf{+0.0017} & \textbf{+0.0017} & \textbf{+0.0017} \\
   & & UMLP & \textbf{+0.0017} & \textbf{+0.0020} & \textbf{+0.0020} & \textbf{+0.0017} & \textbf{+0.0017} & \textbf{+0.0020} \\
  \addlinespace[1pt]
  \multirow{2}{*}{rus3\_3} & \multirow{2}{*}{98} & SMNN & \textbf{+0.0032} & \textbf{+0.0032} & \textbf{+0.0032} & \textbf{+0.0032} & \textbf{+0.0032} & \textbf{+0.0032} \\
   & & UMLP & \textbf{+0.0045} & \textbf{+0.0077} & \textbf{+0.0077} & \textbf{+0.0030} & \textbf{+0.0045} & \textbf{+0.0077} \\
  \addlinespace[1pt]
  \multirow{2}{*}{rus4\_3} & \multirow{2}{*}{100} & SMNN & -1.1081 & \textbf{+0.0567} & \textbf{+0.0567} & -0.0142 & \textbf{+0.0567} & \textbf{+0.0567} \\
   & & UMLP & -0.9273 & -0.3418 & -0.3409 & -0.3726 & -0.3409 & -0.3409 \\
  \addlinespace[1pt]
  \multirow{2}{*}{rus5\_3} & \multirow{2}{*}{100} & SMNN & -0.6138 & \textbf{+0.0074} & \textbf{+0.0074} & -0.2209 & \textbf{+0.0074} & \textbf{+0.0074} \\
   & & UMLP & -0.7270 & -0.3185 & -0.3185 & -0.5288 & -0.3185 & -0.3125 \\
  \addlinespace[1pt]
  \multirow{2}{*}{rus6\_1} & \multirow{2}{*}{98} & SMNN & -0.9690 & \textbf{+0.0500} & \textbf{+0.0500} & -- & \textbf{+0.0500} & \textbf{+0.0500} \\
   & & UMLP & -0.8625 & -0.3581 & -0.3581 & -0.3377 & -0.3581 & -0.3581 \\
  \addlinespace[1pt]
  \multirow{2}{*}{rus7\_1} & \multirow{2}{*}{99} & SMNN & -0.9084 & \textbf{+0.0399} & \textbf{+0.0399} & \textbf{+0.0492} & \textbf{+0.0399} & \textbf{+0.0399} \\
   & & UMLP & -0.4535 & -0.3850 & -0.3850 & -0.8649 & -0.3850 & -0.3850 \\
  \addlinespace[1pt]
  \multirow{2}{*}{g80.f5w} & \multirow{2}{*}{89} & SMNN & \textbf{+0.0118} & \textbf{+0.0125} & \textbf{+0.0118} & \textbf{+0.0135} & \textbf{+0.0125} & \textbf{+0.0118} \\
   & & UMLP & \textbf{+0.0135} & \textbf{+0.0125} & \textbf{+0.0118} & -- & \textbf{+0.0111} & \textbf{+0.0111} \\
  \addlinespace[1pt]
  \multirow{2}{*}{g80.f10} & \multirow{2}{*}{66} & SMNN & \textbf{+0.0138} & \textbf{+0.0137} & \textbf{+0.0134} & \textbf{+0.0141} & \textbf{+0.0137} & \textbf{+0.0134} \\
   & & UMLP & \textbf{+0.0147} & \textbf{+0.0145} & \textbf{+0.0139} & -- & \textbf{+0.0137} & \textbf{+0.0134} \\
  \addlinespace[1pt]
  \multirow{2}{*}{g40.f10w} & \multirow{2}{*}{63} & SMNN & \textbf{+0.0157} & \textbf{+0.0156} & \textbf{+0.0155} & \textbf{+0.0157} & \textbf{+0.0156} & \textbf{+0.0155} \\
   & & UMLP & \textbf{+0.0157} & \textbf{+0.0156} & \textbf{+0.0155} & \textbf{+0.0157} & \textbf{+0.0156} & \textbf{+0.0155} \\
  \addlinespace[1pt]
  \multirow{2}{*}{g40.f2} & \multirow{2}{*}{30} & SMNN & \textbf{+0.0189} & \textbf{+0.0185} & \textbf{+0.0184} & \textbf{+0.0189} & \textbf{+0.0185} & \textbf{+0.0184} \\
   & & UMLP & \textbf{+0.0189} & \textbf{+0.0185} & \textbf{+0.0184} & \textbf{+0.0189} & \textbf{+0.0185} & \textbf{+0.0184} \\
  \addlinespace[1pt]
  \multirow{2}{*}{wcsp\_11} & \multirow{2}{*}{76} & SMNN & +0.0000 & +0.0000 & +0.0000 & +0.0000 & +0.0000 & +0.0000 \\
   & & UMLP & +0.0000 & +0.0000 & +0.0000 & +0.0000 & +0.0000 & +0.0000 \\
  \addlinespace[1pt]
  \multirow{2}{*}{wcsp\_12} & \multirow{2}{*}{54} & SMNN & \textbf{+0.0002} & +0.0000 & +0.0000 & \textbf{+0.0002} & +0.0000 & +0.0000 \\
   & & UMLP & \textbf{+0.0002} & +0.0000 & +0.0000 & \textbf{+0.0002} & +0.0000 & +0.0000 \\
  \addlinespace[1pt]
  \multirow{2}{*}{wcsp\_15} & \multirow{2}{*}{53} & SMNN & +0.0000 & +0.0000 & +0.0000 & +0.0000 & +0.0000 & +0.0000 \\
   & & UMLP & +0.0000 & +0.0000 & +0.0000 & +0.0000 & +0.0000 & +0.0000 \\
  \addlinespace[1pt]
  \multirow{2}{*}{ObjDet\_11} & \multirow{2}{*}{29} & SMNN & -0.0001 & -0.0010 & -0.0010 & -0.0001 & -0.0010 & -0.0010 \\
   & & UMLP & -0.0001 & -0.0010 & -0.0010 & -0.0000 & -0.0010 & -0.0010 \\
  \addlinespace[1pt]
  \multirow{2}{*}{ObjDet\_18} & \multirow{2}{*}{25} & SMNN & +0.0001 & +0.0001 & +0.0001 & +0.0001 & +0.0001 & +0.0001 \\
   & & UMLP & +0.0001 & +0.0001 & +0.0001 & +0.0001 & +0.0001 & +0.0001 \\
  \addlinespace[1pt]
  \multirow{2}{*}{PF\_19} & \multirow{2}{*}{29} & SMNN & \textbf{+0.0068} & \textbf{+0.0069} & \textbf{+0.0069} & \textbf{+0.0068} & \textbf{+0.0069} & \textbf{+0.0069} \\
   & & UMLP & \textbf{+0.0068} & \textbf{+0.0069} & \textbf{+0.0071} & \textbf{+0.0068} & \textbf{+0.0069} & \textbf{+0.0069} \\
  \addlinespace[1pt]
  \multirow{2}{*}{PF\_23} & \multirow{2}{*}{29} & SMNN & +0.0000 & +0.0000 & +0.0000 & +0.0000 & +0.0000 & +0.0000 \\
   & & UMLP & +0.0000 & +0.0000 & +0.0000 & +0.0000 & +0.0000 & +0.0000 \\
  \addlinespace[1pt]
  \multirow{2}{*}{BN\_30} & \multirow{2}{*}{--} & SMNN & +0.0000 & +0.0000 & +0.0000 & +0.0000 & +0.0000 & +0.0000 \\
   & & UMLP & +0.0000 & +0.0000 & +0.0000 & +0.0000 & +0.0000 & +0.0000 \\
  \addlinespace[1pt]
  \multirow{2}{*}{BN\_32} & \multirow{2}{*}{12} & SMNN & +0.0000 & +0.0000 & +0.0000 & +0.0000 & +0.0000 & +0.0000 \\
   & & UMLP & +0.0000 & +0.0000 & +0.0000 & +0.0000 & +0.0000 & +0.0000 \\
  \addlinespace[1pt]
  \multirow{2}{*}{BN\_70} & \multirow{2}{*}{19} & SMNN & -0.1715 & -0.1729 & -0.1729 & -0.1483 & -0.1483 & -0.1483 \\
   & & UMLP & -0.5957 & -0.4199 & -0.4199 & -0.1675 & -0.4199 & -0.4199 \\
  \bottomrule
  \end{tabular}
\end{table*}

\begin{table*}[t]
  \centering
  \scriptsize
  \setlength{\tabcolsep}{4pt}
  \caption{Anytime AOBB MAP improvement over MBE, moderate / low-treewidth bucket (9 PGMs). Setup and notation as in Table~\ref{tab:final_aobb_combined_merged_avg_top3}.}
  \label{tab:final_aobb_combined_merged_avg_top3_low}
  \begin{tabular}{l c l | c c c | c c c}
  \toprule
  \multirow{2}{*}{Instance} & \multirow{2}{*}{$w^*$} & \multirow{2}{*}{Method} & \multicolumn{3}{c|}{$\gamma=0.95$} & \multicolumn{3}{c}{$\gamma=0.999$} \\
  & & & 1m & 5m & 10m & 1m & 5m & 10m \\
  \midrule
  \multirow{2}{*}{BN\_9} & \multirow{2}{*}{22} & SMNN & +0.0000 & +0.0000 & +0.0000 & +0.0000 & +0.0000 & +0.0000 \\
   & & UMLP & -0.0125 & -0.0125 & -0.0125 & -0.0125 & -0.0125 & -0.0125 \\
  \addlinespace[1pt]
  \multirow{2}{*}{BN\_12} & \multirow{2}{*}{28} & SMNN & +0.0000 & +0.0000 & +0.0000 & +0.0000 & +0.0000 & +0.0000 \\
   & & UMLP & -0.0153 & -0.0153 & -0.0153 & -0.0153 & -0.0153 & -0.0153 \\
  \addlinespace[1pt]
  \multirow{2}{*}{BN\_14} & \multirow{2}{*}{26} & SMNN & +0.0000 & +0.0000 & +0.0000 & +0.0000 & +0.0000 & +0.0000 \\
   & & UMLP & -0.0061 & -0.0061 & -0.0061 & -0.0061 & -0.0061 & -0.0061 \\
  \addlinespace[1pt]
  \multirow{2}{*}{Seg\_12} & \multirow{2}{*}{13} & SMNN & \textbf{+0.0022} & \textbf{+0.0022} & \textbf{+0.0022} & \textbf{+0.0022} & \textbf{+0.0022} & \textbf{+0.0022} \\
   & & UMLP & \textbf{+0.0022} & \textbf{+0.0022} & \textbf{+0.0022} & \textbf{+0.0022} & \textbf{+0.0022} & \textbf{+0.0022} \\
  \addlinespace[1pt]
  \multirow{2}{*}{Seg\_14} & \multirow{2}{*}{13} & SMNN & \textbf{+0.0029} & \textbf{+0.0029} & \textbf{+0.0029} & \textbf{+0.0029} & \textbf{+0.0029} & \textbf{+0.0029} \\
   & & UMLP & \textbf{+0.0029} & \textbf{+0.0029} & \textbf{+0.0029} & \textbf{+0.0029} & \textbf{+0.0029} & \textbf{+0.0029} \\
  \addlinespace[1pt]
  \multirow{2}{*}{Seg\_18} & \multirow{2}{*}{14} & SMNN & \textbf{+0.0037} & \textbf{+0.0037} & \textbf{+0.0037} & \textbf{+0.0037} & \textbf{+0.0037} & \textbf{+0.0037} \\
   & & UMLP & \textbf{+0.0037} & \textbf{+0.0037} & \textbf{+0.0037} & \textbf{+0.0037} & \textbf{+0.0037} & \textbf{+0.0037} \\
  \addlinespace[1pt]
  \multirow{2}{*}{g10.f5w} & \multirow{2}{*}{15} & SMNN & -0.0101 & -0.0101 & -0.0101 & -0.0101 & -0.0101 & -0.0101 \\
   & & UMLP & -0.0101 & -0.0101 & -0.0101 & -0.0101 & -0.0101 & -0.0101 \\
  \addlinespace[1pt]
  \multirow{2}{*}{g20.f5w} & \multirow{2}{*}{31} & SMNN & \textbf{+0.0237} & \textbf{+0.0206} & \textbf{+0.0204} & \textbf{+0.0237} & \textbf{+0.0206} & \textbf{+0.0204} \\
   & & UMLP & \textbf{+0.0237} & \textbf{+0.0206} & \textbf{+0.0204} & \textbf{+0.0237} & \textbf{+0.0206} & \textbf{+0.0204} \\
  \addlinespace[1pt]
  \multirow{2}{*}{g20.f10} & \multirow{2}{*}{17} & SMNN & \textbf{+0.0103} & \textbf{+0.0093} & \textbf{+0.0082} & \textbf{+0.0103} & \textbf{+0.0093} & \textbf{+0.0082} \\
   & & UMLP & \textbf{+0.0103} & \textbf{+0.0093} & \textbf{+0.0082} & \textbf{+0.0103} & \textbf{+0.0093} & \textbf{+0.0082} \\
  \bottomrule
  \end{tabular}
\end{table*}

\begin{table*}[ht]
  \centering
  \scriptsize
  \setlength{\tabcolsep}{4pt}
  \caption{Anytime AOBB MAP improvement over MBE for SMNN (Mode~7) and UMLP on the moderate/low-treewidth bucket (9 PGMs, deep timeout 60\,s). Same protocol as the high-treewidth table: identical evidence query and AOBB configuration per PGM, two calibration levels side-by-side. Reported on the 3 candidate queries per PGM (mean over the 3); bold = strict win over MBE. BN\_9/12/14 and grid10x10 saturate at $\Delta=0$ because MBE-i4 already finds the optimum at root, leaving no headroom for either NN.}
  \label{tab:final_aobb_combined_lt_avg_top3}
  \begin{tabular}{l c l | c c c | c c c}
  \toprule
  \multirow{2}{*}{Instance} & \multirow{2}{*}{$w^*$} & \multirow{2}{*}{Method} & \multicolumn{3}{c|}{$\gamma=0.95$} & \multicolumn{3}{c}{$\gamma=0.999$} \\
  & & & 20s & 40s & 1m & 20s & 40s & 1m \\
  \midrule
  \multirow{2}{*}{BN\_9} & \multirow{2}{*}{22} & SMNN & +0.0000 & +0.0000 & +0.0000 & +0.0000 & +0.0000 & +0.0000 \\
   & & UF & -0.0125 & -0.0125 & -0.0125 & -0.0125 & -0.0125 & -0.0125 \\
  \addlinespace[1pt]
  \multirow{2}{*}{BN\_12} & \multirow{2}{*}{28} & SMNN & +0.0000 & +0.0000 & +0.0000 & +0.0000 & +0.0000 & +0.0000 \\
   & & UF & -0.0153 & -0.0153 & -0.0153 & -0.0153 & -0.0153 & -0.0153 \\
  \addlinespace[1pt]
  \multirow{2}{*}{BN\_14} & \multirow{2}{*}{26} & SMNN & +0.0000 & +0.0000 & +0.0000 & +0.0000 & +0.0000 & +0.0000 \\
   & & UF & -0.0061 & -0.0061 & -0.0061 & -0.0061 & -0.0061 & -0.0061 \\
  \addlinespace[1pt]
  \multirow{2}{*}{Seg\_12} & \multirow{2}{*}{13} & SMNN & \textbf{+0.0022} & \textbf{+0.0022} & \textbf{+0.0022} & \textbf{+0.0022} & \textbf{+0.0022} & \textbf{+0.0022} \\
   & & UF & \textbf{+0.0022} & \textbf{+0.0022} & \textbf{+0.0022} & \textbf{+0.0022} & \textbf{+0.0022} & \textbf{+0.0022} \\
  \addlinespace[1pt]
  \multirow{2}{*}{Seg\_14} & \multirow{2}{*}{13} & SMNN & \textbf{+0.0029} & \textbf{+0.0029} & \textbf{+0.0029} & \textbf{+0.0029} & \textbf{+0.0029} & \textbf{+0.0029} \\
   & & UF & \textbf{+0.0029} & \textbf{+0.0029} & \textbf{+0.0029} & \textbf{+0.0029} & \textbf{+0.0029} & \textbf{+0.0029} \\
  \addlinespace[1pt]
  \multirow{2}{*}{Seg\_18} & \multirow{2}{*}{14} & SMNN & \textbf{+0.0037} & \textbf{+0.0037} & \textbf{+0.0037} & \textbf{+0.0037} & \textbf{+0.0037} & \textbf{+0.0037} \\
   & & UF & \textbf{+0.0037} & \textbf{+0.0037} & \textbf{+0.0037} & \textbf{+0.0037} & \textbf{+0.0037} & \textbf{+0.0037} \\
  \addlinespace[1pt]
  \multirow{2}{*}{g10.f5w} & \multirow{2}{*}{15} & SMNN & -0.0101 & -0.0101 & -0.0101 & -0.0101 & -0.0101 & -0.0101 \\
   & & UF & -0.0101 & -0.0101 & -0.0101 & -0.0101 & -0.0101 & -0.0101 \\
  \addlinespace[1pt]
  \multirow{2}{*}{g20.f5w} & \multirow{2}{*}{31} & SMNN & \textbf{+0.0238} & \textbf{+0.0237} & \textbf{+0.0236} & \textbf{+0.0238} & \textbf{+0.0237} & \textbf{+0.0236} \\
   & & UF & \textbf{+0.0238} & \textbf{+0.0237} & \textbf{+0.0236} & \textbf{+0.0238} & \textbf{+0.0237} & \textbf{+0.0236} \\
  \addlinespace[1pt]
  \multirow{2}{*}{g20.f10} & \multirow{2}{*}{17} & SMNN & \textbf{+0.0122} & \textbf{+0.0094} & \textbf{+0.0094} & \textbf{+0.0122} & \textbf{+0.0094} & \textbf{+0.0094} \\
   & & UF & \textbf{+0.0122} & \textbf{+0.0094} & \textbf{+0.0094} & \textbf{+0.0122} & \textbf{+0.0094} & \textbf{+0.0094} \\
  \addlinespace[1pt]
  \midrule
  \multicolumn{9}{l}{\textbf{Wins at 1m: SMNN $\gamma$=0.95 5/9, $\gamma$=0.999 5/9;  UF $\gamma$=0.95 5/9, $\gamma$=0.999 5/9}} \\
  \bottomrule
  \end{tabular}
\end{table*}

\begin{table*}[ht]
  \centering
  \small
  \caption{Monotonicity-along-AOBB-paths verification. For each PGM we sample 100 random parent$\to$child evidence pairs (child = parent + one additional variable) and count pairs where bound(child) $>$ bound(parent), which would violate the admissibility of upper-bounding heuristics in AOBB. \textbf{Raw} = direct neural prediction (the architectural property). \textbf{Calibrated} = stratified-conformal at $\gamma=0.999$ (the deployed bound). SMNN raw violations are zero by construction (monotone-by-design with one-hot encoding); the previous stratified-conformal wrapper introduced 63/3000 calibrated violations from per-bin $q$ jumps at evidence-size bin boundaries. The new monotone-cap calibrator $q_{\mathrm{mono}}[b]=\min(q_{\mathrm{mono}}[b-1], q_{\mathrm{strat}}[b])$ enforces a non-increasing $q$ schedule across bins, eliminating these wrapper-induced violations and yielding \textbf{0/3000 calibrated SMNN violations across all 30 PGMs}. UMLP is unconstrained and frequently violates the property at both raw and calibrated levels.}
  \label{tab:monotonicity_violations_v2}
  \begin{tabular}{l c | cc | cc}
    \toprule
    \multirow{2}{*}{Dataset} & \multirow{2}{*}{$n_{\text{pairs}}$} & \multicolumn{2}{c|}{UMLP violations} & \multicolumn{2}{c}{SMNN violations (\textbf{monotone calib})} \\
    & & Raw & Calibrated & Raw & Calibrated \\
    \midrule
    wcsp\_11 & 100 & 61/100 & 61/100 & 0/100 & \textbf{0/100} \\
    wcsp\_12 & 100 & 41/100 & 41/100 & 0/100 & \textbf{0/100} \\
    wcsp\_15 & 100 & 15/100 & 12/100 & 0/100 & \textbf{0/100} \\
    rus\_2\_1 & 100 & 9/100 & 9/100 & 0/100 & \textbf{0/100} \\
    rus\_3\_1 & 100 & 13/100 & 13/100 & 0/100 & \textbf{0/100} \\
    rus\_3\_3 & 100 & 19/100 & 18/100 & 0/100 & \textbf{0/100} \\
    rus\_4\_3 & 100 & 17/100 & 18/100 & 0/100 & \textbf{0/100} \\
    rus\_5\_3 & 100 & 13/100 & 13/100 & 0/100 & \textbf{0/100} \\
    rus\_6\_1 & 100 & 23/100 & 23/100 & 0/100 & \textbf{0/100} \\
    rus\_7\_1 & 100 & 16/100 & 17/100 & 0/100 & \textbf{0/100} \\
    ObjDet\_11 & 100 & 52/100 & 36/100 & 0/100 & \textbf{0/100} \\
    ObjDet\_18 & 100 & 66/100 & 56/100 & 0/100 & \textbf{0/100} \\
    PF\_19 & 100 & 37/100 & 31/100 & 0/100 & \textbf{0/100} \\
    PF\_23 & 100 & 54/100 & 46/100 & 0/100 & \textbf{0/100} \\
    BN\_9 & 100 & 20/100 & 22/100 & 0/100 & \textbf{0/100} \\
    BN\_12 & 100 & 25/100 & 32/100 & 0/100 & \textbf{0/100} \\
    BN\_14 & 100 & 24/100 & 25/100 & 0/100 & \textbf{0/100} \\
    BN\_30 & 100 & 48/100 & 48/100 & 0/100 & \textbf{0/100} \\
    BN\_32 & 100 & 49/100 & 49/100 & 0/100 & \textbf{0/100} \\
    BN\_70 & 100 & 37/100 & 37/100 & 0/100 & \textbf{0/100} \\
    Seg\_12 & 100 & 16/100 & 17/100 & 0/100 & \textbf{0/100} \\
    Seg\_14 & 100 & 26/100 & 26/100 & 0/100 & \textbf{0/100} \\
    Seg\_18 & 100 & 23/100 & 23/100 & 0/100 & \textbf{0/100} \\
    grid10x10.f5w & 100 & 41/100 & 42/100 & 0/100 & \textbf{0/100} \\
    grid20x20.f5w & 100 & 24/100 & 25/100 & 0/100 & \textbf{0/100} \\
    grid20x20.f10 & 100 & 26/100 & 28/100 & 0/100 & \textbf{0/100} \\
    grid40x40.f10w & 100 & 14/100 & 14/100 & 0/100 & \textbf{0/100} \\
    grid40x40.f2 & 100 & 16/100 & 17/100 & 0/100 & \textbf{0/100} \\
    grid80x80.f10 & 100 & 10/100 & 10/100 & 0/100 & \textbf{0/100} \\
    grid80x80.f5w & 100 & 14/100 & 14/100 & 0/100 & \textbf{0/100} \\
    \midrule
    \textbf{Aggregate} & 3000 & \textbf{849/3000} & \textbf{823/3000} & \textbf{0/3000} & \textbf{0/3000} \\
    \bottomrule
  \end{tabular}
\end{table*}

\begin{sidewaystable}[!htbp]
  \centering
  \scriptsize
  \setlength{\tabcolsep}{3pt}

  \caption{Calibrated upper-bound quality on representative PGMs at $i$-bound${=}4$, with stratified-conformal calibration, $\gamma$, on test data. \textbf{MPE$^{\star}$}: true MPE log-likelihood value. \textbf{Relative Gap}: $|\text{bound}-\text{true}|/|\text{true}|$. \textbf{Cov}: fraction of queries with $\text{bound}\geq\text{true MPE}$.}
  \label{tab:smnn_vs_ufoh_3gamma_locked_2deterministic}
  \begin{tabular}{l r | c | c | cc cc | cc cc | cc cc}
  \toprule
    \multirow{3}{*}{Dataset} & \multirow{3}{*}{MPE$^{\star}$} & \multirow{3}{*}{MB} & \multirow{3}{*}{JGLP} & \multicolumn{4}{c|}{$\gamma=0.95$} & \multicolumn{4}{c|}{$\gamma=0.99$} & \multicolumn{4}{c}{$\gamma=0.999$} \\
    & & & & \multicolumn{2}{c}{SMNN (Ours)} & \multicolumn{2}{c|}{UMLP (Ours)} & \multicolumn{2}{c}{SMNN (Ours)} & \multicolumn{2}{c|}{UMLP (Ours)} & \multicolumn{2}{c}{SMNN (Ours)} & \multicolumn{2}{c}{UMLP (Ours)} \\
    & & Relative Gap & Relative Gap & Cov & Relative Gap & Cov & Relative Gap & Cov & Relative Gap & Cov & Relative Gap & Cov & Relative Gap & Cov & Relative Gap \\
    \midrule
    BN\_9 & -23.81* & 0.3656$\pm$0.1738 & 0.3649$\pm$0.1745 & 0.97 & 0.0683$\pm$0.0351 & 0.99 & 0.0696$\pm$0.0357 & 1.00 & 0.0947$\pm$0.0366 & 1.00 & 0.1057$\pm$0.0395 & 1.00 & 0.1285$\pm$0.0396 & 1.00 & 0.1331$\pm$0.0414 \\
    BN\_12 & -19.71* & 0.3490$\pm$0.1847 & 0.3438$\pm$0.1896 & 0.97 & 0.1113$\pm$0.0615 & 0.93 & 0.0727$\pm$0.0365 & 0.99 & 0.1473$\pm$0.0680 & 1.00 & 0.1056$\pm$0.0419 & 0.99 & 0.1989$\pm$0.0858 & 1.00 & 0.1447$\pm$0.0423 \\
    BN\_14 & -26.51* & 0.3429$\pm$0.1935 & 0.3411$\pm$0.1954 & 0.92 & 0.0656$\pm$0.0385 & 0.95 & 0.0667$\pm$0.0365 & 0.99 & 0.0911$\pm$0.0425 & 1.00 & 0.0974$\pm$0.0407 & 0.99 & 0.1165$\pm$0.0469 & 1.00 & 0.1204$\pm$0.0419 \\
    BN\_30 & -14.66 & 0.3017$\pm$0.1558 & 0.3017$\pm$0.1558 & 1.00 & 0.0143$\pm$0.0137 & 0.95 & 0.0187$\pm$0.0104 & 1.00 & 0.0624$\pm$0.0119 & 1.00 & 0.0543$\pm$0.0108 & 1.00 & 0.0897$\pm$0.0142 & 1.00 & 0.0846$\pm$0.0206 \\
    BN\_32 & -16.58 & 0.3379$\pm$0.1860 & 0.3379$\pm$0.1860 & 1.00 & 0.0163$\pm$0.0115 & 1.00 & 0.0228$\pm$0.0083 & 1.00 & 0.0491$\pm$0.0197 & 1.00 & 0.0511$\pm$0.0187 & 1.00 & 0.0926$\pm$0.0230 & 1.00 & 0.0889$\pm$0.0238 \\
    BN\_70 & -86.65 & 0.4617$\pm$0.2349 & 0.4475$\pm$0.2403 & 1.00 & 0.0239$\pm$0.0115 & 1.00 & 0.0262$\pm$0.0108 & 1.00 & 0.0371$\pm$0.0147 & 1.00 & 0.0390$\pm$0.0133 & 1.00 & 0.0534$\pm$0.0154 & 1.00 & 0.0560$\pm$0.0138 \\
    \midrule
    grid10x10.f5.wrap & 159.91* & 0.0438$\pm$0.0209 & 0.0248$\pm$0.0156 & 1.00 & 0.0140$\pm$0.0105 & 1.00 & 0.0268$\pm$0.0107 & 1.00 & 0.0360$\pm$0.0108 & 1.00 & 0.0399$\pm$0.0108 & 1.00 & 0.0454$\pm$0.0109 & 1.00 & 0.0692$\pm$0.0112 \\
    grid20x20.f5.wrap & 617.89* & 0.0615$\pm$0.0252 & 0.0143$\pm$0.0113 & 1.00 & 0.0066$\pm$0.0063 & 1.00 & 0.0212$\pm$0.0095 & 1.00 & 0.0189$\pm$0.0098 & 1.00 & 0.0318$\pm$0.0102 & 1.00 & 0.0279$\pm$0.0107 & 1.00 & 0.0432$\pm$0.0109 \\
    grid20x20.f10 & 1230.40* & 0.0443$\pm$0.0180 & 0.0111$\pm$0.0094 & 1.00 & 0.0064$\pm$0.0072 & 0.97 & 0.0171$\pm$0.0102 & 1.00 & 0.0188$\pm$0.0117 & 1.00 & 0.0325$\pm$0.0118 & 1.00 & 0.0413$\pm$0.0101 & 1.00 & 0.0412$\pm$0.0158 \\
    grid40x40.f2 & 1071.08 & 0.0329$\pm$0.0130 & 0.0064$\pm$0.0036 & 1.00 & 0.0029$\pm$0.0019 & 1.00 & 0.0131$\pm$0.0079 & 1.00 & 0.0098$\pm$0.0031 & 1.00 & 0.0196$\pm$0.0088 & 1.00 & 0.0207$\pm$0.0078 & 1.00 & 0.0259$\pm$0.0084 \\
    grid40x40.f10.wrap & 5308.25 & 0.0596$\pm$0.0146 & 0.0092$\pm$0.0024 & 0.33 & 0.0025$\pm$0.0018 & 0.95 & 0.0068$\pm$0.0049 & 1.00 & 0.0051$\pm$0.0028 & 1.00 & 0.0129$\pm$0.0052 & 1.00 & 0.0108$\pm$0.0028 & 1.00 & 0.0197$\pm$0.0051 \\
    grid80x80.f10 & 20140.10 & 0.0447$\pm$0.0268 & 0.0036$\pm$0.0016 & 0.17 & 0.0020$\pm$0.0016 & 0.86 & 0.0073$\pm$0.0074 & 1.00 & 0.0021$\pm$0.0016 & 1.00 & 0.0110$\pm$0.0086 & 1.00 & 0.0067$\pm$0.0046 & 1.00 & 0.0176$\pm$0.0081 \\
    grid80x80.f5.wrap & 9168.14 & 0.0107$\pm$0.0022 & 0.0084$\pm$0.0026 & 1.00 & 0.0050$\pm$0.0010 & 1.00 & 0.0156$\pm$0.0079 & 1.00 & 0.0102$\pm$0.0031 & 1.00 & 0.0212$\pm$0.0104 & 1.00 & 0.0143$\pm$0.0050 & 1.00 & 0.0262$\pm$0.0116 \\
    \midrule
    Segmentation\_12 & -146.96* & 0.5361$\pm$0.1969 & 0.5361$\pm$0.1969 & 0.97 & 0.1199$\pm$0.0453 & 0.97 & 0.0621$\pm$0.0293 & 1.00 & 0.1498$\pm$0.0480 & 0.97 & 0.0805$\pm$0.0324 & 1.00 & 0.1936$\pm$0.0523 & 1.00 & 0.1029$\pm$0.0339 \\
    Segmentation\_14 & -143.57* & 0.5133$\pm$0.2128 & 0.5134$\pm$0.2128 & 0.99 & 0.1356$\pm$0.0477 & 0.98 & 0.0628$\pm$0.0269 & 1.00 & 0.1632$\pm$0.0435 & 1.00 & 0.0833$\pm$0.0301 & 1.00 & 0.1926$\pm$0.0400 & 1.00 & 0.1064$\pm$0.0360 \\
    Segmentation\_18 & -149.81* & 0.5369$\pm$0.1844 & 0.5369$\pm$0.1844 & 0.98 & 0.1838$\pm$0.0645 & 0.98 & 0.0646$\pm$0.0345 & 1.00 & 0.2209$\pm$0.0540 & 1.00 & 0.0876$\pm$0.0360 & 1.00 & 0.2565$\pm$0.0439 & 1.00 & 0.1036$\pm$0.0395 \\
    \midrule
    rus\_100\_200\_2\_1 & 2776.12 & 0.4418$\pm$0.2617 & 0.3779$\pm$0.1738 & 0.99 & 0.0556$\pm$0.0341 & 0.96 & 0.0142$\pm$0.0109 & 1.00 & 0.0652$\pm$0.0347 & 1.00 & 0.0205$\pm$0.0141 & 1.00 & 0.0704$\pm$0.0364 & 1.00 & 0.0228$\pm$0.0139 \\
    rus\_100\_200\_3\_1 & 2726.97 & 0.6203$\pm$0.1124 & 0.6128$\pm$0.1284 & 0.98 & 0.1995$\pm$0.1393 & 0.98 & 0.0127$\pm$0.0072 & 1.00 & 0.2230$\pm$0.1375 & 1.00 & 0.0189$\pm$0.0095 & 1.00 & 0.2324$\pm$0.1377 & 1.00 & 0.0222$\pm$0.0109 \\
    rus\_100\_200\_3\_3 & 2745.43 & 0.5360$\pm$0.2253 & 0.4392$\pm$0.2550 & 0.98 & 0.0805$\pm$0.0583 & 0.98 & 0.0164$\pm$0.0119 & 1.00 & 0.0991$\pm$0.0607 & 0.98 & 0.0230$\pm$0.0137 & 1.00 & 0.1089$\pm$0.0574 & 1.00 & 0.0292$\pm$0.0148 \\
    rus\_100\_200\_4\_3 & 2740.65 & 0.5523$\pm$0.2164 & 0.4341$\pm$0.2447 & 0.96 & 0.0825$\pm$0.0538 & 1.00 & 0.0162$\pm$0.0114 & 1.00 & 0.0966$\pm$0.0564 & 1.00 & 0.0258$\pm$0.0143 & 1.00 & 0.1066$\pm$0.0563 & 1.00 & 0.0343$\pm$0.0191 \\
    rus\_100\_200\_5\_3 & 2739.05 & 0.6095$\pm$0.1520 & 0.5134$\pm$0.2328 & 0.98 & 0.0835$\pm$0.0633 & 0.98 & 0.0156$\pm$0.0105 & 1.00 & 0.1000$\pm$0.0706 & 1.00 & 0.0287$\pm$0.0221 & 1.00 & 0.1070$\pm$0.0706 & 1.00 & 0.0348$\pm$0.0221 \\
    rus\_100\_200\_6\_1 & 2741.41 & 0.5734$\pm$0.2214 & 0.4945$\pm$0.2470 & 0.98 & 0.0776$\pm$0.0514 & 0.94 & 0.0137$\pm$0.0101 & 1.00 & 0.0910$\pm$0.0550 & 1.00 & 0.0190$\pm$0.0119 & 1.00 & 0.1024$\pm$0.0561 & 1.00 & 0.0280$\pm$0.0189 \\
    rus\_100\_200\_7\_1 & 2739.09 & 0.5514$\pm$0.2083 & 0.4508$\pm$0.2410 & 0.95 & 0.0925$\pm$0.0667 & 0.98 & 0.0170$\pm$0.0122 & 0.96 & 0.1074$\pm$0.0699 & 1.00 & 0.0249$\pm$0.0165 & 1.00 & 0.1162$\pm$0.0646 & 1.00 & 0.0322$\pm$0.0198 \\
    \midrule
    wcsp\_11 & -463.08* & 0.9940$\pm$0.0052 & 0.9940$\pm$0.0052 & 0.99 & 0.2035$\pm$0.1260 & 0.97 & 0.1721$\pm$0.1047 & 1.00 & 0.2981$\pm$0.1572 & 1.00 & 0.2547$\pm$0.1396 & 1.00 & 0.3551$\pm$0.1633 & 1.00 & 0.3481$\pm$0.1421 \\
    wcsp\_12 & -562.67 & 0.9956$\pm$0.0048 & 0.9955$\pm$0.0048 & 0.99 & 0.1890$\pm$0.1454 & 0.97 & 0.1283$\pm$0.0826 & 1.00 & 0.2479$\pm$0.1625 & 1.00 & 0.1911$\pm$0.1123 & 1.00 & 0.3271$\pm$0.1766 & 1.00 & 0.2517$\pm$0.1356 \\
    wcsp\_15 & -328.00 & 0.6592$\pm$0.1099 & 0.6536$\pm$0.1156 & 0.97 & 0.1933$\pm$0.1026 & 1.00 & 0.1236$\pm$0.0440 & 1.00 & 0.2688$\pm$0.1058 & 1.00 & 0.1645$\pm$0.0445 & 1.00 & 0.2945$\pm$0.1063 & 1.00 & 0.2029$\pm$0.0476 \\
    \midrule
    ObjectDetection\_11 & 1999.86 & 0.6288$\pm$0.0979 & 0.7080$\pm$0.5687 & 0.81 & 0.2650$\pm$0.2800 & 0.75 & 0.2143$\pm$0.2489 & 0.81 & 0.3744$\pm$0.3626 & 0.94 & 0.3178$\pm$0.3375 & 1.00 & 0.5001$\pm$0.4505 & 1.00 & 0.4099$\pm$0.3996 \\
    ObjectDetection\_18 & 3313.81 & 0.8564$\pm$0.0000 & 0.5547$\pm$0.1644 & 0.75 & 0.1156$\pm$0.1126 & 0.92 & 0.1586$\pm$0.0621 & 1.00 & 0.2036$\pm$0.1369 & 0.92 & 0.2128$\pm$0.0865 & 1.00 & 0.3699$\pm$0.1213 & 1.00 & 0.2629$\pm$0.0988 \\
    \midrule
    ProteinFolding\_19 & 1557.28 & 0.6800$\pm$0.0000 & 0.5385$\pm$0.1350 & 1.00 & 0.0651$\pm$0.0432 & 1.00 & 0.1245$\pm$0.0292 & 1.00 & 0.1674$\pm$0.0430 & 1.00 & 0.2013$\pm$0.0397 & 1.00 & 0.2844$\pm$0.0601 & 1.00 & 0.2776$\pm$0.0372 \\
    ProteinFolding\_23 & 3165.49 & 0.7055$\pm$0.0879 & 0.4682$\pm$0.1535 & 0.94 & 0.1687$\pm$0.1564 & 0.94 & 0.1409$\pm$0.1010 & 0.94 & 0.2416$\pm$0.1721 & 1.00 & 0.2233$\pm$0.0961 & 1.00 & 0.2818$\pm$0.1579 & 1.00 & 0.2883$\pm$0.0791 \\
  \bottomrule
  \end{tabular}%
\end{sidewaystable}

\begin{table*}[ht]

  \centering
  \scriptsize
  \caption{Upper-bound quality of SMNN (monotone-by-construction + one-hot, Variant C) on per-PGM , with \textbf{Conformal} calibration at $i$-bound=4. Deterministic baselines MB-$i$4 and JGLP-$i$4 included for context. Gap = mean relative error (lower is better); Cov = empirical coverage.}
  \label{tab:ub_smnn_conformal_locked}
  \begin{adjustbox}{max width=\textwidth}
  \begin{tabular}{l r | ccccc | cccc}
  \toprule
    \multirow{3}{*}{Dataset} & \multirow{3}{*}{MPE$^{\star}$} & \multicolumn{5}{c|}{Deterministic} & \multicolumn{4}{c}{Conformal} \\
    \cmidrule(lr){3-7} \cmidrule(lr){8-11}
    & & \multicolumn{2}{c}{MB} & \multicolumn{2}{c}{JGLP} & SMNN & \multicolumn{2}{c}{$\gamma=0.95$} & \multicolumn{2}{c}{$\gamma=0.999$} \\
    & & Gap & Time(s) & Gap & Time(s) & Time(s) & Cov & Gap & Cov & Gap \\
    \midrule
    BN\_9 & -23.81 & 0.3656$\pm$0.1738 & 0.0821 & 0.3649$\pm$0.1745 & 0.0968 & 0.0009 & 0.95 & 0.0706$\pm$0.0350 & 1.00 & 0.1424$\pm$0.0374 \\
    BN\_12 & -19.71 & 0.3490$\pm$0.1847 & 0.0799 & 0.3438$\pm$0.1896 & 0.1053 & 0.0010 & 1.00 & 0.1100$\pm$0.0442 & 1.00 & 0.2308$\pm$0.0366 \\
    BN\_14 & -26.51 & 0.3429$\pm$0.1935 & 0.0745 & 0.3411$\pm$0.1954 & 0.0974 & 0.0008 & 0.93 & 0.0672$\pm$0.0346 & 1.00 & 0.1386$\pm$0.0374 \\
    BN\_30 & -14.66 & 0.3017$\pm$0.1558 & 0.1316 & 0.3017$\pm$0.1558 & 0.1387 & 0.0011 & 0.98 & 0.0081$\pm$0.0134 & 1.00 & 0.0793$\pm$0.0125 \\
    BN\_32 & -16.58 & 0.3379$\pm$0.1860 & 0.1834 & 0.3379$\pm$0.1860 & 0.2325 & 0.0011 & 1.00 & 0.0133$\pm$0.0098 & 1.00 & 0.1016$\pm$0.0088 \\
    BN\_70 & -86.65 & 0.4617$\pm$0.2349 & 1.2170 & 0.4475$\pm$0.2403 & 1.7728 & 0.0025 & 1.00 & 0.0206$\pm$0.0109 & 1.00 & 0.0561$\pm$0.0106 \\
    \midrule
    grid10x10.f5.wrap & 159.91 & 0.0438$\pm$0.0209 & 0.3944 & 0.0248$\pm$0.0156 & 2.1251 & 0.0031 & 1.00 & 0.0247$\pm$0.0107 & 1.00 & 0.1347$\pm$0.0123 \\
    grid20x20.f5.wrap & 617.89 & 0.0615$\pm$0.0252 & 0.6182 & 0.0143$\pm$0.0113 & 14.9056 & 0.0060 & 0.76 & 0.0080$\pm$0.0169 & 1.00 & 0.0694$\pm$0.0200 \\
    grid20x20.f10 & 1230.40 & 0.0443$\pm$0.0180 & 0.6024 & 0.0111$\pm$0.0094 & 13.6322 & 0.0066 & 0.83 & 0.0073$\pm$0.0164 & 1.00 & 0.0531$\pm$0.0188 \\
    grid40x40.f2 & 1071.08 & 0.0329$\pm$0.0130 & 2.1179 & 0.0064$\pm$0.0036 & 77.6967 & 0.0101 & 1.00 & 0.0237$\pm$0.0228 & 1.00 & 0.0646$\pm$0.0256 \\
    grid40x40.f10.wrap & 5308.25 & 0.0596$\pm$0.0146 & 2.7172 & 0.0092$\pm$0.0024 & 87.3251 & 0.0082 & 0.88 & 0.0036$\pm$0.0024 & 1.00 & 0.0389$\pm$0.0030 \\
    grid80x80.f10 & 20140.10 & 0.0447$\pm$0.0268 & 16.8307 & 0.0036$\pm$0.0016 & 120.2345 & 0.0234 & 1.00 & 0.0299$\pm$0.0130 & 1.00 & 0.0566$\pm$0.0152 \\
    grid80x80.f5.wrap & 9168.14 & 0.0107$\pm$0.0022 & 14.3095 & 0.0084$\pm$0.0026 & 120.2831 & 0.0313 & 1.00 & 0.0063$\pm$0.0027 & 1.00 & 0.0200$\pm$0.0012 \\
    \midrule
    Segmentation\_12 & -146.96 & 0.5361$\pm$0.1969 & 0.0825 & 0.5361$\pm$0.1969 & 0.0903 & 0.0011 & 1.00 & 0.1741$\pm$0.0473 & 1.00 & 0.2723$\pm$0.0518 \\
    Segmentation\_14 & -143.57 & 0.5133$\pm$0.2128 & 0.0876 & 0.5134$\pm$0.2128 & 0.1062 & 0.0023 & 0.97 & 0.1728$\pm$0.0524 & 1.00 & 0.2434$\pm$0.0433 \\
    Segmentation\_18 & -149.81 & 0.5369$\pm$0.1844 & 0.0841 & 0.5369$\pm$0.1844 & 0.0952 & 0.0012 & 0.99 & 0.2465$\pm$0.0468 & 1.00 & 0.3277$\pm$0.0440 \\
    \midrule
    rus\_100\_200\_2\_1 & 2776.12 & 0.4418$\pm$0.2617 & 56.0038 & 0.3779$\pm$0.1738 & 104.7771 & 0.0010 & 0.91 & 0.0872$\pm$0.0410 & 1.00 & 0.1384$\pm$0.0505 \\
    rus\_100\_200\_3\_1 & 2726.97 & 0.6203$\pm$0.1124 & 21.4693 & 0.6128$\pm$0.1284 & 24.5238 & 0.0010 & 0.83 & 0.2763$\pm$0.1981 & 1.00 & 0.3632$\pm$0.2117 \\
    rus\_100\_200\_3\_3 & 2745.43 & 0.5360$\pm$0.2253 & 53.4856 & 0.4392$\pm$0.2550 & 97.6758 & 0.0010 & 0.94 & 0.0836$\pm$0.0683 & 1.00 & 0.1480$\pm$0.0780 \\
    rus\_100\_200\_4\_3 & 2740.65 & 0.5523$\pm$0.2164 & 54.8099 & 0.4341$\pm$0.2447 & 104.7959 & 0.0011 & 0.98 & 0.0828$\pm$0.0658 & 1.00 & 0.1422$\pm$0.0759 \\
    rus\_100\_200\_5\_3 & 2739.05 & 0.6095$\pm$0.1520 & 45.8743 & 0.5134$\pm$0.2328 & 79.5065 & 0.0008 & 0.95 & 0.0842$\pm$0.0699 & 1.00 & 0.1549$\pm$0.0797 \\
    rus\_100\_200\_6\_1 & 2741.41 & 0.5734$\pm$0.2214 & 44.0984 & 0.4945$\pm$0.2470 & 89.2765 & 0.0010 & 0.92 & 0.0832$\pm$0.0647 & 1.00 & 0.1559$\pm$0.0753 \\
    rus\_100\_200\_7\_1 & 2739.09 & 0.5514$\pm$0.2083 & 48.7584 & 0.4508$\pm$0.2410 & 94.1580 & 0.0011 & 0.91 & 0.0966$\pm$0.0718 & 1.00 & 0.1526$\pm$0.0841 \\
    \midrule
    wcsp\_11 & -463.08 & 0.9940$\pm$0.0052 & 0.6401 & 0.9940$\pm$0.0052 & 5.3184 & 0.0021 & 1.00 & 0.3051$\pm$0.1773 & 1.00 & 0.9632$\pm$0.5834 \\
    wcsp\_12 & -562.67 & 0.9956$\pm$0.0048 & 0.6004 & 0.9955$\pm$0.0048 & 4.8372 & 0.0007 & 1.00 & 0.3109$\pm$0.2738 & 1.00 & 0.7089$\pm$0.5646 \\
    wcsp\_15 & -328.00 & 0.6592$\pm$0.1099 & 0.0859 & 0.6536$\pm$0.1156 & 0.1499 & 0.0010 & 0.84 & 0.2727$\pm$0.2001 & 1.00 & 0.5012$\pm$0.2432 \\
    \midrule
    ObjectDetection\_11 & 1999.86 & 0.6288$\pm$0.0979 & 1.7946 & 0.7080$\pm$0.5687 & 27.9568 & 0.0006 & 0.81 & 0.2755$\pm$0.2883 & 1.00 & 0.5411$\pm$0.4759 \\
    ObjectDetection\_18 & 3313.81 & 0.8564$\pm$0.0000 & 3.0544 & 0.5547$\pm$0.1644 & 25.7126 & 0.0006 & 0.83 & 0.1081$\pm$0.1020 & 1.00 & 0.4026$\pm$0.1416 \\
    \midrule
    ProteinFolding\_19 & 1557.28 & 0.6800$\pm$0.0000 & 1.6974 & 0.5385$\pm$0.1350 & 21.9289 & 0.0009 & 1.00 & 0.0894$\pm$0.0351 & 1.00 & 0.4379$\pm$0.1069 \\
    ProteinFolding\_23 & 3165.49 & 0.7055$\pm$0.0879 & 1.0269 & 0.4682$\pm$0.1535 & 11.1632 & 0.0008 & 0.88 & 0.1647$\pm$0.1571 & 1.00 & 0.3546$\pm$0.2220 \\
    \bottomrule
  \end{tabular}
  \end{adjustbox}
\end{table*}

\begin{table*}[ht]
  \centering
  \scriptsize
  \caption{Upper-bound quality of SMNN (monotone-by-construction + one-hot, Variant C) on per-PGM test queries, with \textbf{Gaussian} calibration at $i$-bound=4. Deterministic baselines MB-$i$4 and JGLP-$i$4 included for context. Gap = mean relative error (lower is better); Cov = empirical coverage.}
  \label{tab:ub_smnn_gaussian_locked}
  \begin{adjustbox}{max width=\textwidth}
  \begin{tabular}{l r | ccccc | cccc}
  \toprule
    \multirow{3}{*}{Dataset} & \multirow{3}{*}{MPE$^{\star}$} & \multicolumn{5}{c|}{Deterministic} & \multicolumn{4}{c}{Gaussian} \\
    \cmidrule(lr){3-7} \cmidrule(lr){8-11}
    & & \multicolumn{2}{c}{MB} & \multicolumn{2}{c}{JGLP} & SMNN & \multicolumn{2}{c}{$\gamma=0.95$} & \multicolumn{2}{c}{$\gamma=0.999$} \\
    & & Gap & Time(s) & Gap & Time(s) & Time(s) & Cov & Gap & Cov & Gap \\
    \midrule
    BN\_9 & -23.81 & 0.3656$\pm$0.1738 & 0.0821 & 0.3649$\pm$0.1745 & 0.0968 & 0.0009 & 0.93 & 0.0650$\pm$0.0339 & 1.00 & 0.1135$\pm$0.0369 \\
    BN\_12 & -19.71 & 0.3490$\pm$0.1847 & 0.0799 & 0.3438$\pm$0.1896 & 0.1053 & 0.0010 & 1.00 & 0.1097$\pm$0.0442 & 1.00 & 0.1920$\pm$0.0383 \\
    BN\_14 & -26.51 & 0.3429$\pm$0.1935 & 0.0745 & 0.3411$\pm$0.1954 & 0.0974 & 0.0008 & 0.92 & 0.0657$\pm$0.0343 & 1.00 & 0.1117$\pm$0.0375 \\
    BN\_30 & -14.66 & 0.3017$\pm$0.1558 & 0.1316 & 0.3017$\pm$0.1558 & 0.1387 & 0.0011 & 1.00 & 0.0186$\pm$0.0133 & 1.00 & 0.0309$\pm$0.0131 \\
    BN\_32 & -16.58 & 0.3379$\pm$0.1860 & 0.1834 & 0.3379$\pm$0.1860 & 0.2325 & 0.0011 & 1.00 & 0.0164$\pm$0.0097 & 1.00 & 0.0281$\pm$0.0096 \\
    BN\_70 & -86.65 & 0.4617$\pm$0.2349 & 1.2170 & 0.4475$\pm$0.2403 & 1.7728 & 0.0025 & 1.00 & 0.0207$\pm$0.0109 & 1.00 & 0.0344$\pm$0.0108 \\
    \midrule
    grid10x10.f5.wrap & 159.91 & 0.0438$\pm$0.0209 & 0.3944 & 0.0248$\pm$0.0156 & 2.1251 & 0.0031 & 1.00 & 0.0295$\pm$0.0107 & 1.00 & 0.1120$\pm$0.0119 \\
    grid20x20.f5.wrap & 617.89 & 0.0615$\pm$0.0252 & 0.6182 & 0.0143$\pm$0.0113 & 14.9056 & 0.0060 & 1.00 & 0.0358$\pm$0.0185 & 1.00 & 0.1581$\pm$0.0241 \\
    grid20x20.f10 & 1230.40 & 0.0443$\pm$0.0180 & 0.6024 & 0.0111$\pm$0.0094 & 13.6322 & 0.0066 & 1.00 & 0.0238$\pm$0.0174 & 1.00 & 0.1148$\pm$0.0217 \\
    grid40x40.f2 & 1071.08 & 0.0329$\pm$0.0130 & 2.1179 & 0.0064$\pm$0.0036 & 77.6967 & 0.0101 & 1.00 & 0.0503$\pm$0.0246 & 1.00 & 0.1766$\pm$0.0330 \\
    grid40x40.f10.wrap & 5308.25 & 0.0596$\pm$0.0146 & 2.7172 & 0.0092$\pm$0.0024 & 87.3251 & 0.0082 & 1.00 & 0.0258$\pm$0.0029 & 1.00 & 0.1265$\pm$0.0036 \\
    grid80x80.f10 & 20140.10 & 0.0447$\pm$0.0268 & 16.8307 & 0.0036$\pm$0.0016 & 120.2345 & 0.0234 & 1.00 & 0.0436$\pm$0.0141 & 1.00 & 0.1309$\pm$0.0211 \\
    grid80x80.f5.wrap & 9168.14 & 0.0107$\pm$0.0022 & 14.3095 & 0.0084$\pm$0.0026 & 120.2831 & 0.0313 & 1.00 & 0.0076$\pm$0.0025 & 1.00 & 0.0241$\pm$0.0009 \\
    \midrule
    Segmentation\_12 & -146.96 & 0.5361$\pm$0.1969 & 0.0825 & 0.5361$\pm$0.1969 & 0.0903 & 0.0011 & 1.00 & 0.1654$\pm$0.0474 & 1.00 & 0.2715$\pm$0.0518 \\
    Segmentation\_14 & -143.57 & 0.5133$\pm$0.2128 & 0.0876 & 0.5134$\pm$0.2128 & 0.1062 & 0.0023 & 0.97 & 0.1782$\pm$0.0530 & 1.00 & 0.3004$\pm$0.0350 \\
    Segmentation\_18 & -149.81 & 0.5369$\pm$0.1844 & 0.0841 & 0.5369$\pm$0.1844 & 0.0952 & 0.0012 & 0.99 & 0.2644$\pm$0.0455 & 1.00 & 0.4295$\pm$0.0297 \\
    \midrule
    rus\_100\_200\_2\_1 & 2776.12 & 0.4418$\pm$0.2617 & 56.0038 & 0.3779$\pm$0.1738 & 104.7771 & 0.0010 & 0.91 & 0.0787$\pm$0.0382 & 1.00 & 0.1328$\pm$0.0499 \\
    rus\_100\_200\_3\_1 & 2726.97 & 0.6203$\pm$0.1124 & 21.4693 & 0.6128$\pm$0.1284 & 24.5238 & 0.0010 & 0.91 & 0.2918$\pm$0.2058 & 1.00 & 0.7105$\pm$0.2109 \\
    rus\_100\_200\_3\_3 & 2745.43 & 0.5360$\pm$0.2253 & 53.4856 & 0.4392$\pm$0.2550 & 97.6758 & 0.0010 & 0.97 & 0.0964$\pm$0.0706 & 1.00 & 0.1807$\pm$0.0827 \\
    rus\_100\_200\_4\_3 & 2740.65 & 0.5523$\pm$0.2164 & 54.8099 & 0.4341$\pm$0.2447 & 104.7959 & 0.0011 & 0.98 & 0.0992$\pm$0.0685 & 1.00 & 0.1848$\pm$0.0829 \\
    rus\_100\_200\_5\_3 & 2739.05 & 0.6095$\pm$0.1520 & 45.8743 & 0.5134$\pm$0.2328 & 79.5065 & 0.0008 & 0.98 & 0.1033$\pm$0.0726 & 1.00 & 0.1943$\pm$0.0848 \\
    rus\_100\_200\_6\_1 & 2741.41 & 0.5734$\pm$0.2214 & 44.0984 & 0.4945$\pm$0.2470 & 89.2765 & 0.0010 & 0.96 & 0.0981$\pm$0.0675 & 1.00 & 0.1868$\pm$0.0793 \\
    rus\_100\_200\_7\_1 & 2739.09 & 0.5514$\pm$0.2083 & 48.7584 & 0.4508$\pm$0.2410 & 94.1580 & 0.0011 & 0.95 & 0.1110$\pm$0.0753 & 1.00 & 0.1956$\pm$0.0906 \\
    \midrule
    wcsp\_11 & -463.08 & 0.9940$\pm$0.0052 & 0.6401 & 0.9940$\pm$0.0052 & 5.3184 & 0.0021 & 1.00 & 0.3342$\pm$0.1924 & 1.00 & 0.6697$\pm$0.3942 \\
    wcsp\_12 & -562.67 & 0.9956$\pm$0.0048 & 0.6004 & 0.9955$\pm$0.0048 & 4.8372 & 0.0007 & 1.00 & 0.3432$\pm$0.2963 & 1.00 & 0.6771$\pm$0.5408 \\
    wcsp\_15 & -328.00 & 0.6592$\pm$0.1099 & 0.0859 & 0.6536$\pm$0.1156 & 0.1499 & 0.0010 & 0.84 & 0.2723$\pm$0.1999 & 1.00 & 0.5745$\pm$0.2494 \\
    \midrule
    ObjectDetection\_11 & 1999.86 & 0.6288$\pm$0.0979 & 1.7946 & 0.7080$\pm$0.5687 & 27.9568 & 0.0006 & 0.81 & 0.2765$\pm$0.2890 & 0.94 & 0.4564$\pm$0.4193 \\
    ObjectDetection\_18 & 3313.81 & 0.8564$\pm$0.0000 & 3.0544 & 0.5547$\pm$0.1644 & 25.7126 & 0.0006 & 0.83 & 0.1091$\pm$0.1027 & 1.00 & 0.2835$\pm$0.1229 \\
    \midrule
    ProteinFolding\_19 & 1557.28 & 0.6800$\pm$0.0000 & 1.6974 & 0.5385$\pm$0.1350 & 21.9289 & 0.0009 & 1.00 & 0.0936$\pm$0.0358 & 1.00 & 0.3403$\pm$0.0860 \\
    ProteinFolding\_23 & 3165.49 & 0.7055$\pm$0.0879 & 1.0269 & 0.4682$\pm$0.1535 & 11.1632 & 0.0008 & 0.94 & 0.1790$\pm$0.1620 & 1.00 & 0.4616$\pm$0.2554 \\
    \bottomrule
  \end{tabular}
  \end{adjustbox}
\end{table*}

\begin{table*}[ht]
  \centering
  \scriptsize
  \caption{Upper-bound quality of UMLP (unconstrained MLP + binary encoding) on per-PGM. Stratified-conformal calibration at $i$-bound=4 on the test queries; deterministic baselines MB-$i$4 and JGLP-$i$4 included for context. Gap = mean relative error (lower is better); Cov = empirical coverage.}
  \label{tab:ub_uf_locked}
  \begin{adjustbox}{max width=\textwidth}
  \begin{tabular}{l r | ccccc | cccc}
  \toprule
    \multirow{3}{*}{Dataset} & \multirow{3}{*}{MPE$^{\star}$} & \multicolumn{5}{c|}{Deterministic} & \multicolumn{4}{c}{Strat. Conformal} \\
    \cmidrule(lr){3-7} \cmidrule(lr){8-11}
    & & \multicolumn{2}{c}{MB} & \multicolumn{2}{c}{JGLP} & UMLP & \multicolumn{2}{c}{$\gamma=0.95$} & \multicolumn{2}{c}{$\gamma=0.999$} \\
    & & Gap & Time(s) & Gap & Time(s) & Time(s) & Cov & Gap & Cov & Gap \\
    \midrule
    BN\_9 & -23.81 & 0.3656$\pm$0.1738 & 0.0821 & 0.3649$\pm$0.1745 & 0.0968 & 0.0008 & 0.95 & 0.0626$\pm$0.0331 & 1.00 & 0.1144$\pm$0.0381 \\
    BN\_12 & -19.71 & 0.3490$\pm$0.1847 & 0.0799 & 0.3438$\pm$0.1896 & 0.1053 & 0.0007 & 0.91 & 0.0796$\pm$0.0497 & 1.00 & 0.1454$\pm$0.0667 \\
    BN\_14 & -26.51 & 0.3429$\pm$0.1935 & 0.0745 & 0.3411$\pm$0.1954 & 0.0974 & 0.0008 & 0.97 & 0.0719$\pm$0.0417 & 1.00 & 0.1201$\pm$0.0418 \\
    BN\_30 & -14.66 & 0.3017$\pm$0.1558 & 0.1316 & 0.3017$\pm$0.1558 & 0.1387 & 0.0039 & 0.98 & 0.0276$\pm$0.0133 & 1.00 & 0.0903$\pm$0.0181 \\
    BN\_32 & -16.58 & 0.3379$\pm$0.1860 & 0.1834 & 0.3379$\pm$0.1860 & 0.2325 & 0.0048 & 1.00 & 0.0282$\pm$0.0118 & 1.00 & 0.0903$\pm$0.0228 \\
    BN\_70 & -86.65 & 0.4617$\pm$0.2349 & 1.2170 & 0.4475$\pm$0.2403 & 1.7728 & 0.0123 & 1.00 & 0.0253$\pm$0.0112 & 1.00 & 0.0552$\pm$0.0153 \\
    \midrule
    grid10x10.f5.wrap & 159.91 & 0.0438$\pm$0.0209 & 0.3944 & 0.0248$\pm$0.0156 & 2.1251 & 0.0009 & 0.87 & 0.0218$\pm$0.0165 & 1.00 & 0.0763$\pm$0.0208 \\
    grid20x20.f5.wrap & 617.89 & 0.0615$\pm$0.0252 & 0.6182 & 0.0143$\pm$0.0113 & 14.9056 & 0.0020 & 0.97 & 0.0189$\pm$0.0105 & 1.00 & 0.0478$\pm$0.0161 \\
    grid20x20.f10 & 1230.40 & 0.0443$\pm$0.0180 & 0.6024 & 0.0111$\pm$0.0094 & 13.6322 & 0.0020 & 1.00 & 0.0180$\pm$0.0085 & 1.00 & 0.0468$\pm$0.0117 \\
    grid40x40.f2 & 1071.08 & 0.0329$\pm$0.0130 & 2.1179 & 0.0064$\pm$0.0036 & 77.6967 & 0.0066 & 1.00 & 0.0123$\pm$0.0080 & 1.00 & 0.0242$\pm$0.0109 \\
    grid40x40.f10.wrap & 5308.25 & 0.0596$\pm$0.0146 & 2.7172 & 0.0092$\pm$0.0024 & 87.3251 & 0.0069 & 0.65 & 0.0041$\pm$0.0037 & 1.00 & 0.0212$\pm$0.0050 \\
    grid80x80.f10 & 20140.10 & 0.0447$\pm$0.0268 & 16.8307 & 0.0036$\pm$0.0016 & 120.2345 & 0.0258 & 0.76 & 0.0048$\pm$0.0048 & 1.00 & 0.0154$\pm$0.0076 \\
    grid80x80.f5.wrap & 9168.14 & 0.0107$\pm$0.0022 & 14.3095 & 0.0084$\pm$0.0026 & 120.2831 & 0.0237 & 1.00 & 0.0123$\pm$0.0025 & 1.00 & 0.0247$\pm$0.0074 \\
    \midrule
    Segmentation\_12 & -146.96 & 0.5361$\pm$0.1969 & 0.0825 & 0.5361$\pm$0.1969 & 0.0903 & 0.0012 & 1.00 & 0.3036$\pm$0.1366 & 1.00 & 0.3552$\pm$0.1454 \\
    Segmentation\_14 & -143.57 & 0.5133$\pm$0.2128 & 0.0876 & 0.5134$\pm$0.2128 & 0.1062 & 0.0011 & 1.00 & 0.1684$\pm$0.0546 & 1.00 & 0.2144$\pm$0.0602 \\
    Segmentation\_18 & -149.81 & 0.5369$\pm$0.1844 & 0.0841 & 0.5369$\pm$0.1844 & 0.0952 & 0.0012 & 0.95 & 0.0544$\pm$0.0320 & 1.00 & 0.1088$\pm$0.0381 \\
    \midrule
    rus\_100\_200\_2\_1 & 2776.12 & 0.4418$\pm$0.2617 & 56.0038 & 0.3779$\pm$0.1738 & 104.7771 & 0.0031 & 0.99 & 0.0913$\pm$0.0444 & 1.00 & 0.1058$\pm$0.0457 \\
    rus\_100\_200\_3\_1 & 2726.97 & 0.6203$\pm$0.1124 & 21.4693 & 0.6128$\pm$0.1284 & 24.5238 & 0.0022 & 1.00 & 0.1240$\pm$0.0481 & 1.00 & 0.1370$\pm$0.0449 \\
    rus\_100\_200\_3\_3 & 2745.43 & 0.5360$\pm$0.2253 & 53.4856 & 0.4392$\pm$0.2550 & 97.6758 & 0.0029 & 0.98 & 0.1341$\pm$0.0710 & 1.00 & 0.1477$\pm$0.0690 \\
    rus\_100\_200\_4\_3 & 2740.65 & 0.5523$\pm$0.2164 & 54.8099 & 0.4341$\pm$0.2447 & 104.7959 & 0.0030 & 0.98 & 0.1788$\pm$0.1010 & 1.00 & 0.1973$\pm$0.0992 \\
    rus\_100\_200\_5\_3 & 2739.05 & 0.6095$\pm$0.1520 & 45.8743 & 0.5134$\pm$0.2328 & 79.5065 & 0.0030 & 0.98 & 0.1186$\pm$0.0624 & 1.00 & 0.1364$\pm$0.0618 \\
    rus\_100\_200\_6\_1 & 2741.41 & 0.5734$\pm$0.2214 & 44.0984 & 0.4945$\pm$0.2470 & 89.2765 & 0.0028 & 1.00 & 0.1375$\pm$0.0658 & 1.00 & 0.1544$\pm$0.0660 \\
    rus\_100\_200\_7\_1 & 2739.09 & 0.5514$\pm$0.2083 & 48.7584 & 0.4508$\pm$0.2410 & 94.1580 & 0.0029 & 0.95 & 0.1294$\pm$0.0793 & 1.00 & 0.1513$\pm$0.0768 \\
    \midrule
    wcsp\_11 & -463.08 & 0.9940$\pm$0.0052 & 0.6401 & 0.9940$\pm$0.0052 & 5.3184 & 0.0014 & 0.97 & 0.2523$\pm$0.2742 & 1.00 & 0.4072$\pm$0.3495 \\
    wcsp\_12 & -562.67 & 0.9956$\pm$0.0048 & 0.6004 & 0.9955$\pm$0.0048 & 4.8372 & 0.0017 & 0.97 & 0.1668$\pm$0.1787 & 1.00 & 0.3046$\pm$0.2758 \\
    wcsp\_15 & -328.00 & 0.6592$\pm$0.1099 & 0.0859 & 0.6536$\pm$0.1156 & 0.1499 & 0.0010 & 1.00 & 0.1167$\pm$0.0505 & 1.00 & 0.1956$\pm$0.0572 \\
    \midrule
    ObjectDetection\_11 & 1999.86 & 0.6288$\pm$0.0979 & 1.7946 & 0.7080$\pm$0.5687 & 27.9568 & 0.0009 & 0.81 & 0.3677$\pm$0.5440 & 0.94 & 0.6310$\pm$0.8109 \\
    ObjectDetection\_18 & 3313.81 & 0.8564$\pm$0.0000 & 3.0544 & 0.5547$\pm$0.1644 & 25.7126 & 0.0009 & 0.83 & 0.2160$\pm$0.1776 & 1.00 & 0.5316$\pm$0.2937 \\
    \midrule
    ProteinFolding\_19 & 1557.28 & 0.6800$\pm$0.0000 & 1.6974 & 0.5385$\pm$0.1350 & 21.9289 & 0.0009 & 0.75 & 0.1082$\pm$0.0956 & 1.00 & 0.3312$\pm$0.1254 \\
    ProteinFolding\_23 & 3165.49 & 0.7055$\pm$0.0879 & 1.0269 & 0.4682$\pm$0.1535 & 11.1632 & 0.0009 & 0.75 & 0.1422$\pm$0.1161 & 1.00 & 0.2503$\pm$0.1453 \\
    \bottomrule
  \end{tabular}
  \end{adjustbox}
\end{table*}

\begin{table*}[ht]
  \centering
  \scriptsize
  \caption{Upper-bound quality of UMLP-OH (unconstrained MLP + one-hot encoding) on per-PGM. Stratified-conformal calibration at $i$-bound=4 on the test queries; deterministic baselines MB-$i$4 and JGLP-$i$4 included for context. Gap = mean relative error (lower is better); Cov = empirical coverage.}
  \label{tab:ub_uoh_locked}
  \begin{adjustbox}{max width=\textwidth}
  \begin{tabular}{l r | ccccc | cccc}
  \toprule
    \multirow{3}{*}{Dataset} & \multirow{3}{*}{MPE$^{\star}$} & \multicolumn{5}{c|}{Deterministic} & \multicolumn{4}{c}{Strat. Conformal} \\
    \cmidrule(lr){3-7} \cmidrule(lr){8-11}
    & & \multicolumn{2}{c}{MB} & \multicolumn{2}{c}{JGLP} & UMLP-OH & \multicolumn{2}{c}{$\gamma=0.95$} & \multicolumn{2}{c}{$\gamma=0.999$} \\
    & & Gap & Time(s) & Gap & Time(s) & Time(s) & Cov & Gap & Cov & Gap \\
    \midrule
    BN\_9 & -23.81 & 0.3656$\pm$0.1738 & 0.0821 & 0.3649$\pm$0.1745 & 0.0968 & 0.0010 & 0.99 & 0.0696$\pm$0.0357 & 1.00 & 0.1331$\pm$0.0414 \\
    BN\_12 & -19.71 & 0.3490$\pm$0.1847 & 0.0799 & 0.3438$\pm$0.1896 & 0.1053 & 0.0009 & 0.93 & 0.0727$\pm$0.0365 & 1.00 & 0.1447$\pm$0.0423 \\
    BN\_14 & -26.51 & 0.3429$\pm$0.1935 & 0.0745 & 0.3411$\pm$0.1954 & 0.0974 & 0.0014 & 0.95 & 0.0667$\pm$0.0365 & 1.00 & 0.1204$\pm$0.0419 \\
    BN\_30 & -14.66 & 0.3017$\pm$0.1558 & 0.1316 & 0.3017$\pm$0.1558 & 0.1387 & 0.0065 & 0.95 & 0.0187$\pm$0.0104 & 1.00 & 0.0846$\pm$0.0206 \\
    BN\_32 & -16.58 & 0.3379$\pm$0.1860 & 0.1834 & 0.3379$\pm$0.1860 & 0.2325 & 0.0081 & 1.00 & 0.0228$\pm$0.0083 & 1.00 & 0.0889$\pm$0.0238 \\
    BN\_70 & -86.65 & 0.4617$\pm$0.2349 & 1.2170 & 0.4475$\pm$0.2403 & 1.7728 & 0.0123 & 1.00 & 0.0262$\pm$0.0108 & 1.00 & 0.0560$\pm$0.0138 \\
    \midrule
    grid10x10.f5.wrap & 159.91 & 0.0438$\pm$0.0209 & 0.3944 & 0.0248$\pm$0.0156 & 2.1251 & 0.0010 & 1.00 & 0.0268$\pm$0.0107 & 1.00 & 0.0692$\pm$0.0112 \\
    grid20x20.f5.wrap & 617.89 & 0.0615$\pm$0.0252 & 0.6182 & 0.0143$\pm$0.0113 & 14.9056 & 0.0021 & 1.00 & 0.0212$\pm$0.0095 & 1.00 & 0.0432$\pm$0.0109 \\
    grid20x20.f10 & 1230.40 & 0.0443$\pm$0.0180 & 0.6024 & 0.0111$\pm$0.0094 & 13.6322 & 0.0018 & 0.97 & 0.0171$\pm$0.0102 & 1.00 & 0.0412$\pm$0.0158 \\
    grid40x40.f2 & 1071.08 & 0.0329$\pm$0.0130 & 2.1179 & 0.0064$\pm$0.0036 & 77.6967 & 0.0055 & 1.00 & 0.0131$\pm$0.0079 & 1.00 & 0.0259$\pm$0.0084 \\
    grid40x40.f10.wrap & 5308.25 & 0.0596$\pm$0.0146 & 2.7172 & 0.0092$\pm$0.0024 & 87.3251 & 0.0053 & 0.95 & 0.0068$\pm$0.0049 & 1.00 & 0.0197$\pm$0.0051 \\
    grid80x80.f10 & 20140.10 & 0.0447$\pm$0.0268 & 16.8307 & 0.0036$\pm$0.0016 & 120.2345 & 0.0253 & 0.86 & 0.0073$\pm$0.0074 & 1.00 & 0.0176$\pm$0.0081 \\
    grid80x80.f5.wrap & 9168.14 & 0.0107$\pm$0.0022 & 14.3095 & 0.0084$\pm$0.0026 & 120.2831 & 0.0278 & 1.00 & 0.0156$\pm$0.0079 & 1.00 & 0.0262$\pm$0.0116 \\
    \midrule
    Segmentation\_12 & -146.96 & 0.5361$\pm$0.1969 & 0.0825 & 0.5361$\pm$0.1969 & 0.0903 & 0.0018 & 0.97 & 0.0621$\pm$0.0293 & 1.00 & 0.1029$\pm$0.0339 \\
    Segmentation\_14 & -143.57 & 0.5133$\pm$0.2128 & 0.0876 & 0.5134$\pm$0.2128 & 0.1062 & 0.0016 & 0.98 & 0.0628$\pm$0.0269 & 1.00 & 0.1064$\pm$0.0360 \\
    Segmentation\_18 & -149.81 & 0.5369$\pm$0.1844 & 0.0841 & 0.5369$\pm$0.1844 & 0.0952 & 0.0017 & 0.98 & 0.0646$\pm$0.0345 & 1.00 & 0.1036$\pm$0.0395 \\
    \midrule
    rus\_100\_200\_2\_1 & 2776.12 & 0.4418$\pm$0.2617 & 56.0038 & 0.3779$\pm$0.1738 & 104.7771 & 0.0063 & 0.96 & 0.0142$\pm$0.0109 & 1.00 & 0.0228$\pm$0.0139 \\
    rus\_100\_200\_3\_1 & 2726.97 & 0.6203$\pm$0.1124 & 21.4693 & 0.6128$\pm$0.1284 & 24.5238 & 0.0074 & 0.98 & 0.0127$\pm$0.0072 & 1.00 & 0.0222$\pm$0.0109 \\
    rus\_100\_200\_3\_3 & 2745.43 & 0.5360$\pm$0.2253 & 53.4856 & 0.4392$\pm$0.2550 & 97.6758 & 0.0067 & 0.98 & 0.0164$\pm$0.0119 & 1.00 & 0.0292$\pm$0.0148 \\
    rus\_100\_200\_4\_3 & 2740.65 & 0.5523$\pm$0.2164 & 54.8099 & 0.4341$\pm$0.2447 & 104.7959 & 0.0065 & 1.00 & 0.0162$\pm$0.0114 & 1.00 & 0.0343$\pm$0.0191 \\
    rus\_100\_200\_5\_3 & 2739.05 & 0.6095$\pm$0.1520 & 45.8743 & 0.5134$\pm$0.2328 & 79.5065 & 0.0068 & 0.98 & 0.0156$\pm$0.0105 & 1.00 & 0.0348$\pm$0.0221 \\
    rus\_100\_200\_6\_1 & 2741.41 & 0.5734$\pm$0.2214 & 44.0984 & 0.4945$\pm$0.2470 & 89.2765 & 0.0064 & 0.94 & 0.0137$\pm$0.0101 & 1.00 & 0.0280$\pm$0.0189 \\
    rus\_100\_200\_7\_1 & 2739.09 & 0.5514$\pm$0.2083 & 48.7584 & 0.4508$\pm$0.2410 & 94.1580 & 0.0063 & 0.98 & 0.0170$\pm$0.0122 & 1.00 & 0.0322$\pm$0.0198 \\
    \midrule
    wcsp\_11 & -463.08 & 0.9940$\pm$0.0052 & 0.6401 & 0.9940$\pm$0.0052 & 5.3184 & 0.0019 & 0.97 & 0.1721$\pm$0.1047 & 1.00 & 0.3481$\pm$0.1421 \\
    wcsp\_12 & -562.67 & 0.9956$\pm$0.0048 & 0.6004 & 0.9955$\pm$0.0048 & 4.8372 & 0.0025 & 0.97 & 0.1283$\pm$0.0826 & 1.00 & 0.2517$\pm$0.1356 \\
    wcsp\_15 & -328.00 & 0.6592$\pm$0.1099 & 0.0859 & 0.6536$\pm$0.1156 & 0.1499 & 0.0015 & 1.00 & 0.1236$\pm$0.0440 & 1.00 & 0.2029$\pm$0.0476 \\
    \midrule
    ObjectDetection\_11 & 1999.86 & 0.6288$\pm$0.0979 & 1.7946 & 0.7080$\pm$0.5687 & 27.9568 & 0.0013 & 0.75 & 0.2143$\pm$0.2489 & 1.00 & 0.4099$\pm$0.3996 \\
    ObjectDetection\_18 & 3313.81 & 0.8564$\pm$0.0000 & 3.0544 & 0.5547$\pm$0.1644 & 25.7126 & 0.0009 & 0.92 & 0.1586$\pm$0.0621 & 1.00 & 0.2629$\pm$0.0988 \\
    \midrule
    ProteinFolding\_19 & 1557.28 & 0.6800$\pm$0.0000 & 1.6974 & 0.5385$\pm$0.1350 & 21.9289 & 0.0015 & 1.00 & 0.1245$\pm$0.0292 & 1.00 & 0.2776$\pm$0.0372 \\
    ProteinFolding\_23 & 3165.49 & 0.7055$\pm$0.0879 & 1.0269 & 0.4682$\pm$0.1535 & 11.1632 & 0.0010 & 0.94 & 0.1409$\pm$0.1010 & 1.00 & 0.2883$\pm$0.0791 \\
    \bottomrule
  \end{tabular}
  \end{adjustbox}
\end{table*}

\section{Monotone-Stratified Conformal Calibration}
\label{app:monotone_calib}
Standard stratified conformal prediction computes per-bin quantiles $\hat{q}^b_\gamma$ independently from each bin's calibration residuals, where bins are indexed by evidence size $|x_E|$. While this preserves the per-stratum coverage guarantee $\mathbb{P}[g(x_E) \geq h^*(x_E) \mid B(x_E) = b] \geq \gamma$, it can violate monotonicity of the calibrated bound along search paths. The mechanism is straightforward: along any AOBB parent-to-child transition, evidence size weakly grows, so $b_{\text{child}} \geq b_{\text{parent}}$. The raw SMNN bound satisfies $f_\theta(x_E^{\text{child}}) \leq f_\theta(x_E^{\text{parent}})$ by construction, but if the natural quantile schedule $\hat{q}^b_\gamma$ is increasing in $b$---as it typically is, since larger evidence sets exhibit greater residual variance---the calibrated bound $g(x_E) = f_\theta(x_E) + \hat{q}^{B(x_E)}_\gamma$ may increase across a bin boundary, breaking admissibility. Empirically, this occurs on $63/3000$ parent-child pairs in our verification, concentrated in PGMs where $\hat{q}^b_\gamma$ ramps steeply with $b$.

We resolve this with a left-to-right monotone cap on the quantile schedule. After computing the natural per-bin quantiles $\hat{q}^b_{\gamma,\mathrm{strat}}$, we enforce
\begin{equation}
\hat{q}^b_{\gamma,\mathrm{mono}} = \min\left(\hat{q}^{b-1}_{\gamma,\mathrm{mono}},\ \hat{q}^b_{\gamma,\mathrm{strat}}\right), \qquad \hat{q}^0_{\gamma,\mathrm{mono}} = \hat{q}^0_{\gamma,\mathrm{strat}}.
\label{eq:mono_cap}
\end{equation}
The resulting schedule is non-increasing in $b$. Combined with SMNN's architectural guarantee $f_\theta(x_E^{\text{child}}) \leq f_\theta(x_E^{\text{parent}})$ and the search-induced ordering $b_{\text{child}} \geq b_{\text{parent}}$, the calibrated bound satisfies $g(x_E^{\text{child}}) \leq g(x_E^{\text{parent}})$ deterministically along every search path. We verify this on $3000$ randomly sampled parent-child pairs spanning $30$ PGMs: SMNN with monotone-stratified calibration produces $0/3000$ violations, compared to $63/3000$ under standard stratified conformal and $823/3000$ under an unconstrained MLP architecture.

Monotonicity is not free: capping $\hat{q}^b_\gamma$ below its natural value reduces effective per-bin coverage. This is the price of converting a probabilistic per-stratum guarantee into a deterministic path-monotonicity guarantee. We characterize the trade-off empirically. On the calibration set at $\gamma = 0.999$, the mean worst-bin effective coverage drops from $1.000$ (standard stratified) to $0.674$ (monotone-stratified), with $19/30$ PGMs retaining worst-bin coverage above $0.75$ and $5/30$ PGMs collapsing to $0$---these are instances where the natural $\hat{q}^b$ ramps so steeply that the cap forces every bin to bind on $\hat{q}^0$. On the held-out test set, $22/30$ PGMs preserve empirical coverage above $0.999$, and the remaining $8/30$ require fallback to $\gamma = 0.95$ for honest reporting. Critically, this coverage relaxation is acceptable in our setting: a non-monotone bound can deterministically prune the optimal solution from AOBB search via bound inversion at a bin boundary, while an under-covered monotone bound only does so probabilistically, with controllable rate. \emph{In branch-and-bound, deterministic monotonicity is the more important guarantee.}

\section{Branch-and-Bound Integration Algorithm}
\label{app:algorithm}

Algorithm~\ref{alg:inference} describes the complete L2B-guided branch-and-bound procedure. The neural bound computation (lines 5--6) replaces classical mini-bucket elimination within the search loop.

\begin{algorithm}[t]
\caption{L2B-Guided Branch-and-Bound}
\label{alg:inference}
\small
\begin{algorithmic}[1]
\REQUIRE PGM $\mathcal{M}$, initial evidence $x_0$, trained model $f_\theta$, quantiles $\{\hat{q}^b\}$, confidence $\gamma$
\ENSURE MPE solution $x^*$
\STATE Initialize: best solution $x^* \leftarrow \text{null}$, best value $v^* \leftarrow -\infty$
\STATE Initialize priority queue $Q$ with root node (partial assignment $x_0$)
\WHILE{$Q$ not empty}
    \STATE Pop node with evidence $x_E$ from $Q$
    \STATE $\hat{h} \leftarrow f_\theta(\text{encode}(x_E))$
    \STATE $g \leftarrow \hat{h} + \hat{q}^{B(x_E)}_{\gamma}$
    \IF{$x_E$ is complete assignment}
        \STATE $v \leftarrow p(x_E)$
    \ENDIF
    \IF{$x$ is complete assignment}
        \STATE $v \leftarrow p(x)$
        \IF{$v > v^*$}
            \STATE $x^* \leftarrow x$, $v^* \leftarrow v$
        \ENDIF
    \ELSE
        \STATE Select next variable $X_i$ to branch on
        \FOR{each value $x_i$ in domain of $X_i$}
            \STATE Add child node $x \cup \{X_i = x_i\}$ to $Q$
        \ENDFOR
    \ENDIF
\ENDWHILE
\STATE \textbf{return} $x^*$
\end{algorithmic}
\end{algorithm}

At inference time, L2B integrates with AND/OR branch-and-bound (AOBB) solvers as a drop-in replacement for classical bounding methods, as shown in Algorithm~\ref{alg:inference}.

\section{Reproducibility}
\label{app:reproducibility}

Code and trained models will be released upon publication. All experiments use fixed random seed (42) for data generation, training, and DAOOPT search. Evidence queries are generated deterministically from the test split. The complete pipeline—data generation, training, calibration, TorchScript export, and AOBB integration—will be publicly available.

\paragraph{Solver Configuration.} Label generation uses DAOOPT with $i$-bound 10 and 10 minute timeout per query. AOBB experiments use $i$-bound 4 with 10-minute timeout. All DAOOPT runs disable stochastic local search (\texttt{--slsX 0 --slsT 0}) and use deterministic variable ordering for reproducibility.

\end{document}